\documentclass[format=acmsmall, review=false, screen=true]{acmart}

\usepackage{booktabs} % For formal tables

\usepackage[ruled]{algorithm2e} % For algorithms
\renewcommand{\algorithmcfname}{ALGORITHM}
\SetAlFnt{\small}
\SetAlCapFnt{\small}
\SetAlCapNameFnt{\small}
\SetAlCapHSkip{0pt}
\IncMargin{-\parindent}

\acmJournal{CSUR}
\setcopyright{acmlicensed}
\usepackage{verbatim}
\usepackage{mathtools}

\usepackage[ruled]{algorithm2e}
\renewcommand{\algorithmcfname}{ALGORITHM}
\SetAlFnt{\small}
\SetAlCapFnt{\small}
\SetAlCapNameFnt{\small}
\SetAlCapHSkip{0pt}
\IncMargin{-\parindent}
\usepackage{graphicx}
\usepackage{subfigure}

\usepackage{wrapfig}
\begin{document}

\title{Spatio-Temporal Data Mining: A Survey of Problems and Methods}  

\author{Gowtham Atluri}
\affiliation{%
  \institution{University of Cincinnati}}
\email{atlurigm@ucmail.uc.edu}
\authornote{Both authors contributed equally to the paper.}
\author{Anuj Karpatne}
\affiliation{%
  \institution{University of Minnesota}}
\email{karpa009@umn.edu}
\authornotemark[1]
\author{Vipin Kumar}
\affiliation{%
  \institution{University of Minnesota}}
\email{kumar001@umn.edu}

\begin{abstract}
Large volumes of spatio-temporal data are increasingly collected and studied in diverse domains including, climate science, social sciences, neuroscience, epidemiology, transportation, mobile health, and Earth sciences. Spatio-temporal data differs from relational data for which computational approaches are developed in the data mining community for multiple decades, in that both spatial and temporal attributes are available in addition to the actual measurements/attributes. The presence of these attributes introduces additional challenges that needs to be dealt with. Approaches for mining spatio-temporal data have been studied for over a decade in the data mining community. In this article we present a broad survey of this relatively young field of spatio-temporal data mining. We discuss different types of spatio-temporal data and the relevant data mining questions that arise in the context of analyzing each of these datasets. Based on the nature of the data mining problem studied, we classify literature on spatio-temporal data mining into six major categories: clustering, predictive learning, change detection, frequent pattern mining, anomaly detection, and relationship mining. We discuss the various forms of spatio-temporal data mining problems in each of these categories.   
\end{abstract}

\maketitle

\renewcommand{\shortauthors}{Atluri et al.}

% \vspace{-1ex}
\section{Introduction}

Space and time are ubiquitous aspects of observations in a number of domains, including, climate science, neuroscience, social sciences, epidemiology, transportation, criminology, and Earth sciences, that are rapidly being transformed by the deluge of data. Since the real-world processes being studied in these domains are inherently spatio-temporal in nature, a number of data collection methodologies have been devised to record the spatial and temporal information of every measurement in the data, hereby referred to as spatio-temporal (ST) data. For example, in neuroimaging data, activity measured from the human brain is stored along with the spatial location from which the activity was measured and the time at which the measurement was made. Similarly, web-search requests arriving at Google's servers have a geographic location and time from which they are made. Effective analysis of such increasingly prevalent ST data holds great promise for advancing the state-of-the-art in several scientific disciplines.

A unique quality of ST data that differentiates it from other data studied in classical data mining literature (e.g., see \cite{tan2018introduction}) is the presence of dependencies among measurements induced by the spatial and temporal dimensions. 
% The presence of these dependencies in the data poses challenges to classical data mining algorithms that are typically designed for relational data. 
For example, many of the widely used data mining methods are founded on the assumption that data instances are independent and identically distributed ($i.i.d$). However, this assumption is violated when dealing with ST data, where instances are structurally related to each other in the context of space and time and show varying properties in different spatial regions and time periods. Ignoring these dependencies during data analysis can lead to poor accuracy and interpretability of results \cite{eklund2016cluster}. 

Apart from limiting the effectiveness of classical data mining algorithms, the presence of spatial and temporal information also makes it possible to consider novel formulations for analyzing data in the emerging field of spatio-temporal data mining (STDM). Contrary to traditional data mining that deals with distinct objects (also referred to as data instances) having well-defined features, in STDM, one can define objects and features in a variety of ways. One scenario involves treating spatial locations as objects and using the measurements collected from a spatial location over time to define the features. For example, in climate science, one of the goals is to group locations that experience similar climatic phenomenon over time. In this case, locations are treated as instances/objects and features are defined based on climate variables measured over time \cite{steinbach2002data}. Another scenario involves treating time points as objects and using measurements collected from all the spatial locations under consideration to define features. For example, in the application of discovering patterns of human brain activity from neuroimages, the goal is to identify the time points at which similar brain activity is observed in the brain. In this case, time points are treated as objects/instances and features are defined using the observed spatial map of activity \cite{liu2013decomposition}. There are also scenarios where events are treated as objects and features are defined based on the spatial and temporal information of events. For example, in the context of discovering crimes that are committed in close proximity in space and time, incidence of a crime is treated as an object and the location and time stamp of the crime are treated as features, in addition to other features such as nature of crime and number of victims involved \cite{tompson2015uk}. 
Hence, the coupling of spatial and temporal information in ST data introduces novel problems, challenges, and opportunities for STDM research, with a broad scope of application in several domains of scientific and commercial significance. 

There exists a vast literature on approaches for mining data that is purely spatial in nature, spanning multiple decades of research in spatial statistics \cite{cressie2015statistics}, spatial data mining \cite{shekhar2008spatial,shekhar2011identifying,aggarwal2015mining}, and spatial database management \cite{ester1997spatial,shashi2003spatial}. An extensive taxonomy of spatial data types and representations has been explored in the field of spatial data mining for improving the efficiency and effectiveness of data mining tasks such as clustering, prediction, anomaly detection, and pattern mining when dealing with spatial data \cite{shekhar2011identifying}. Another related area of research is time series data mining \cite{keogh2003need,liao2005clustering,esling2012time}, where approaches for mining useful information from time-series databases have been explored. Existing research in STDM includes foundational research in the statistics community \cite{cressie2015statistics}, e.g., research on spatio-temporal point processes \cite{diggle2013statistical}.  Approaches for handling spatial and temporal information have also been explored in the data mining literature for problems such as spatio-temporal clustering \cite{kisilevich2010spatio} and trajectory pattern mining \cite{giannotti2007trajectory}. 

There are a few recent surveys that have reviewed the literature on STDM in certain contexts from different perspectives. Articles by \cite{vatsavai2012spatiotemporal} and  \cite{chandola2015analyzing} discuss the computational issues for STDM algorithms in the era of `big-data' for application domains such as remote sensing, climate science, and social media analysis. The review by \cite{cheng2014spatiotemporal} covers STDM approaches for prediction, clustering, and visualization problems in several applications. An extensive survey of approaches for mining trajectory data, one of the many types of ST data, is presented in \cite{li2014spatiotemporal,zheng2015trajectory,mamoulis2009spatio}. A survey on STDM by \cite{shekhar2015spatiotemporal} provides a semantic categorization of ST data types and pattern families from a database-centric perspective.

Given the richness of problems and the variety of methods being explored in the rapidly advancing field of STDM, there is a need for developing an over-arching structure of research in STDM that highlights the similarities and differences of different problems and methods in diverse ST applications. This can enable the cross-pollination of ideas across disparate research areas and application domains, by making it possible to see how a solution developed for a certain problem in a particular domain (e.g., identifying patterns in climate data) can be useful for solving a different problem in another domain (e.g., understanding the working of the brain). This can also help in connecting the traditional data mining community with the challenges and opportunities in analyzing ST data, thus exposing some of the open questions and motivating future directions of research in STDM. 

This review paper on STDM problems and methods attempts to fulfill this need as follows.
% In contrast to existing surveys, this paper presents an comprehensive review of ST data types, problems, and methods, with an extensive breadth of algorithms from different research areas as well as application domains. The motivation for presenting this broad summary of STDM research is two-fold. 
First, it builds a foundation of ST data types and properties that can help in identifying the relevant problems and methods for any class of ST data encountered in real-world applications. In particular, we provide a broad taxonomy of the different types of ST data, different ways of defining and describing ST data instances, and different ways of computing similarity among ST data instances. Second, it presents a survey of STDM approaches for a number of commonly studied data mining problems such as clustering, predictive learning, frequent pattern mining, anomaly detection, change detection, and relationship mining. For every category of problems, we review the novel issues that arise in dealing with the unique properties of ST data types in classical data mining frameworks. 
% This survey can help in connecting the data mining community with the challenges and opportunities in analyzing spatio-temporal data that are becoming increasingly prevalent in different scientific and commercial disciplines. 
This paper can be used as a guide by data mining researchers and real-world practitioners working with ST data, to identify STDM formulations fit for their data and to make most effective use of STDM research in their problem.
In addition, by bridging the gap between classical data mining literature and the novel aspects of spatio-temporal data, this paper helps in  opening novel possibilities of future research. % \begin{comment}

The rest of the paper is organized as follows. In Section 2 we review the variety of application areas where analyzing ST data is important. In Section 3, we discuss the types and characteristics of ST data, and the different ways of defining instances and similarity measures using ST data types. Section 4 presents a survey of STDM methods developed for different types of ST data instances in the context of six major data mining problems, viz., clustering, predictive learning, frequent pattern mining, anomaly detection, change detection, and relationship mining. 
% We discuss future research directions in Section 5. 
Section 5 presents concluding remarks and discusses future research directions.

\section{Applications}

Large volumes of ST data are collected in several application domains such as social media, health-care, agriculture, transportation, and climate science. In this section, we briefly describe the different sources of ST data and the motivation for analyzing ST data in different application domains.

% structure for each application - Data, typical resolution, What it means, Some high level domain questions that are pursued.

\textbf{Climate Science:} 
Data pertaining to historic and current atmospheric and oceanic conditions (e.g., temperature, pressure, wind-flow, and humidity) is collected and studied in climate science \cite{karpatne2013earth}. In addition to observational data \footnote{https://www.ncdc.noaa.gov/cdo-web/datasets} collected from weather stations and reanalysis data that is gridded in space \cite{kistler2001ncep}, simulated data generated using climate models \cite{voldoire2013cnrm} is also studied in this domain. 
% For example, the World Climate Reserach Programme (WCRP) develops and distributes simulations of General Circulation Models (GCM) of climate variables such as sea surface temperature and pressure under the Coupled Model Intercomparison Project (CMIP) \cite{cmip}.
% This data is available for a duration of over three decades at a spatial resolution of $2.5^\circ \times 2.5^\circ $ and at a temporal resolution of four measurements per day (which is typically aggregated to weekly or monthly resolution for analysis). 
%Measurements capture prevailing atmospheric phenomenon such a temperature, pressure, and humidity at each square location (e.g., 2.5 Km) on the surface of the planet at regular intervals (e..g, daily or weekly). 
The purpose in studying this data is to discover relationships and patterns in climate science that advance our understanding of the Earth's system and help us better prepare for future adverse conditions by informing adaptation and mitigation actions in a timely manner.

\textbf{Neuroscience:}
Continuous neural activity captured using a variety of technologies such as Functional Magnetic Resonance Imaging (fMRI), Electroencephalogram (EEG), and Magnetoencephalography (MEG) is studied in neuroscience \cite{atluri2016brain}. Spatial resolution of neural activity measured using these technologies is quite different from another. For example, neural activity is measured from millions of locations in fMRI data, while it is only measured from tens of locations in the case of EEG data. Temporal resolution of the data collected using these technologies is also quite different. 
For example, fMRI typically measures activity for every two seconds, while the temporal resolution of EEG data is is typically 1 millisecond. 
% The subject from which the data is collected is either resting or working actively in a task while the measurements are being made. 
The purpose in studying this data is to understand the governing principles of the brain and thereby determine the disruptions to normal conditions that arise in the case of mental disorders \cite{atluri2015connectivity,atluri2013complex}. Discovering such disruptions can be useful for designing diagnostic procedures and in developing therapeutic procedures for patients. 

\textbf{Environmental Science:}
Studying the data pertaining to the quality of air, water, and environment is one of the objectives of environmental science. While air quality is measured based on the presence of pollutants such as particles, carbon monoxide, nitrogen dioxide, sulphur dioxide, ozone etc., water quality is measured based on factors such as dissolved oxygen, conductivity, turbidity, and pH. Air quality sensors are typically placed on streets or on top of buildings, and water quality sensors are placed in lakes, rivers, and streams. In addition to air and water quality, data pertaining to sound pollution is also collected. These environmental data sets are studied to detect changes in levels of pollution, identify the causal factors that contribute to pollution, and to design effective policies to reduce the different types of pollution \cite{thompson2014systems}.

%These sensors are placed all over the US in a non-gridded fashion. However, the data measured from the sensors can be summarized into a low resolution gridded data by taking mean of all the sensors present in the grid.

\textbf{Precision Agriculture:}
Multi-band high-resolution (ranging from 0.25m to 1m) areal or remote-sensing images of large farms are being collected at regular intervals (e.g., daily to weekly). One of the purposes of collecting and studying this data is to detect plant diseases \cite{mahlein2016plant} and understand the impact of several factors such as misapplication of fertilizer, compaction during planting and weeds on crop yield, as well as their inter-relationships. With the help of this knowledge, steps can be taken in future crop cycles to mitigate the risks due to the factors that adversely affect the crop yield. 

\textbf{Epidemiology/ Health care:}
Electronic health record data that is widely stored in hospitals provide demographic information pertaining to patients as well diagnosis made on patients at different time points. This dataset can be represented as a spatio-temporal dataset where each diagnosis has a spatial location and a time-point associated with it. 
%One can further convert this dataset into a gridded space time dataset by first defining the spatial and temporal resolution and then counting the number of patients that received a diagnosis within each space-time grid. 
One can construct such spatio-temporal instances for different types diseases such as cancers and diabetes, as well as for infectious diseases such as influenza. This data is studied to discover spatio-temporal patterns in different diseases \cite{matsubara2014funnel} and to study the spread of an epidemic. This data is also used in conjunction with environmental, climate science data sets to discover relationships between environmental factors and public health \cite{ryan2007comparison}. Discovery of such relationships will allow policy-makers to develop effective policies that will ensure the well-being of the population.

\textbf{Social media:}
Users of social media portals such as Twitter and Facebook post their experience at a given place and time. Each social media post captures the experience of a user at a given place and time. Using this data one can study collective user experience at a given place for a given time period \cite{tang2014mining}. One can also capture the spread of epidemics such as \emph{Influenza} or \emph{ebola} based on users' posts. More recently, there is also increased interest in studying the spread of social and political movements using social media data \cite{carney2016all}. In addition, events such as earthquakes, tsunamis, and fires can also be automatically detected from this data.

%\textbf{Air quality data}

%\textbf{Bike/Car share data}
%A number of cities in the US have both bike and car sharing programs where bikes and cars are made available to members of this service to commute from one place to another within the city. Because members commute from one place to another and leave the vehicle near their destination the number of vehicles in a given area keeps changing with time. This information can be summarized at a high level (relatively low-resolution) to represent the number of bikes available in a location in a given time. Although the number of bikes available in an region tends to be driven by the number of bike stands placed in the area, one can still study this data to explore the relative changes with time. 

\textbf{Traffic Dynamics:}
Large scale taxi pick-up/drop-off data is publicly available for several major cities across the world \cite{castro2013taxi}. This data contains information about each trip made by customers of the taxi service, including the time and location of pick-up and drop-off, and GPS locations for each second during the taxi ride. 
%One can represent this data as a gridded space time data by summarizing the number of pickups (and drop-offs) in a given grid region for a given interval.
This data can be used to understand how the population in a city moves spatially as a function of time and also the influence of extraneous factors such as traffic and weather. In addition, this data can also be studied to explore traffic dynamics based on the collective movement patterns of the taxis. This will enable transportation engineers to design effective policies to reduce traffic congestion. In addition, the behavior of taxi drivers can also be studied using this data so effective practices can be designed to detect abnormal behavior, increasing likelihood of finding new passengers, and taking optimal routes to arrive at a destination.
%This data can be useful to study where the demand for taxi's is higher and at what times. This can be used by taxi services to optimize their revenue. Information about taxis GPS traces can also be used to summarize the traffic at a given location for a given time. This will enable transportation engineers as well as taxi services to plan alternative routes for commuting. 

\textbf{Heliophysics:}
Heliophysics studies the events that occur in the Sun and their impact on the Solar System. The publicly available Heliophysics Events Knowledgebase \cite{hurlburt2010heliophysics} provides various observations that include solar events and their annotations on a daily basis. Examples of these events include Active region, Emerging flux, Filament, Flare, Sigmoid, and Sunspot. The time and the location of where these events were observed on the Sun are also provided in the knowledgebase. The spatial and temporal information along with the different observations are studied to discover patterns in the solar events \cite{pillai2012spatio}. The Heliophysics knowledgebase also enables the study of the impact of solar events and the Earth's climate system.

\textbf{Crime data:}
Law enforcement agencies store information about reported crimes in many cities and this information is made publicly available in the spirit of \emph{open-data} \cite{tompson2015uk}. This data typically has the type of crime (e.g., arson, assault, burglary, robbery, theft, and vandalism), as well as the time and location of the crime. Patterns in crime and the effect of law enforcement policies on the amount of crime in a region can be studied using this data with the goal of reducing crime.

\section{Data}

% Before we describe the problems and methods in the context of spatio-temporal data mining (STDM), it is important to discuss some of the key properties of spatio-temporal (ST) data, that differentiates it from classical data sets used in traditional data mining frameworks. These properties lie at the foundations of STDM methods, that attempt to comply with and leverage these properties in varying ways during the process of knowledge discovery. 
The presence of space and time introduces a rich diversity of ST data types and representations, which leads to multiple ways of formulating STDM problems and methods. In this section, we first describe some of the generic properties of ST data, and then describe the basic types of ST data available in different applications. Building on this discussion, we describe some of the common ways of defining and representing ST data instances, and generic methods for computing similarity among different types of ST instances.

\subsection{Properties}
% In ST data, every measurement is accompanied by the location and time at which it is collected. 
There are two generic properties of ST data that introduces challenges as well as opportunities for classical data mining algorithms, as described in the following.

%\begin{itemize}
%\item a
%\end{itemize}

\subsubsection{Auto-correlation} In domains involving ST data, the observations made at nearby locations and time stamps are not \emph{independent} but are correlated with each other. This auto-correlation in ST data sets results in a coherence of spatial observations (e.g., surface temperature values are consistent at nearby locations) and smoothness in temporal observations (e.g., changes in traffic activity occurs smoothly over time). As a result, classical data mining algorithms that assume independence among observations are not well-suited for ST applications, often resulting in poor performance with \emph{salt-and-pepper} errors \cite{jiang2015focal}. Further, standard evaluation schemes such as cross-validation may become invalid in the presence of ST data, because the test error rate can be contaminated by the training error rate when random sampling approaches are used to generate training and test sets that are correlated with each other. We also need novel ways of evaluating the predictions of STDM methods, because estimates of the location/time of an ST object (e.g., a crime event) may be useful even if they are not exact but in the close ST vicinity of ground-truth labels. Hence, there is a need to account  for the structure of auto-correlation among observations while analyzing ST data sets.

%The spatial and temporal autocorrelation makes the existing work in mining traditional datasets obsolete as they require data to follow $i.i.d$ property. Spatial data mining and time series mining approaches have been designed independently to tackle these challenges in isolation. Spatial data mining community has developed Spatial regression models to take spatial autocorrelation into account, where as ARIMA models were developed in econometrics community to take temporal autocorrelation into account. However, these developments are limited to a small set of problems (regression and forecasting, specifically) and much work is yet to be done to address state-of-the-art problems arising in several scientific domains.

\subsubsection{Heterogeneity} 
Another basic assumption that is made by classical data mining formulations is the {homogeneity} (or stationarity) of instances, which implies that every instance belongs to the same population and is thus \emph{identically distributed}. However, ST data sets can show heterogeneity (or non-stationarity) both in space and time in varying ways and levels. For example, satellite measurements of vegetation at a location on Earth show a cyclical pattern in time due to the presence of seasonal cycles. Hence, observations made in winter are differently distributed than the observations made in summer. There can also be inter-annual changes due to regime shifts in the Earth's climate, e.g., El Nino phase transitions, which can impact climate patterns to change on a global scale. As another example, different spatial regions of the brain perform different functions and hence show varying physiological responses to a stimuli. This heterogeneity in space and time requires the learning of different models for varying spatio-temporal regions.

%The nature of time series as well as the phenomenon captured in the time series can be different for different regions in space. For example, amount of precipitation in tropical and non-tropical regions is very different and so the properties of the time series at these locations are different (??). The change in vegetation, as seen in remote sensing data, due to fire is known to be different in different types of vegetation. Due to this heterogeneity in the nature of the underlying signal in the data, mining approaches have to be robust to be able to capture the target phenomenon.
%
%The characteristics of the time series collected at each location can change with time. For example, precipitation at a location in El Nino years can be different from that of La Nina years. Similarly, a phenomenon of interest can only be limited to a certain time period. Therefore data mining approaches need to be able to take this nonstationarity into account.

\subsection{Data Types}
There is a variety of ST data types that one can encounter in different real-world applications. They differ in the way space and time are used in the process of data collection and representation, and lead to different categories of STDM problem formulations. For this reason, it is important to establish the type of ST data available in a given application to make the most effective use of STDM methods. In the following, we describe four common categories of ST data types: (i) event data, which comprises of discrete events occurring at point locations and times (e.g., incidences of crime events in the city), (ii) trajectory data, where trajectories of moving bodies are being measured (e.g., the patrol route of a police surveillance car), (iii) point reference data \cite{cressie2015statistics}, where a continuous ST \emph{field} is being measured at moving ST reference sites (e.g., measurements of surface temperature collected using weather balloons), and (iv) raster data, where observations of an ST field is being collected at fixed cells in an ST grid (e.g., fMRI scans of brain activity). While the first two data types (events and trajectories) record observations of discrete events and objects, the next two data types (point reference and rasters) capture information of continuous or discrete ST fields. We discuss the basic properties of these four data types using illustrative examples from diverse applications. Indeed, if an ST data set is collected in a native data type that is different from the one we intend to use, in some cases, it is possible to convert from one ST data type to another, e.g., from point reference data to raster data. We briefly discuss some of the possible ways of converting an ST data type to other data types, for leveraging the STDM methods developed for those data types in a particular application.

% \vspace{-1ex}
\subsubsection{Event Data}
% In this data type, the basic focus is on measuring the behavior of discrete objects that have spatial and temporal components. There are two common types of ST objects: (a) \emph{events}, that have a fixed location and time point of occurrence, and (b) \emph{trajectories}, that denote the paths taken by moving bodies. Information about these ST objects are obtained using different collection technologies, and they in-turn give rise to different types of STDM problem formulations. In the following, we briefly describe the characteristics of these two ST objects using illustrative examples.
% \vspace{0.1in}

% {\bf $\bullet$ Event Data:}
% \vspace{0.05in}

\begin{wrapfigure}{r}{1.8in}
\vspace{-0.1in}
\centering
\subfigure[Events belonging to three types: $A$ (circles), $B$ (squares), and $C$ (triangles).]{\label{fig:events} \includegraphics[width=1.8in]{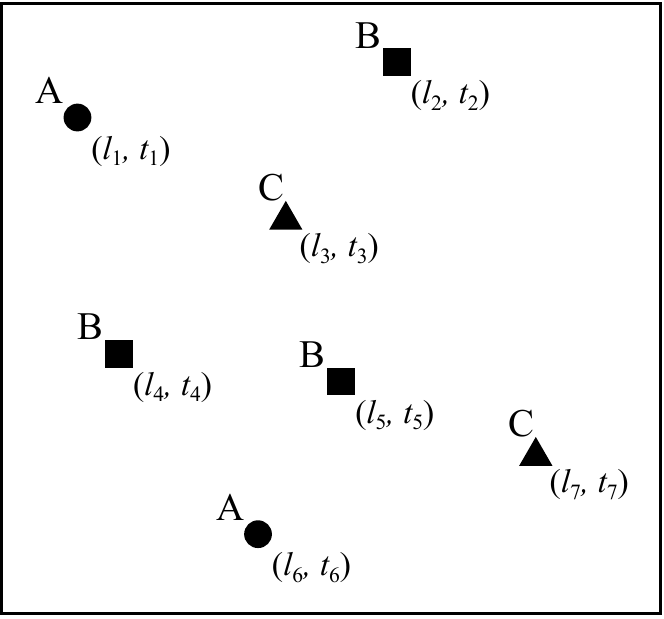}}
\quad
\subfigure[Trajectories of three moving bodies, $A$, $B$, and $C$.]{\label{fig:trajectories} \includegraphics[width=1.8in]{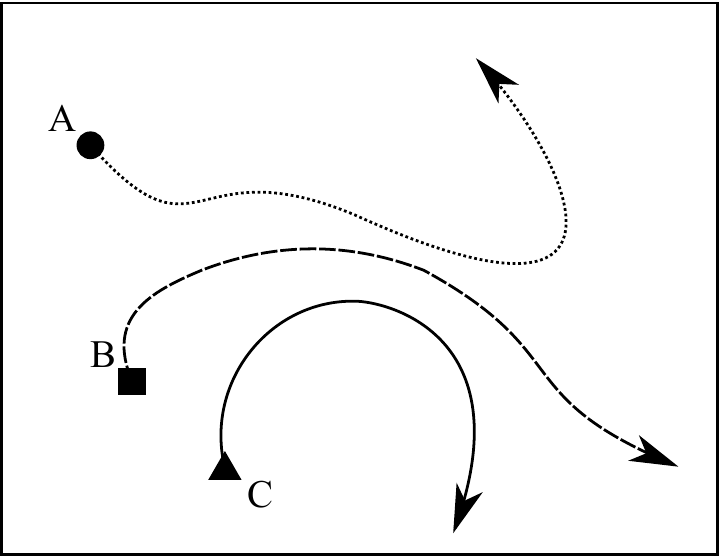}}
\caption{Illustrations of event data and trajectory data types.}
\label{fig:objects}
\vspace{-0.1in}
% \vspace{-0.25cm}
\end{wrapfigure}

An ST event can generally be characterized by a \emph{point} location and time, which denotes where and when the event occurred, respectively. For example, a crime event can be characterized by the location of the crime along with the time at which the crime activity occurred. Similarly, a disease outbreak can be represented using the location and time where the patient was first infected. A collection of ST point events is called as a spatial point pattern \cite{gatrell1996spatial} in the spatial statistics literature.
Figure \ref{fig:events} shows an example of a spatial point pattern in a two-dimensional Euclidean coordinate system, where $(l_i, t_i)$ denotes the location and time point of an event. Note that apart from the location and time information, every ST event may also contain non-ST variables, known as \emph{marked variables}, that provide additional information about every ST event. For example, events can belong to different types such as the type of disease or the nature of crime, which can be denoted by a categorical marked variable. In Figure \ref{fig:events}, the marked variable over events can take three categorical values: $A$, $B$, and $C$. ST events are quite common in real-world applications such as criminology (incidence of crime and related events), epidemiology (disease outbreak events), transportation (road accidents), Earth science (land cover change events like forest fires and insect disease), and social media (Twitter activity or Google search requests).

% \begin{figure}
% % \vspace{-0.1in}
% \centering
% \subfigure[Events belonging to three types: $A$ (circles), $B$ (squares), and $C$ (triangles).]{\label{fig:events} \includegraphics[width=1.8in]{figures/events-eps-converted-to.pdf}}
% \quad \quad
% \subfigure[Trajectories of three moving bodies, $A$, $B$, and $C$.]{\label{fig:trajectories} \includegraphics[width=1.8in]{figures/trajectories-eps-converted-to.pdf}}
% \caption{Illustrations of event data and trajectory data types.}
% \label{fig:objects}
% % \vspace{-0.1in}
% % \vspace{-0.25cm}
% \end{figure}

% \begin{figure}[t]
% \centering
% \subfigure[Events belonging to three types: $A$ (circles), $B$ (squares), and $C$ (triangles).]{\label{fig:events} \includegraphics[height=1.5in]{figures/events-eps-converted-to.pdf}}
% \quad
% \subfigure[Trajectories of three moving bodies, $A$, $B$, and $C$.]{\label{fig:trajectories} \includegraphics[height=1.5in]{figures/trajectories-eps-converted-to.pdf}}
% \caption{Illustrations of event data and trajectory data types.}
% \label{fig:objects}
% % \vspace{-0.25cm}
% \end{figure}

While the spatial nature of most ST events can be represented using a Euclidean coordinate system (where every dimension is equally important), sometimes it is more relevant to explore alternative representations of this data. For example, accidents on freeways can be considered as events occurring on a spatial road network, where the distance between any two events is measured not by their Euclidean distance but by the shortest distance of the road segments connecting the events. 
Further, events may not always be point objects in space but instead be characterized by other geometric shapes such as lines and polygons. For example, a forest fire event can be represented as a spatial polygon that delineates the extent of damage due to the fire.
Similarly, an event may not have an instantaneous time point of occurrence but instead be associated with a time period of appearance, denoting the birth and death of the event. For example, a music concert happening in the city can be represented using the start and end times of the event. While these simple extensions of ST events are quite common in real-world applications, most of the existing STDM methods are tailored for analyzing point ST events occurring in Euclidean spaces.

\subsubsection{Trajectory  Data}
Trajectories denote the paths traced by bodies moving in space over time. Some examples of trajectories include the route taken by a taxi from the pick-up to the drop-off location or the migration patterns of animals traveling for better access to food, water, and shelter \cite{farine2016both}. Trajectory data is commonly collected by mounting sensors on the moving bodies that periodically transmits information about the location of the body over time. Figure \ref{fig:trajectories} illustrates the trajectories of three moving bodies, $A$, $B$, and $C$. Apart from measuring the series of locations traversed by every moving body over time, trajectory data may also contain additional marked variables of the moving body, such as the heart-rate of a person running on a circuit track or the sequence of messages transmitted and received by a patrolling police car. Trajectories are common in applications such as transportation, ecology, and law enforcement.

%\vspace{0.1in}
%\vspace{-0.05in}

% {\bf $\bullet$ Point Reference Data:}
% \vspace{0.05in}

\subsubsection{Point Reference Data}
% The second type of ST data involves measuring spatio-temporal fields such as temperature, vegetation, and population. In contrast to object-based data, the focus of field-based data is to capture the behavior of ST fields over space and time.
% Depending on the technique used for measuring ST fields, field-based data can be broadly categorized into two basic types: (a) \emph{point reference} data, where a sample of reference points over space and time is used to measure a continuous ST field, and (b) \emph{gridded} data, where the field is measured over a fixed set of locations and time-points, arranged on a spatio-temporal grid. In the following, we briefly describe these two types of field-based ST data using illustrative examples from diverse disciplines.

% \begin{figure}
% % \vspace{-0.2in}
% \centering
% \subfigure[Reference points on time stamp 1.]{ \includegraphics[width=2.2in]{figures/point_ref1-eps-converted-to.pdf}}
% \subfigure[Reference points on time stamp 2.]{ \includegraphics[width=2.2in]{figures/point_ref3-eps-converted-to.pdf}}
% \vspace{-0.1in}
% \caption{An example of ST reference points (shown as dots) on two different time stamps. The colorbars show the distribution of the ST field on the two time stamps.}
% \label{fig:point_ref}
% % \vspace{-0.1in}
% \end{figure}

\begin{figure}[t]
\centering
\subfigure[Reference points on time stamp 1.]{ \includegraphics[width=0.48\textwidth]{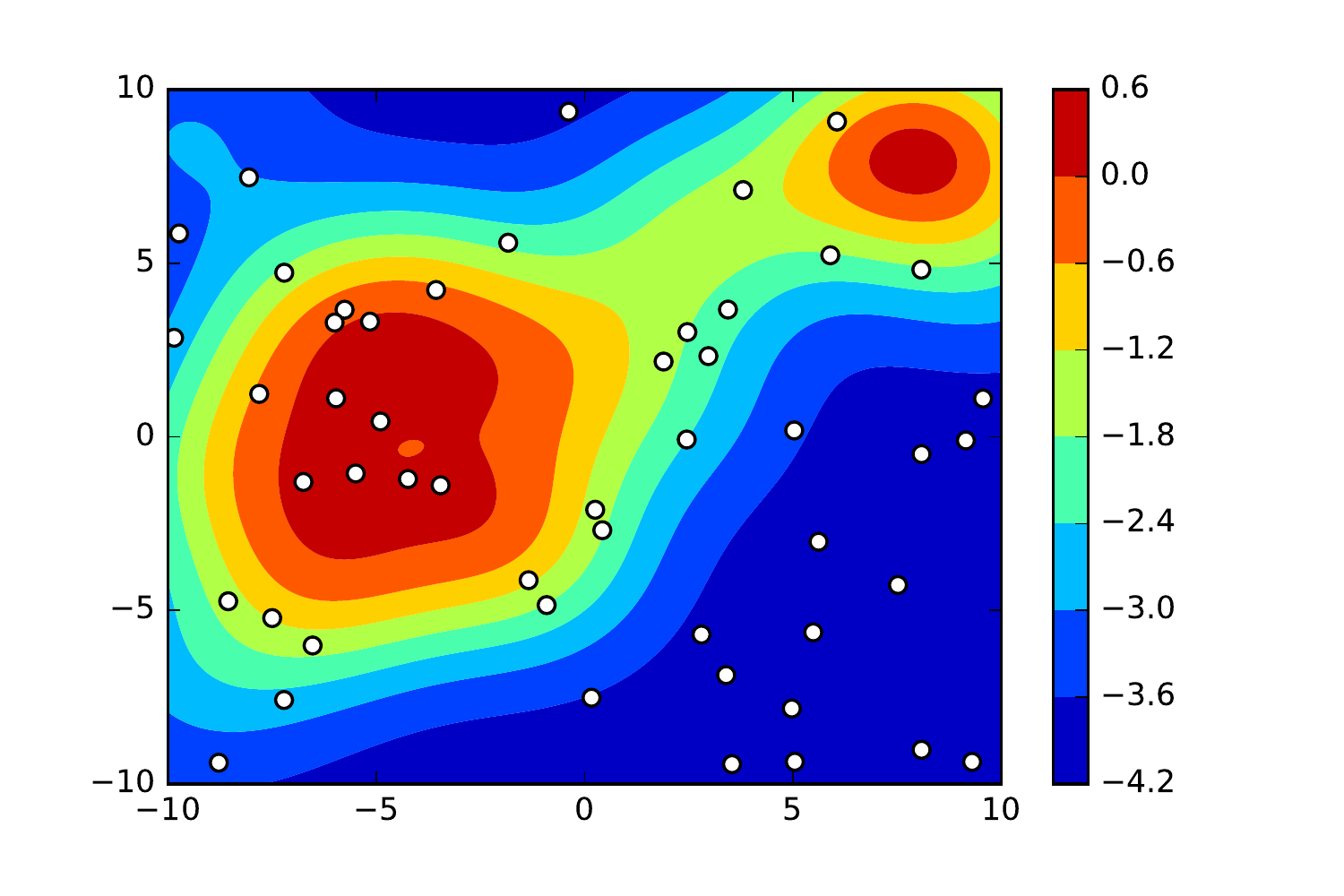}}
\subfigure[Reference points on time stamp 2.]{ \includegraphics[width=0.48\textwidth]{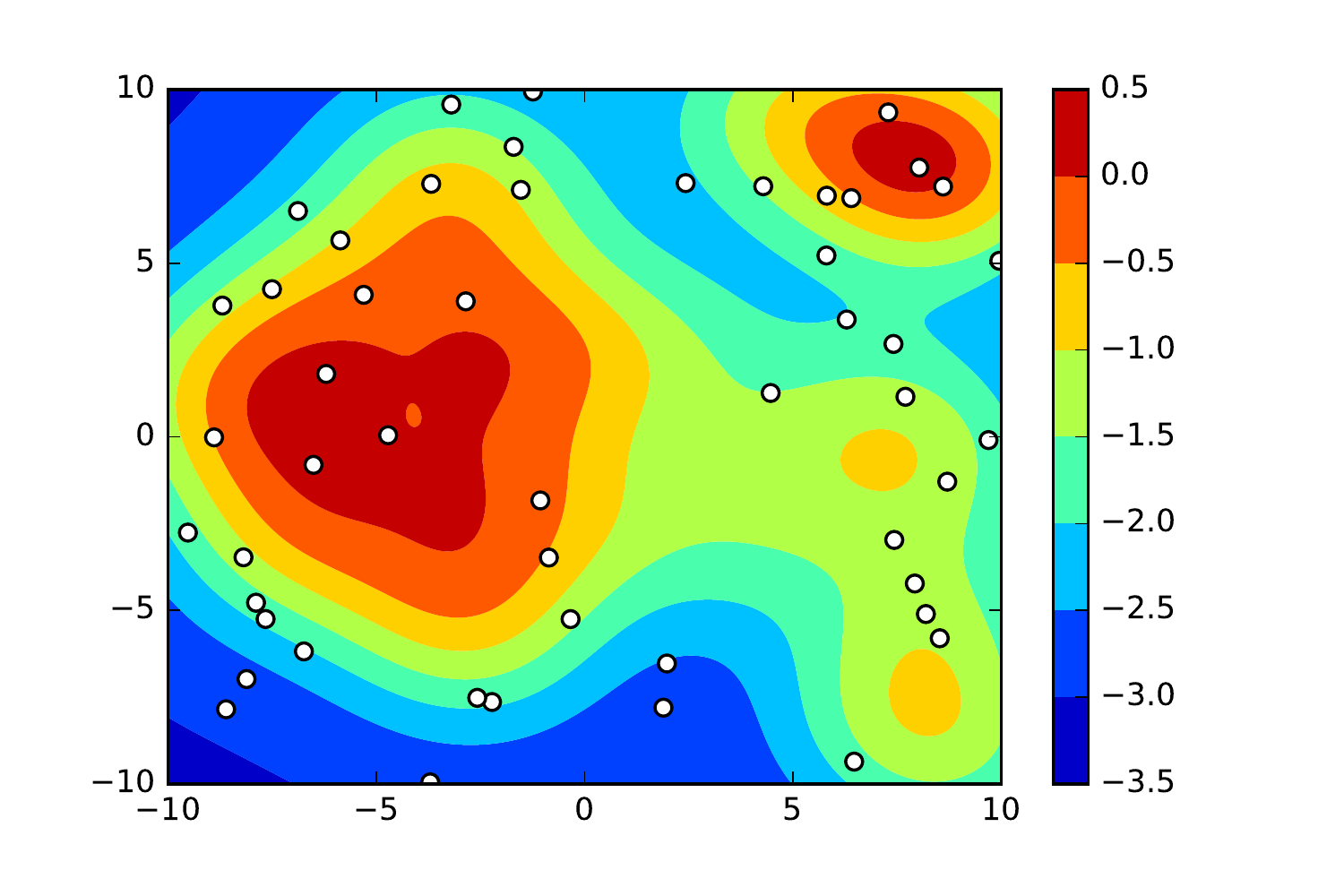}}
\caption{An example of ST reference points (shown as dots) on two different time stamps. The colorbars show the distribution of the ST field on the two time stamps.}
\label{fig:point_ref}
% \vspace{-0.25cm}
\end{figure}

A point reference data consists of measurements of a continuous ST field such as temperature, vegetation, or population over a set of moving \emph{reference} points in space and time. For example, meteorological variables such as temperature and humidity are commonly measured using weather balloons floating in space, which continuously record weather observations at point locations in space and time. Another example is that of buoy sensors used for measuring ocean variables such as sea surface temperature at locations that keep changing over time. In both these examples, a finite sample of ST reference points is used to represent the behavior of a continuous spatio-temporal field. Figure \ref{fig:point_ref} shows an example of the spatial distribution of a continuous ST field at two different time stamps, that are being measured at reference locations (shown as dots) on the two time stamps. The observations at these discrete reference points can be used to reconstruct the ST field at any arbitrary location and time using data-driven methods (e.g., smoothing techniques) or physics-based methods (e.g., meteorological reanalysis approaches \cite{saha2010ncep}).
Point reference data is also known as geostatistical data in the spatial statistics literature.

% \begin{wrapfigure}{r}{2.2in}
% \vspace{-0.2in}
% \centering
% \subfigure[Reference points on time stamp 1.]{ \includegraphics[width=2.2in]{figures/point_ref1-eps-converted-to.pdf}}
% \subfigure[Reference points on time stamp 2.]{ \includegraphics[width=2.2in]{figures/point_ref3-eps-converted-to.pdf}}
% \vspace{-0.1in}
% \caption{An example of ST reference points (shown as dots) on two different time stamps. The colorbars show the distribution of the ST field on the two time stamps.}
% \label{fig:point_ref}
% \vspace{-0.1in}
% \end{wrapfigure}

\subsubsection{Raster Data}
\label{sec:raster}

\begin{figure}[bt]
\centering
\subfigure[Regular space.]{\label{fig:s_regular} \includegraphics[width=0.26\textwidth]{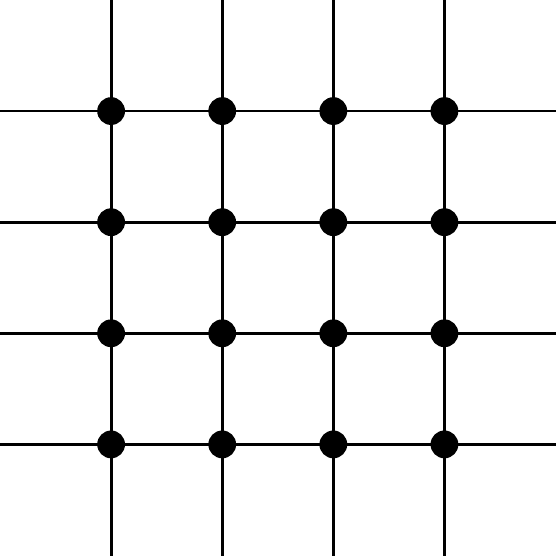}}
~~~~~~
\subfigure[Irregular space.]{\label{fig:s_irregular}  \includegraphics[width=0.26\textwidth]{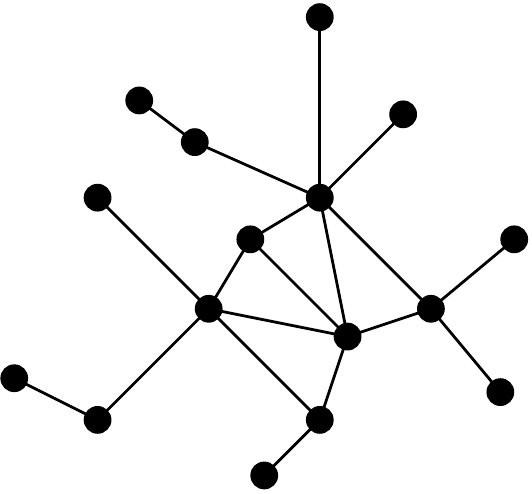}}
\\
\subfigure[Regular time.]{\label{fig:t_regular}  \includegraphics[width=0.35\textwidth]{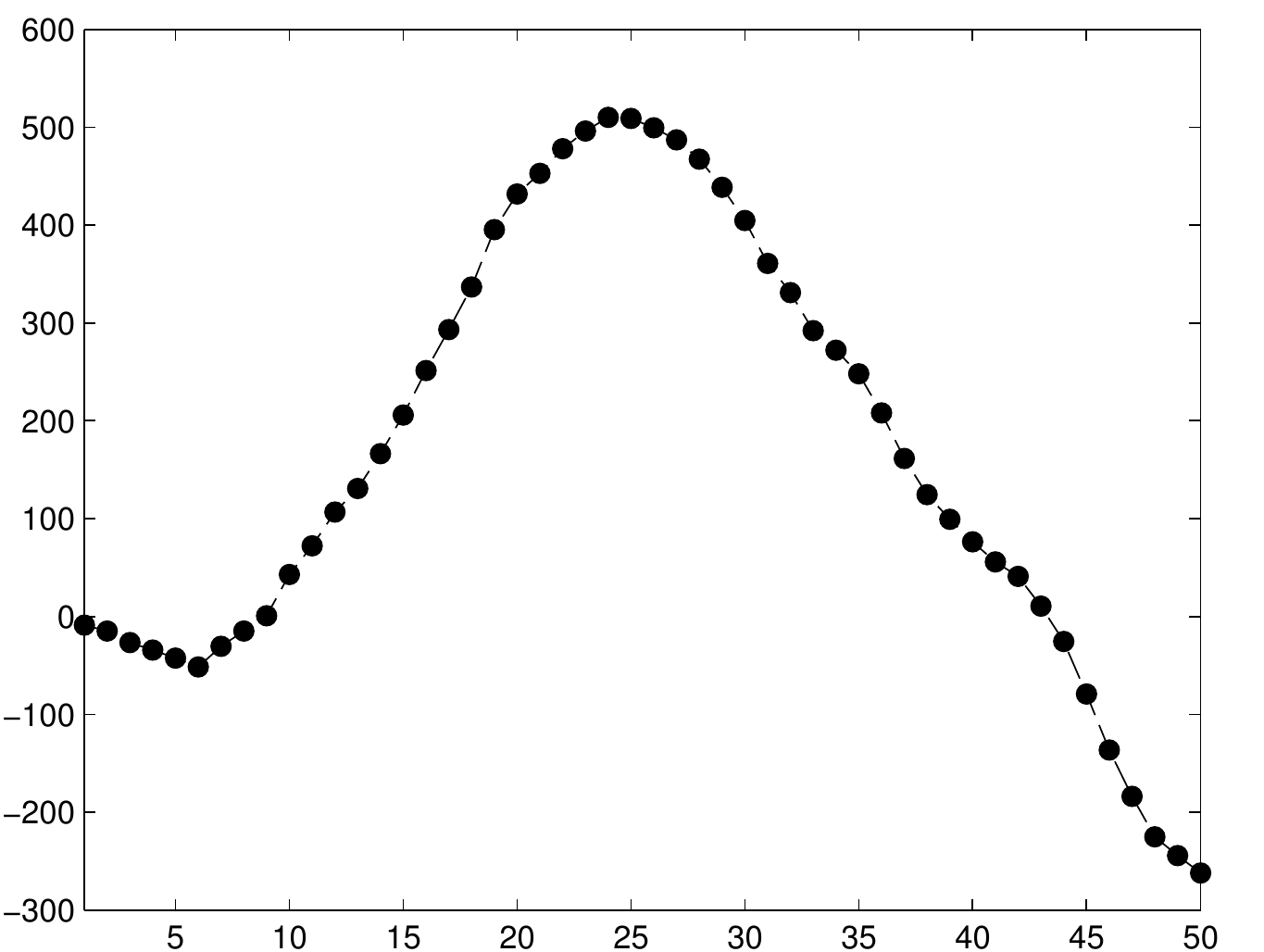}}
\quad
\subfigure[Irregular time.]{\label{fig:t_irregular}  \includegraphics[width=0.35\textwidth]{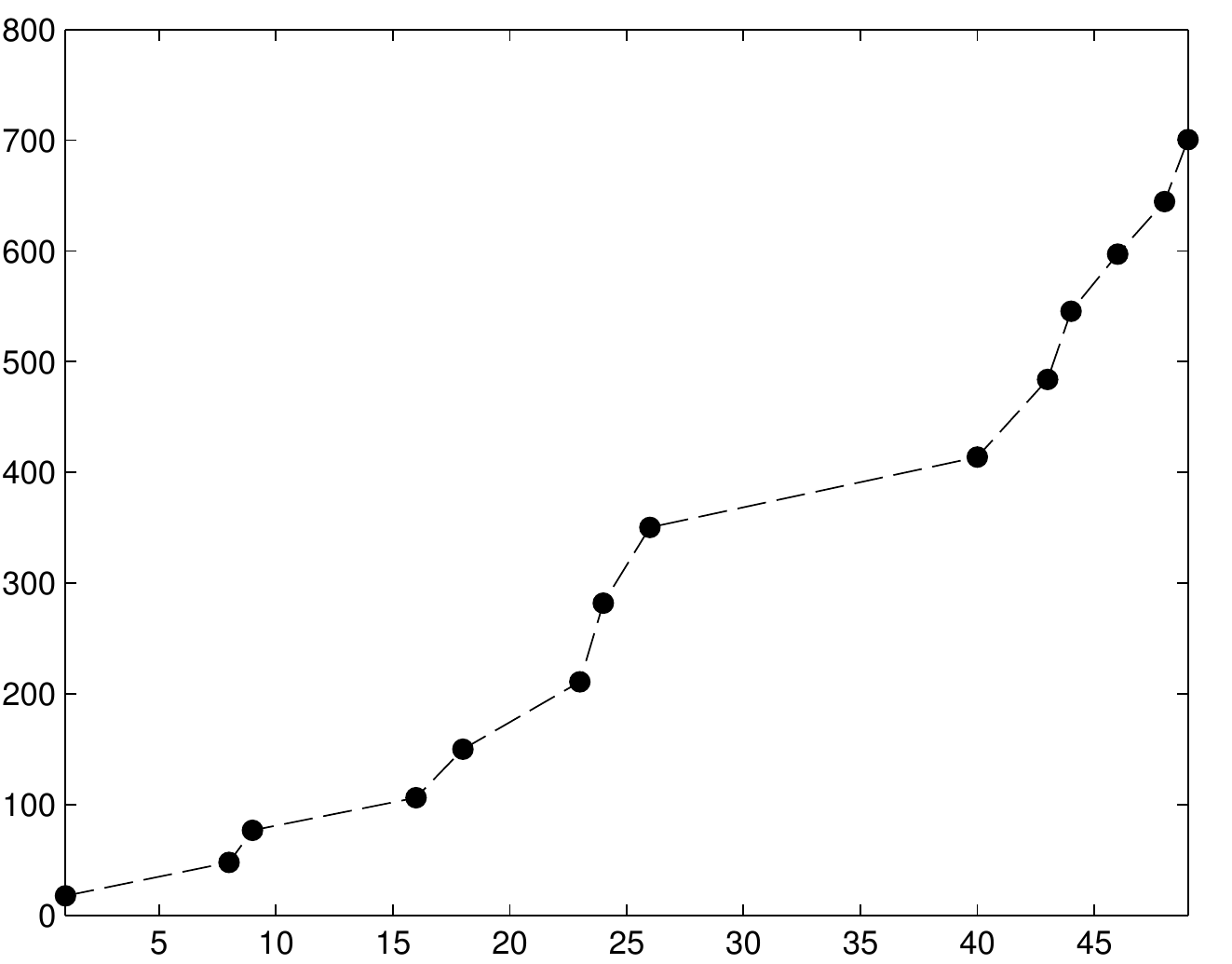}}
\caption{Different aspects of ST grids used for representing a raster data. The set of locations in the ST raster can either be located regularly or irregularly in space. The set of time stamps can also be either spaced regularly or irregularly.}
\label{fig:st_raster}
% \vspace{-0.25cm}
\end{figure}

In raster data, measurements of a continuous or discrete ST field are recorded at fixed locations in space and at fixed points in time. This is in contrast to point reference data where the ST reference sites may keep changing their location over time and collect recordings on different time stamps. To formally describe a raster, consider a set of fixed locations, $\mathcal{S} = \{s_1, s_2, \ldots, s_m\}$, either distributed regularly in space with constant distance between adjacent locations, e.g., pixels in an image (see Figure \ref{fig:s_regular}) or distributed in an irregular spatial pattern, e.g., ground-based sensor networks (see Figure \ref{fig:s_irregular}). 
For every location, we record observations on a fixed set of time stamps, $\mathcal{T} = \{t_1, t_2, \ldots, t_n\}$, which can again be regularly spaced with equal delays between consecutive measurements (see Figure \ref{fig:t_regular}) or irregularly spaced (see Figure \ref{fig:t_irregular}). 
It is the Cartesian product of $\mathcal{S}$ and $\mathcal{T}$ that results in the complete spatio-temporal grid, $\mathcal{S} \times \mathcal{T}$, where every vertex on the ST grid, $(s_i, t_j)$, has a distinct measurement.

% An example of irregularly distributed locations includes the sites placed by the U.S. Environmental Protection Agency to measure air pollutants near areas with high population.
% An example of irregularly spaced temporal observations is the gene expression images of Drosophila that are collected at non-uniform times. 

ST raster data is quite common in several real-world applications such as remote sensing, climate science, brain imaging, epidemiology, and demography. Some examples of ST raster data include measurements collected by ground-based sensors of ST fields such as air quality or weather information, geo-registered images of the Earth's surface collected by satellites at regular revisit times, and fMRI video sequences of brain activity. 
Note that while some examples of ST raster record observations at point vertices (e.g., measurements collected by a sensor network), others make aggregate measurements over the region at every grid cell. For example, demographic information is often collected at aggregate scales over political divisions such as cities, counties, districts, and states at annual scale. 
% In such a representation, the aggregate information over every political division of the map at every time stamp can be considered as an ST grid cell.
Another feature of an ST raster data is the resolution of the grid (both in space and in time) used for collecting measurements. 
In many applications, one commonly encounters ST raster data sets at varying resolutions of space and time, collected from different instruments or sensors. 
For example, satellite measurements of Earth's surface may be obtained via Landsat instruments at 30 meter spatial resolution every 16 days or via MODIS instruments at 500 meter spatial resolution on a daily scale. As another example, fMRI technology can be used to measure brain activity at each 1mm x 1mm x 1mm location, whereas EEG technology measures activity at a selected set of tens of locations. In problems involving ST rasters with varying resolutions, we often need to convert an ST raster from its native resolution to a finer or coarser resolution, so that a seamless analysis of all ST rasters can be performed at a common resolution. Interpolation techniques, also referred to as resampling methods in Geographic Information System (GIS) literature, are commonly used to convert a raster data to a finer resolution in space or time. Note that the computational requirements of STDM methods generally increase as we move to finer resolutions in space and time. A raster data can also be converted to a coarser resolution by aggregating over collections of ST cells. Aggregation generally helps in removing redundancies in the observations, especially when there is high spatial and temporal auto-correlation at finer resolutions. However, it is important to keep in mind that aggressive aggregation of data to coarser resolutions may result in loss of information about the ST field being measured.

% \vspace{0.1in}

\subsubsection{Converting Data Types}
Even if an ST data is naturally collected in a particular data type in a certain application, it is possible to \emph{transform} it to a different ST type so that the relevant family of STDM tools are used for their analyses. We provide some examples of inter-conversion among ST data types in the following. An event data type can be converted to a raster data type by aggregating the counts of events at every cell of an ST grid. For example, crime events can be counted at the levels of counties in a city at an hourly scale, thus producing an ST raster of crime occurrences. In some cases, a raster data can also be converted to ST events by using special algorithms for event extraction, e.g., techniques to find ST regions with abnormal activity. As an example, ecosystem disturbance events such as forest fires can be extracted from geo-registered satellite images of vegetation cover \cite{mithal2011monitoring}.
Another common type of conversion among ST data types is between point reference data and raster data.
Observations at ST reference points can be transformed into an ST raster format by interpolating or aggregating over an ST grid. Raster data can also be converted to a point reference data where every vertex of the ST grid is viewed as an ST reference point.

\subsection{Data Instances}

\begin{wrapfigure}{r}{2.2in}
\vspace{-0.15in}
\centering
\includegraphics[width=2.2in]{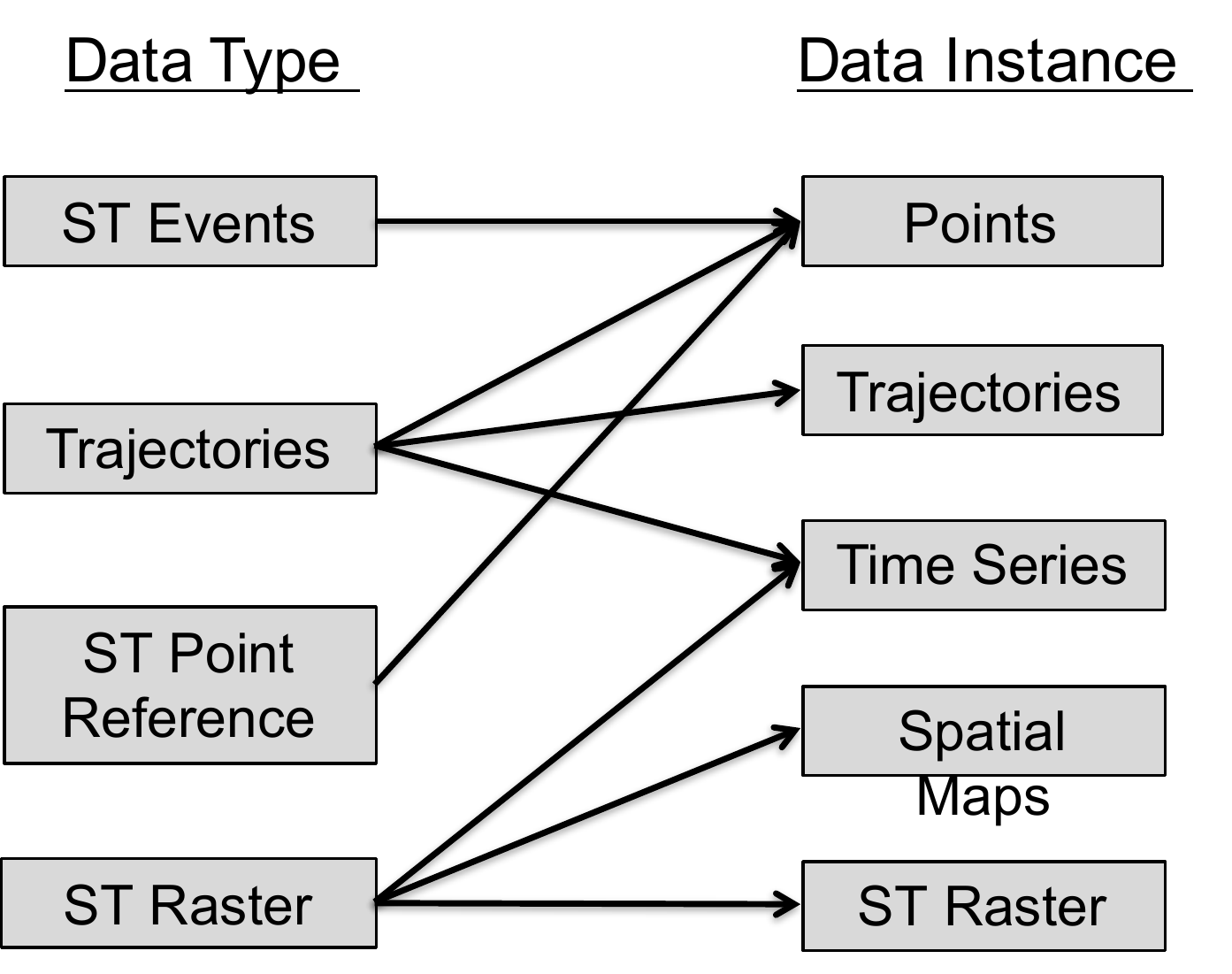}
\vspace{-0.1in}
\caption{A mapping showing the different categories of ST data instances that can be constructed from ST data types.}
\label{fig:type_instance}
\vspace{-0.1in}
\end{wrapfigure}

The basic unit of data that a data mining algorithm operates upon is called a \emph{data instance}.
In classical data mining settings, a data instance is unambiguously represented as a set of observed features with optional supervised labels. 
However, in the context of ST data, there are multiple ways of defining instances for a given data type, 
each resulting in a different STDM formulation. In this section, we review five common categories of ST instances that one encounters in STDM problems, namely, points, trajectories, time series, spatial maps, and ST rasters. These data instances form the canonical building blocks of analysis for a broad range of problems and methods in STDM that we will discuss in Section \ref{sec:problems}. 

Figure \ref{fig:type_instance}  summarizes the different types of data instances that can be constructed for each ST data type. While ST events can be naturally represented as point instances, a trajectory data type can either be represented as a collection of point instances (ordered list of locations visited by the moving object), a trajectory instance, or as a time-series of spatial identifiers (e.g., location coordinates). A ST reference data type comprises of point instances, where every instance is a \emph{reference} point of the ST field in space and time. There are three different ways of constructing instances for ST raster data. First, we can define an instance to be the set of measurements at any location, $s_i$, represented as a time series, $T_i$. Second, we can also define an instance to be the set of measurements at any time stamp, $t_j$, represented as a spatial map, $S_j$. A third approach is to consider the entire ST raster (collection of observations over the entire ST grid) as a single data instance.
It is the plurality of ways for defining instances for every ST data types that results in multiple STDM problems and methods. The choice of the right approach for constructing ST instances from a ST data type depends on the nature of the question being investigated, and the family of available STDM methods that can be used.

In the following, we introduce the five categories of ST data instances and discuss some of the common questions that can be investigated for each category. Since ST data instances can be summarized using a number of ways that go beyond standard dimensionality reduction techniques such as principal component analysis, we also describe some of the common ways of representing and summarizing the features for every instance category.

% Note that some of the ST data instances can be derived from multiple types of ST data, e.g., points can either be ST events (given event data) or ST reference sites  (given point reference data). 

% For example, a time-series can be represented using a dictionary of frequent temporal sub-sequences known as ``temporal motifs.'' 

% Hence, a basic understanding of ST data instances will be instrumental in building a taxonomy of existing research in STDM, so that it is easy to identify the type of data instance a certain problem formulation caters towards. This can also help in identifying open problems for data instances that have been less explored and there is an opportunity for novel research.

% \begin{figure}
% % \vspace{-0.15in}
% \centering
% \includegraphics[width=2in]{figures/STinstance.pdf}
% % \vspace{-0.1in}
% \caption{A mapping showing the different categories of ST data instances that can be constructed from ST data types.}
% \label{fig:type_instance}
% % \vspace{-0.1in}
% \end{figure}

% A unique quality of ST raster data is that the spatial and temporal components of an ST raster data can be represented using two different views as follows. First, the set of measurements at any location, $s_i$, can be represented as a time series, $T_i$. Second, the measurements at any time stamp, $t_j$, can be considered as a spatial map, $S_j$. 

\subsubsection{Points}

An ST point can be represented as a tuple containing the spatial and temporal information of a discrete observation, as well as any additional variables associated with the observation.
ST points are frequently used as data instances in STDM analyses involving event data, e.g., point events such as the occurrence of crimes at certain locations and times can be treated as basic instances to group similar instances or to find anomalous instances. ST points are also used as data instances when dealing with point reference data, where the measurements at ST reference points are used as instances to estimate the ST field at unseen instances. Additionally, a trajectory can also be viewed as an ordered collection of ST points, which are the locations visited by the moving object.
% A basic way to represent a point can be represented as a tuple containing the location, time, and additional variables at every point. 

Some of the common questions that can be asked using ST points as data instances include: \emph{how are ST points clustered in space and time? What are the frequently occurring patterns of ST points? Can we identify ST points that do not follow the general behavior of other ST points?
Can we estimate a target variable of interest at an ST point that has not been seen during training?}

A collection of ST points can be summarized by measures that capture the strength of interaction (or auto-correlation) among the points. For example, the Ripley's $K$ function \cite{dixon2002ripley} is a commonly used statistic for describing the amount of attraction or repulsion among spatial locations beyond its expected value. Spatio-temporal extensions of the Ripley's $K$ function have also been explored to measure the strength of interaction among ST point events \cite{lynch2008spatiotemporal}. The strength of auto-correlation among ST reference points can also be measured using the Moran's I function \cite{li2007beyond}, local Moran's I \cite{anselin1995local}, and its ST extension \cite{hardisty2010analysing}.

\subsubsection{Trajectories}

Trajectories are a different class of data instances that can be used in STDM analyses involving moving bodies. Trajectories can be represented as multi-dimensional sequences that contain a temporally ordered list of locations visited by the moving object, along with any other information recorded by the object.

Some of the common questions that can be asked using trajectories as data instances include: \emph{can we cluster a collection of trajectories into a small set of representative groups? Are there frequent sequences of locations within the trajectories that are traversed by multiple moving bodies?}

A common approach for representing and extracting features from trajectories
is by the use of generative models \cite{gaffney1999trajectory}, where a parametric model is used to approximate the behavior of every trajectory. The learned parameters can then used as succinct representations of the trajectories. A trajectory can also be represented using the frequent sub-sequences of locations that are visited by the moving body. Techniques for identifying frequent trajectory patterns are discussed in detail later in Section \ref{sec:trajectory_pattern}. Distributed and efficient  indexing structures for answering trajectory-related similarity queries have been developed in \cite{zeinalipour2006distributed,al2007dimensionality}. Apart from these methods, different schemes such as semantic trajectories \cite{bogorny2014constant,li2017semantic}, symbolic trajectories \cite{guting2015symbolic}, and spatio-textual trajectories \cite{damiani2016spatial} are also recently being explored for representing trajectory data.

\subsubsection{ Time Series}

Time series can be used as data instances in two different scenarios involving ST data. First, given an ST raster data, we can consider the set of observations at every spatial cell in the ST grid as a time series that can be used as a data instance in an STDM analysis.
Second, a trajectory data can also be treated as a multi-dimensional time-series data, where the multiple dimensions correspond to the spatial identifiers (e.g., location coordinates) traversed by the moving objects over time, and any other variables recorded by the moving object in the course of its trajectory. While representing spatial identifiers as multiple (and independent) dimensions of a time-series may not preserve the spatial context among the identifiers, 
it opens up the vast literature on time series data mining that can be used for analyzing trajectories in novel ways, as will be described in Section \ref{sec:problems}.

Some of the common questions that can be asked using time series as data instances include: \emph{can we identify groups of time-series that show similar temporal activity and are located nearby in space? Are there some temporal patterns that commonly repeat in a number of time-series? Can we identify time stamps where the time-series deviate from their normal behavior for a short period of time? Can we discover time stamps where the time-series show a change in its profile? Can we use time-series as input features to predict a target variable? Can we predict the value of a time-series at a future time stamp using its historical values? Can we find distant groups of spatially contiguous time-series that are related to each other?}

A number of approaches exist for extracting useful features from a collection of time-series. This include methods that can identify temporally frequent sub-sequences that occur in a majority of time-series (e.g., temporal motifs \cite{mueen2014time}) or contain discriminatory information about a particular time-series class (e.g., shapelets \cite{ye2009time,hills2014classification}). A review of techniques for identifying such frequent patterns in time-series is presented in \cite{fu2011review} and \cite{esling2012time}.

\subsubsection{Spatial Maps}
An ST raster data can be viewed as a collection of spatial maps observed at every time stamp, which can also be used as data instance in analyses involving ST rasters.

Some of the common questions that can be asked using spatial maps as data instances in STDM analyses include: \emph{can we cluster the spatial maps to find groups of time stamps showing similar spatial activity? Can we identify spatial patterns that are observed in a number of spatial maps? Can we use spatial maps as input variables to predict a target variable? Can we predict the value at a certain location using observed values at other locations in the map?}

A common approach for extracting features among spatial maps is using image segmentation techniques \cite{haralick1985image}. The presence or absence of different types of image segments can then be used as features to represent spatial maps. 
% Another approach for representing spatial maps is using a catalog of frequent spatial patterns, termed as spatial motifs \cite{chen2015introducing}. Techniques for extracting spatial motifs from collections of spatial maps will be described in detail later in Section \ref{sec:frequent}.

\subsubsection{ST Rasters}
\label{sec:raster_instance}
An ST raster data in its entirety, with measurements spanning the entire set of locations and time stamps, can also be treated as individual data instances in STDM analyses. 

Some of the common questions that can be asked using ST rasters as data instances include: \emph{can we cluster ST rasters into groups that show similar behavior in space and time? Can we find frequent spatio-temporal behavior that occurs in a number of ST raster data sets? Can we find ST rasters that show distinctly different behavior than other ST rasters? Can we find timestamps where the behavior of an ST raster changes over time? Can we use ST rasters as input features for predicting a target variable of interest? Can we predict the value at a certain location and time stamp in the ST grid using observed values at other locations and time stamps? Can we find subsets of locations in the ST grid that show interesting relationships in their temporal activity?}

A basic approach for representing ST rasters is using $N$-way arrays also called as \emph{tensors}. In a tensor representation of an ST raster data, some dimensions are used to represent the set of locations while the remaining dimension is used to represent the set of time stamps available in the ST grid. For example, precipitation data is represented as a 3-dimensional array where the first two dimensions capture $2D$ space and the third dimension captures time. Similarly, fMRI data is represented as a 4-dimensional array where the first three dimensions capture $3D$ space and the third dimension captures time. A tensor representation of an ST raster data can then be summarized using space-time subspaces that have similar values, which are the equivalent of image segmentation in ST domains. 
% Such space-time subspaces are expected to capture short-lived phenomenon that are restricted to a small number of locations. Examples of such phenomenon are floods that occur for a few weeks in small areas around water bodies whose impact is changing mildly over its life time. 
Existing image segmentation techniques that have been extended to work with videos to discover moving objects are relevant to address this problem \cite{haritaoglu2000w,prati2003detecting}. 
% The performance of these techniques needs to be evaluated for discovering space-time subspaces. 

% Tensors even though are a natural representation that capture the entire raw data, it is often the case that only small amount of data is useful in many problems such as classification and clustering. Moreover, features constructed from subsets of data tend to be informative than the raw data itself. The type of features vary with the application domains and the nature of the problem. One may be able to determine relevant features subjectively with the aid of domain scientists. Once the relevant features are determined the next task is to discover them from the underlying data.

One way to summarize or extract features from ST rasters is by using network-based representations, where the nodes correspond to the locations and the edges denote the similarity among the time-series at locations \cite{atluri2016brain,feldhoff2015complex}. Techniques for computing similarity among time-series are discussed in detail later in Section \ref{sec:ts_sim}. The topological properties of nodes such as degree and different variants of centrality measures can then be used to characterize the `role' and `influence' of locations. Such properties have been found to be useful in characterizing nodes in different types of networks such as social, biological, and transportation networks. For example, the use of network-based properties such as coherence among a set of locations \cite{de2007classification,sui2009ica} or relationship between distant locations \cite{lynall2010functional,pettersson2011dysconnectivity} has been explored in previous studies.

% Here we will discuss some of the classification approaches that have been proposed for classifying the stimulus of ST rasters obtained from fMRI brain scans. 
% One approach is to only use those voxels (ST grid cells) that are found to be highly responsive to a stimulus \cite{ku2008comparison}. Another approach is to use 'Regions of Interest' (ROIs) that are defined using external data sources or domain knowledge (e.g., anatomical regions in brain fMRI) and to aggregate the activation scores from all the voxels within selected ROIs \cite{etzel2009introduction}. Approaches that directly determine the relevant features as part of the classification model have also been explored \cite{ryali2010sparse,rasmussen2012model}. These approaches typically aggregate the temporal dimension and so they are limited in their ability to make use of any variability along this dimension. 

% $\bullet$ {\color{red} Discuss different ways of constructing networks and dynamic networks}

% $\bullet$ {\color{red} Different types of networks - nodes are locations, locations+time, edges are directional and non-directional, edges can be based on univariate or multivariate space time variables - Nagiza's work}

% $\bullet$ {\color{red} Kolda's work in time series data.}

% $\bullet$ {\color{red} Recent tensors work in ST data.}

% $\bullet$ {\color{red} Discuss examples of fMRI, Climate where different types of features are used. can be a page long discussion.}

\subsection{Similarity Among Instances}
\label{sec:sim_instances}
Determining similarity (or dissimilarity) between data instances is key to many data mining problems such as clustering, classification, pattern discovery, and relationship  mining. Similarity measures have been extensively studied for data instances in traditional data mining settings, that are able to deal with varying types and formats of attributes. However, in the presence of space and time, there is a variety of ways we can define similarity among ST data instances such as points, trajectories, time-series, spatial maps, and ST rasters.
Our ability to capture similarities among ST data instances using proximity measures underpins the effectiveness of STDM methods that build upon them.
In the following, we describe some of the common ways for computing similarity among different types of ST data instances.

\subsubsection{Point Similarity}

Two points are considered close if they lie within the ST neighborhoods of each other. The ST neighborhood of a point can be defined using a fixed distance threshold in space and time, e.g., within 1 km radius and 1 hour time difference. Alternatively, the ST neighborhood of every point can also be defined in terms of a fixed number, $k$, of closest points. The choice of the right notion of locality depends on the application context and can be decided by the domain analyst.

\subsubsection{Trajectory Similarity}
\label{sec:trajectory_sim}
Similarity among trajectories is often measured in terms of the co-location frequency, which is the number of times two moving bodies appear spatially close to one another. Other approaches for measuring similarity among trajectories include subsequence similarity metrics such as the length of the longest common subsequence, Fr\'{e}chet distance, dynamic time warping (DTW), and edit distance \cite{toohey2015trajectory}. Trajectory similarity can also be computed using feature-based representations such as the frequent trajectory patterns extracted from the data.

\subsubsection{Time Series Similarity}
\label{sec:ts_sim}
If we consider every time-series as a 1D-array of observations, the similarity among two time-series can be simply computed using  proximity measures such as the Euclidean distance and the correlation strength, that consider a one-to-one correspondence between the elements of the two arrays. However, sometimes it is the case that two similar time-series are not exactly aligned with one another but show the same pattern of activity over time. Measures such as dynamic time warning (DTW) \cite{keogh2005exact} and Fr\'{e}chet distance \cite{alt1995computing} are able to capture such forms of similarity among time-series. We can also compute the similarity among two time-series based on their closeness in time-series features such as temporal motifs and shapelets. When it is expected to observe a certain delay or time lag between the observations of two time series, a common approach is to translate one of the time series with a range of candidate values of time lag and then choose the time lag that provides maximum similarity (e.g., highest absolute correlation).

While most measures of time-series similarity consider the entire time duration into account, it is possible that the similarity structure among time-series is subject to variation over time. For example, when a subject's mental activity is switching between planning their day (i.e., an executive task) and reminding themselves of a recent meeting (i.e., a memory task), the similarity structure among the time-series at brain regions could be different. In such cases, a desired pattern of similarity among time-series may only be exhibited for short periods of time, which need to be determined from the data. An example of an approach that simultaneously identifies the relevant time windows for computing time-series similarity and uses this metric to cluster the time-series can be found in \cite{atluri2014discovering}.

\subsubsection{Spatial Map Similarity}
\label{sec:map_sim}

Two spatial maps can be considered similar if they show similar values at corresponding locations, which can be captured using standard proximity measures such as Euclidean distance. However, spatial maps can often suffer from small misalignments due to geo-registration errors, which can result in misleading distance metrics. Further, it is often more useful to compute similarity over smaller sub-regions in the map that contain foreground objects than considering the similarity over the entire map. The Earth's mover distance (EMD) \cite{rubner2000earth} is one such metric that is robust to changes in the alignment of images, which is based on the minimal cost that must be paid to transform one image into the other. Spatial map similarity can also be computed on the basis of features such as image segments extracted from the data.

\subsubsection{ST Raster Similarity}
\label{sec:raster_sim}

Network-based representations of ST rasters can be used to assess if two rasters are similar or not. For example, link and node similarity scores \cite{berg2006cross,li2009comparing} can be used to compute the similarity among the corresponding links and nodes of ST rasters. ST raster similarity can also be computed on the basis of features extracted from their network representations \cite{soundarajan2013network}.

\section{Problems and Methods}
\label{sec:problems}
\begin{comment}
{\color{blue} Old Text:\\

Traditionally in data mining several key tasks such as classification, clustering, pattern mining, and anomaly detection have been studied in the context of relational data (where each object is characterized by a set of attributes). These tasks have emerged based on the fundamental questions that are being studied over time in different domains. For example, pattern mining emerged out of the need to find frequently purchased groups of items in market-basket data. Later on with the increased availability of time series data and the need to mine such data, the traditional data mining tasks have been extended and new tasks such as segmentation and motif discovery have also been defined. Similarly, spatial data has also been studied predominantly by extending clustering framework to take spatial information into account and by generalizing pattern mining to co-occurrence mining where the objective is to find objects that frequently co-occur within a given spatial proximity.

As discussed above the time series and spatial data mining communities have largely extended traditional data mining tasks to address questions emerging in the relevant domains. However, in the context of spatio-temporal data the nature of questions that can be posed are very different from those that are posed in relational, time series or spatial datasets.
}
\end{comment}

\subsection{Clustering}

Clustering refers to the grouping of instances in a data set that share similar feature values. 
When clustering ST data, novel challenges arise due to the spatial and temporal aspects of different types of ST instances. For example, clustering locations based on their time series of observations in an ST raster data has to ensure that the discovered clusters are spatially contiguous. Ignoring this spatial information can lead to fragmented clusters of locations that are difficult to interpret and have salt-and-pepper errors. Since many clustering algorithms use similarity measures for identifying groups of similar instances, techniques for computing similarity among ST instances (described in Section \ref{sec:sim_instances}) will come in handy while clustering ST data. In the following, we describe some of the common methods for clustering different types of ST data instances.

\subsubsection{Clustering Points}

% A number of  approaches for clustering data points that only has spatial attributes and no temporal attributes have been studied. They include the seminal work of spatial scan statistics to discover regions of arbitrary shapes and sizes are shown to be useful for studying infant mortality \cite{kulldorff1997spatial}, ring-shaped hot-spot detection for crime data \cite{eftelioglu2014ring}, and spatial network activity summarization that enforces constraints that are in the form of a `spatial network' (e.g., road network) for discovering road segments with high pedestrain mortality \cite{zheng2012mining}.
There are two different objectives that are  of interest when clustering ST points. One objective involves finding clusters that have an unusually high density of ST points, also termed as \emph{hot-spots}. This can be used for finding outbreaks of diseases or social movements, where there is a dense conglomeration of events both in space and time
 \cite{gomide2011dengue,kulldorff2005space}. This problem is also referred to as `event detection' in the literature on mining social media data \cite{feng2015streamcube}. The problem of finding hot-spots has been studied in the spatial statistics literature using methods such as the spatial scan statistic \cite{kulldorff1997spatial}, which explores all possible circular shaped regions of different sizes to determine regions where the incidence of points is significantly higher than expected. Generalization of the scan statistic for ST data, termed as space-time scan statistics, have been explored  for studying disease outbreaks, crime hot-spots, and events in twitter data \cite{kulldorff2001prospective,kulldorff2005space,cheng2014event}.  While these early works provide a promising framework, open questions pertaining to shapes of regions \cite{takahashi2008flexibly,eftelioglu2016ring},  background distribution \cite{tango2011space}, and speed of search \cite{agarwal2006spatial} are being investigated. Approaches for clustering spatial objects such as CLARANS \cite{ng2002clarans} can also be extended in the ST domain for finding groups of objects that are related in both space and time.
 
 The second objective involves finding clusters of ST points that also have similar non-ST attributes. For example, given a collection of crime events, we may be interested in finding regions in space and time with similar crime activities. This clustering objective has been studied in the context of  crime data \cite{eftelioglu2014ring}, twitter data \cite{abdelhaq2013eventweet,chierichetti2014event,ihler2006adaptive,walther2013geo,weng2011event}, geo-tagged photos \cite{zheng2012mining}, traffic accidents \cite{zheng2012mining}, and epidemiological data \cite{glatman2016near}. A number of techniques for clustering ST points are based on the DBSCAN algorithm \cite{ester1996density}, which is a widely used method for finding arbitrarily shaped clusters of spatial points based on the density of points. ST-DBSCAN \cite{birant2007st} is one of the popular extensions of DBSCAN that defines two separate distances between ST points: one that captures spatial attributes and another that captures temporal and non-ST attributes. Computing distance based on spatial, temporal and non-ST attributes separately and having separate thresholds for them provides the user a flexibility to determine the desired spatial density that is relevant to the problem at hand. However, several challenges including heterogeneity in space and time, varying densities of clusters, and  sampling bias inherent in the data are yet to be addressed.

\begin{figure}[t]
\centering
\includegraphics[width=0.8\textwidth]{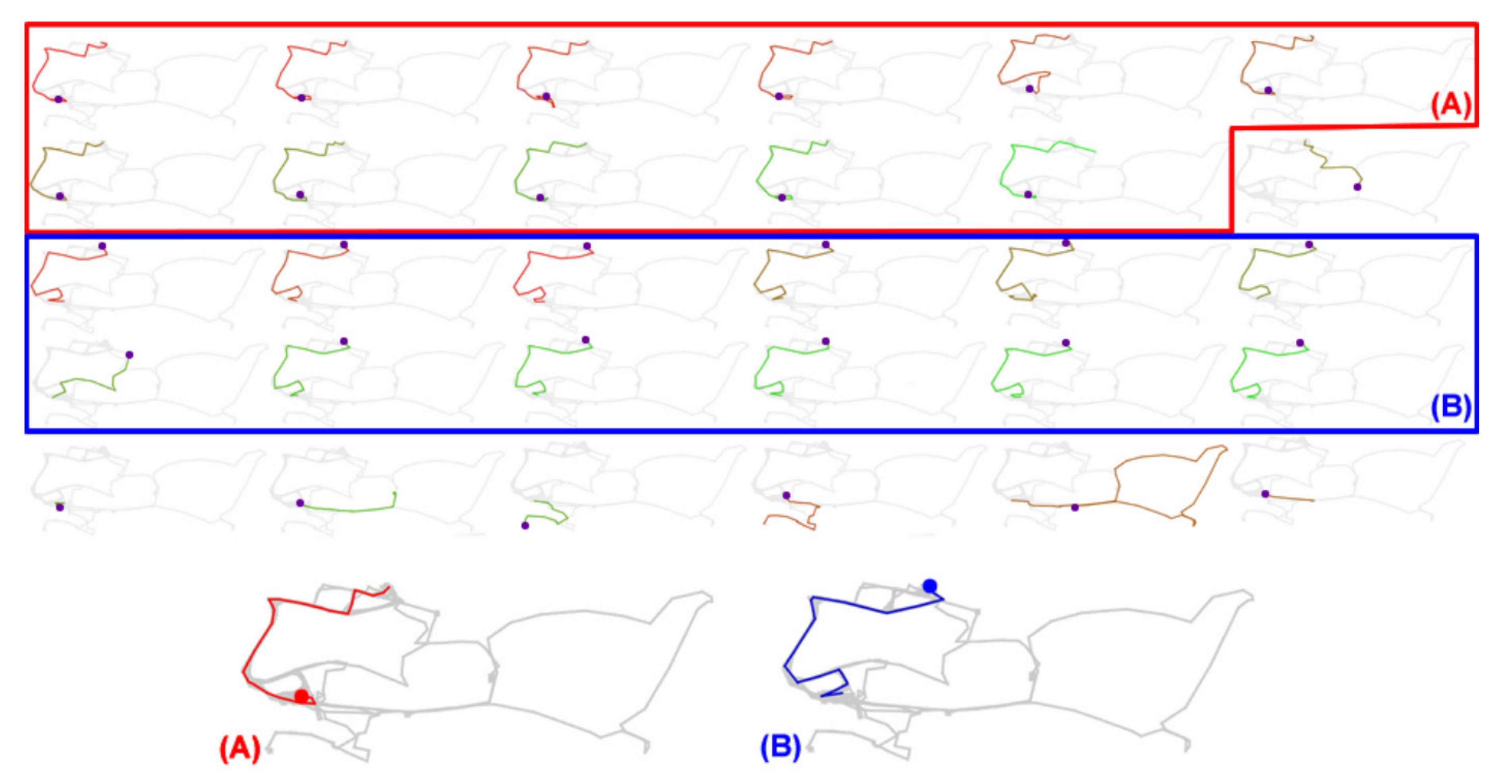}
\caption{Mobility profiles discoverd from a user's GPS traces based on clustering of trips. (Figure taken from \cite{trasarti2011mining})}
\label{fig:clustering_trajectories}
% \vspace{-0.25cm}
\end{figure}

\subsubsection{Clustering Trajectories}
\label{sec:clus_trajectories}
When clustering trajectory data, we are often interested in finding groups of trajectories that are similar to each other across the entire duration of trajectories. For example, clustering of trajectories has been used to find groups of hurricanes with similar trajectories \cite{lee2007trajectory}, groups of taxi traces that follow a similar route \cite{liu2010towards}, or groups of moving objects that follow the same motion in video streams \cite{gaffney1999trajectory}. A review of techniques for clustering trajectories is presented in \cite{kisilevich2010spatio}. Two important aspects of trajectory clustering methods are the choice of distance measure (see Section \ref{sec:trajectory_sim}) and the choice of clustering technique \cite{morris2009learning}. 
One category of methods for clustering trajectories involves using mixture modeling approaches \cite{gaffney1999trajectory,chudova2003translation,alon2003discovering}, e.g., mixtures of regression models where a different regression model is learned for every cluster of trajectories \cite{gaffney1999trajectory}. A different approach has been explored in \cite{trasarti2011mining}, where a two-step clustering method was proposed to find mobility profiles of users based on their GPS traces. Figure~\ref{fig:clustering_trajectories}  shows two sets of discovered profiles (A and B), along with several noisy trips that do not conform to these profiles.

In some trajectory clustering problems, we are interested in finding groups of trajectories that share similarity in only a short duration of the trajectory. One of the methods developed for this problem is a partition-and-group framework called TRACLUS \cite{lee2007trajectory}, which first partitions each trajectory into smaller line segments based on a minimum description length (MDL) principle, and then groups line segments based on their similarity using a DBSCAN-based approach. 

A related area of research is the identification of ``moving clusters'' of trajectories, where moving bodies may join or leave a cluster as it progresses in space over time. This is a common pattern that is observed in several applications, e.g., migrating flocks of animals or convoys of cars. Algorithms for detecting moving clusters of trajectories have been developed in \cite{kalnis2005discovering} and in subsequent studies \cite{jeung2008discovery,li2010swarm,dodge2008towards}. More recently, \cite{zhang2016gmove} proposed an iterative framework, GMove, that alternates between two tasks: assigning users to groups and modeling group-level mobility, using an ensemble of hidden Markov models.

\begin{comment}
{\color{blue} Old Text:\\
One could determine the evolution of groups with time which allows one to study how a group in the first time interval is changing (shifting, shrinking in size, growing in size, disappearing, merging with another group etc.) in successive time intervals. A simpler variant of this can be defined such that contiguous set of locations with similar measurement scores are to be grouped for each time point and the evolution of these groups over time is to be studied. The first half of this objective is similar to the image segmentation problem that is extensively studied in image processing literature where the goal is to determine groups of contiguous locations with similar characteristics that represent objects in an image ~\cite{pal1993review}. Evolution of groups of spatial locations can further be investigated for recurring patterns of changes in the grouping. Some work in studying evolution in dynamic graph provides a good starting point to pursue research in this direction \cite{greene2010tracking,rossi2013modeling}.
}
\end{comment}

\subsubsection{Clustering Time Series}

\begin{wrapfigure}{r}{2.5in}
\vspace{-0.2in}
\centering
\includegraphics[width=2.5in]{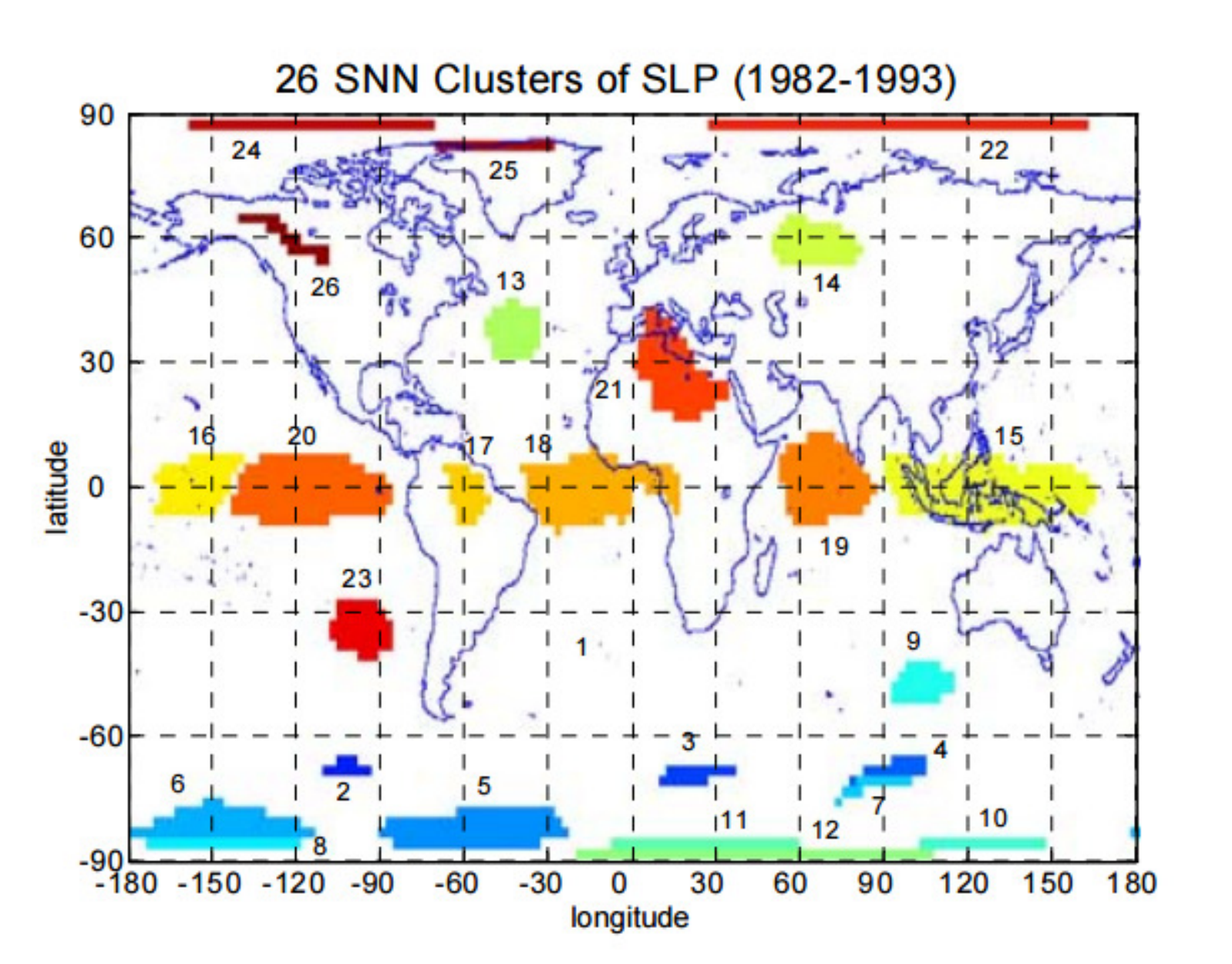}
\caption{Clusters found using SNN clustering of Sea Level Pressure data (1982-1993). (Figure taken from \cite{steinbach2002data})}
\label{fig:clustering_ts}
\vspace{-0.1in}
\end{wrapfigure}

% Using the similarity among time-series discussed in Section \ref{}, one can perform clustering of time-series. However, the spatial information needs to be brought in, either as spatial constraints, or as region-growing/image segmentation approaches.
A central objective when clustering time series derived from ST raster data (using the series of measurements at every location) is to find spatially coherent groups of locations with similar temporal activity. This problem has been approached from multiple directions. One direction is to use traditional clustering schemes such as $k$-means clustering \cite{mezer2009cluster}, hierarchical clustering \cite{goutte1999clustering}, shared nearest neighbor (SNN) clustering \cite{steinbach2003discovery}, and normalized-cut spectral clustering \cite{van2008normalized} to cluster time series (using the time series similarity measures discussed in Section \ref{sec:ts_sim}). The clustering of Sea Level Pressure time series, studied in climate science, from all locations on the Earth's surface is shown in Figure~\ref{fig:clustering_ts}. Data from 1982 to 1993 was used to find these clusters and Pearson's correlation coefficient was used to assess similarity between time series.

% \begin{figure}
% % \vspace{-0.2in}
% \centering
% \includegraphics[width=2.7in]{figures/examples/clustering_timeseries-eps-converted-to.pdf}
% \caption{Clusters found using SNN clustering of Sea Level Pressure data (1982-1993). (Figure taken from \cite{steinbach2002data})}
% \label{fig:clustering_ts}
% % \vspace{-0.1in}
% \end{figure}

Since traditional approaches for time series clustering do not guarantee that the resultant clusters are spatially contiguous, this is typically addressed in a post-processing step either by increasing the number of clusters ~\cite{smith2009correspondence} or by separating clusters with non-contiguous sets into multiple contiguous clusters. Another direction is to directly incorporate spatial contiguity in the clustering process, e.g., by using `region growing' approaches or by enforcing contiguity constraints in the clustering technique. Region growing approaches \cite{heller2006cluster,lu2003region,bellec2006identification} work by merging spatially adjacent locations that are highly similar to each other (defined using similarity measures discussed in Section \ref{sec:ts_sim}) into a single cluster until a minimum number of clusters has been achieved or no cluster can be further grown without violating a similarity criterion. While region growing approaches ensure that every cluster has similar locations, they do not ensure that the different clusters are dissimilar. 
On the other hand, clustering approaches that utilize contiguity constraints \cite{craddock2012whole,blumensath2013spatially} do not have this problem as these approaches ensure that locations within a cluster are highly similar to each other than locations that are in two different clusters.

% Another possibility while clustering locations by looking at their time-series is when clusters involve evolving patterns in time. Locations follow a particular pattern
% Cite Xi's work.

% \begin{figure}[t]
% \centering
% \includegraphics[width=0.48\textwidth]{figures/examples/clustering_timeseries-eps-converted-to.pdf}
% \caption{Clusters found using SNN clustering of Sea Level Pressure data (1982-1993). (Figure taken from \cite{steinbach2002data})}
% \label{fig:clustering_ts}
% % \vspace{-0.25cm}
% \end{figure}

\begin{comment}
{\color{blue} Old Text:\\
Spatial locations in gridded space-time datasets can be characterized based on the temporal behavior, i.e., the time series at a given location. One characteristic of spatio-temporal datasets is the spatial smoothness, also referred to as a spatial autocorrelation. This spatial smoothness introduces redundancy into the data where often a set of contiguous locations have highly similar time series. To reduce this redundancy and to represent spatio-temporal data succinctly it is desirable to represent every set of contiguous locations with highly similar time series as one entity or a group. One example of this is to group highly contiguous locations in the surface of earth that experiences highly similar climatic conditions. Another example of this is to group locations in the brain to determine anatomical locations that potentially exhibit similar time series. 
}
\end{comment}

\subsubsection{Clustering Spatial Maps}

% Using spatial map similarity, we can clustering time points. Extensions to use temporal auto-correlation need to be explored.
When clustering spatial maps (sets of measurements from all locations in an ST raster on different times), the central objective is to find groups of time stamps that have similar spatial maps, e.g., time stamps with similar maps of brain activity \cite{liu2013time}. 
Similarity measures among spatial maps, such as those discussed in Section \ref{sec:map_sim}, can help in discovering meaningful clusters of spatial maps irrespective of data artifacts such as changes in alignment and registration among the locations of different maps \cite{liu2013decomposition}. While these methods are useful in producing temporally contiguous segments with similar spatial activity (due to the temporal auto-correlation in the data), in some cases, we may be interested in identifying non-contiguous groups of time stamps. One way to ensure that a given cluster of spatial maps is not due to temporal auto-correlation is to evaluate this in a post-processing step or use a smaller $k$ such that distant time points are grouped into clusters.  

% This similarity could be either natural (for distant time points) or artefactual (due to temporal autocorrelation for nearby time points, i.e., measurements captured over space between two successive timepoints are known to be more similar than measurements captured at very different time points). Therefore finding sets of time points that are not continuous and whose measurements in a spatial field are highly similar is desired. 
% Such groups of time points are useful to study the patterns of measurements that recur at different points in time \cite{chen2015introducing}.
% One approach that has been used in neuroimaging literature is to employ a $k$-means clustering approach \cite{}. 

%{\color{red} $\bullet$ Discuss the challenge of computing similarity among sub-regions (that have to be determined). Bi-clustering approaches.}

\subsubsection{Finding Dynamic ST Clusters}

A common clustering problem that is of interest when dealing with ST raster data is to identify sub-regions of space and time that show coherent measurements, termed as `dynamic ST clusters' \cite{chen2015clustering}. The discovery of dynamic ST clusters can help in detecting phenomena that only influence a subset of locations during a subset of time points.
For example, water bodies that grow and shrink in space over time, e.g., lakes and reservoirs, can be identified as dynamic ST clusters in remote sensing data,  as they appear as coherent observations in subsets of space and time. Note that a dynamic ST cluster may evolve over time and thus change its shape, size, and appearance as we progress in time. Hence, while some locations are retained across consecutive time stamps, the cluster assignments of locations are dynamic and the clusters can grow and shrink over time. A recent work \cite{chen2015clustering} explored a novel approach to identify dynamic ST clusters in raster data sets for detecting surface water dynamics, which used an iterative algorithm to first identify the set of `core' locations that are part of a dynamic cluster across all time stamps, and then growing around the core locations at every time stamps to capture the dynamic behavior occurring at the boundaries. Similar formulations can be developed for identifying other types of dynamic clusters in ST raster data, e.g., a moving cluster of locations such as the evolution of a hurricane.

\subsubsection{Clustering ST Rasters}

Given a collection of ST raster data sets, possibly collected over different spatial regions and time periods, it is useful to identify groups of ST raster data sets considered as individual instances. This has applications in several domains dealing with ST raster data such as climate science and neuroimaging. For example, in climate science, different climate models produce simulations of global climate variables as ST rasters, which when clustered sheds light on the similarity of climate models \cite{steinhaeuser2014climate}. ST rasters can be clustered by using similarity measures among ST rasters as basic building blocks, which typically involve extracting features from network representations of ST rasters (discussed in Section \ref{sec:raster_sim}).
For example, 
\cite{yu2015assessing} found groups of ST rasters by first constructing networks from ST rasters and then grouping the resultant networks using a network module-finding approach from \cite{newman2006modularity}. However, extracting such high higher-order features from ST rasters is non-trivial and often requires domain expertise.

% Another research direction is the identification of subsets of space and time in a given ST raster that show similar activity and thus appear as coherent ST clusters. For example, water bodies that grow and shrink in space over time, e.g., lakes and reservoirs, can be identified as ST clusters in remote sensing data,  as they appear as coherent observations in subsets of space and time. A recent work \cite{chen2015clustering} explored a novel approach to identify ST clusters in raster data sets for detecting surface water dynamics, which used an iterative algorithm to first identify the most confident locations (that are always part of the cluster), which where iteratively grown around their periphery to include the dynamic locations at every time-step.

% ii) The similarities could exist at a particular spatial and temporal resolution that is different from the native resolution in which the raw data is available. Therefore, significant work in feature learning and resolution selection is desired to make progress on this front. 

\subsection{Predictive Learning}

The basic objective of predictive learning methods is to learn a mapping from the input features (also called as independent variables) to the output variables (also called as dependent variables) using a representative training set. In spatio-temporal applications, both the input and output variables can belong to different types of ST data instances, thus resulting in a variety of predictive learning problem formulations. 
% For example, a scalar output variable can be predicted using ST data instances as input, e.g., time-series, spatial maps, or ST rasters. Furthermore, there can also be scenarios where both the input and output variables show spatio-temporal properties, e.g., while predicting ST points using other ST points. 
In the following, we discuss some of the predictive learning problems that are commonly encountered in ST applications based on the type of ST data instance used as input variables.

\subsubsection{Time Series}
Given an ST raster data, we can consider the problem of predicting an output variable (continuous or categorical) at every location of the raster using the time-series at every location as input variables. The use of time-series as input features is quite common in a number of classification and regression problems. For example, the temporal dynamics of audio frequencies is used to classify words or sentences in human speech recognition problems. This can be achieved by using recurrent neural networks \cite{mikolov2010recurrent,graves2009offline}, which are extensions of artificial neural networks with appropriate skip connections among the neural nodes to model information delay. Another approach to incorporate the temporal context of time-series features in classification problems is the use of shapelets \cite{ye2009time,hills2014classification}. Shapelets are time-series subsequences that are discriminative in nature, i.e. their occurrence is selective to certain classes.  While these techniques provide an ability to model the temporal characteristics of input features, there is a need to develop novel methods that can take into account the spatial information among the time-series in ST rasters. For example, instead of predicting the class label at every location using its time-series independently, we can leverage information about spatial neighborhoods to enforce spatial consistence among the labels at nearby locations. Variants of recurrent neural networks that include spatial features for spatio-temporal prediction have been explored in \cite{jia2017predict,jia2017incremental,jain2016structural}.
Latent space models that use topological as well as temporal attributes of locations have also been developed for real-time traffic prediction using time-varying information from sensor recordings \cite{deng2016latent}. Time series can also be constructed from trajectory data, where the objective is to predict the future location of a moving object (or a group of objects), given their past history of visited locations \cite{li2016knowledge,horton2014classification}.

\subsubsection{Spatial Maps}
In this class of predictive problems, a scalar output variable has to be predicted at a time-step of the ST raster,  using the spatial map at the same time-step as input variables. Some examples of applications that use spatial maps as input features include image classification and object recognition problems, where a categorical value has to be assigned to every image or sub-region in an image using the information contained in spatial maps. A classification approach that has recently gained widespread recognition in the computer vision community is \emph{deep} convolutional neural networks (CNN) \cite{lecun1995convolutional,krizhevsky2012imagenet}, that use the spatial nature of inputs to share model parameters and provide robust generalization performance. In spatio-temporal applications, a promising research direction is to use the temporal auto-correlation among consecutive spatial maps to share the parameters of CNN models over time. Spatio-temporal extensions of CNN based learning frameworks have been developed in \cite{taylor2010convolutional,karpathy2014large}. Extensions of neural networks that use both the spatial and temporal information of data have also been developed in \cite{stiles1997habituation,ghosh1995classification}, where the network design was inspired by the biological information of habituation mechanisms in neuroscience.

\subsubsection{ST Rasters}
Another class of predictive learning problems is to use the entire information in an ST raster as input variables to predict a scalar output variable. 
An example of this is to predict if a subject has a mental disorder or note based on their fMRI scan, stored as an ST raster. This has tremendous applications in diagnosing mental disorders which is currently done in a very subjective manner.
% Another example can be found in the climate domain, where the problem of predicting the number of hurricanes in a given year based on climate variables such as sea surface temperature and pressure \cite{race2010knowledge}, observed over multiple time stamps across a large set of locations, is commonly studied. 
Another application that is increasingly becoming popular in the realm of brain imaging is that of `brain reading' \cite{norman2006beyond}, where the objective is to determine the nature of activity (e.g., planning, memorizing, recollecting etc.) based on measured spatio-temporal activity from the brain, represented as an ST raster. 

% Another widely studied problem is one of predicting Indian Monsoon \cite{delsole2012climate}. \cite{culotta2014estimating} and \cite{culotta2010towards} used a regression based approach for each location independently without taking into account any temporal and spatial relationships. 

A na\"ive approach to this problem could involve representing every ST raster with $|\mathcal{S}|$ locations and $|\mathcal{T}|$ time stamps as a vector of size $|\mathcal{S}| \times |\mathcal{T}|$, and employing traditional classification schemes on these vector representations, e.g., linear discriminant analysis and support vector machines \cite{ku2008comparison}. 
Note that this is only possible when the ST grid of every grid is perfectly aligned with each other, such that their sets of locations and time stamps are identical. This is not usually the case with resting state fMRI scans, because the time points in one subject's scan cannot be matched with those of another subject. 
Furthermore, especially in fMRI data, the large number of spatial locations (typically hundreds of thousands) and time points (typically hundreds) leads to millions of potential features, where the number of instances are often tens to hundreds, leading to the phenomena of model overfitting. Hence, it is important to use \emph{derived} features from ST rasters that summarize the ST activity of every raster, as described in Section \ref{sec:raster_instance}. Alternatively, tensor learning based approaches \cite{zhou2013tensor,bahadori2014fast,yu2015accelerated} provide a way to reduce the model complexity by making use of the spatial and temporal dependencies among the input features, thus showing a promise in predictive learning problems involving input ST rasters.

\subsubsection{ST Reference Points}

A common predictive learning problem in spatio-temporal applications is to predict the response at a certain location and time using observations collected at other locations and time stamps (often in ST neighborhoods). This is important in a number of domains, e.g., while estimating an ecological variable over every location and time using remote sensing observations at nearby locations and time stamps. The problem of land cover classification (estimating a categorical label at every location and time indicating its propensity to belong to a land cover type) has been heavily studied in the remote sensing literature for a variety of problems \cite{defries2000multiple,vatsavai2008machine,jun2011spatially,li2014review}, e.g., the mapping of surface water dynamics using multi-spectral remote sensing data \cite{rse,karpatne2016global}.
As another example, the outbreak of influenza at a given location and time can be  predicted based on web searches \cite{ginsberg2009detecting} and twitter messages \cite{culotta2010towards} at neighboring locations and times. There are two classes of methods that are relevant for making predictions at ST reference points: methods that use the temporal information to predict values at nearby time points, and methods that use the spatial information to estimate values at nearby spatial points. We discuss both these classes of methods in the following.

\paragraph{Using Temporal Information:}
% An example is that of predicting precipitation for the upcoming season based on previous precipitation data. This is highly relevant in countries that have an agriculture based economy as they can select the crops for the impending season appropriately. 
In many domains such as climate and health, estimation (or forecasting) of the future conditions based on present and past conditions is desired. For example, the sea surface pressure and temperature for the present month can be predicted based on values in the previous months. 
Similar problems are studied in predicting closing stock prices at the New York Stock Exchange and in predicting sales in the manufacturing industry \cite{montgomery2015introduction}. 	
Some of the widely used methods for time-series forecasting problems include exponential smoothing techniques \cite{gardner2006exponential}, ARIMA models \cite{box1976time}, and state-space models \cite{aoki2013state}. Another type of methods for making predictions at time stamps include dynamic Bayesian networks such as hidden Markov models and Kalman filters \cite{rabiner1986introduction,harvey1990forecasting}, that estimate the most likely sequence of latent values at time stamps using the temporal auto-correlation structure. Techniques that make predictions in time need to be modified to include the spatial context in spatio-temporal applications. As an example spatio-temporal Granger causality models, that use both spatial and temporal information in the regression models, have been explored to model relationships among ST variables \cite{lozano2009spatial,luo2013spatio}.

\paragraph{Using Spatial Information:}
There is a vast body of literature on spatial prediction methods that take into account the spatial auto-correlation structure in the data to ensure spatially coherent results.
This includes the use of spatial auto-regressive (SAR) models \cite{kelejian1999generalized}, geographically weighted regression (GWR) models \cite{brunsdon1998geographically}, and Kriging \cite{oliver1990kriging} in the spatial statistics literature.  Markov random field based approaches that are naturally suited to handle the spatial auto-correlation in the data have also been widely studied \cite{kasetkasem2002image,schroder1998spatial,zhao2007classification}. There is a promise in informing such techniques with the temporal nature of ST points in spatio-temporal applications, such that both spatial and temporal auto-correlation are incorporated in the modeling framework. For example, spatio-temporal Kriging approaches have been used in a number of applications in climate and environmental modeling \cite{cressie2015statistics}, where time is treated as another dimension while learning covariance structures in space and time. A related area of research is spatial item recommendation \cite{wang2017st}, where the preference over spatial items (e.g., restaurants or tourist attractions) have to be predicted in a time-varying manner, using social network information and the history of preferences of every user.

% {\color{green} Old text:\\
% Predicting the current prevailing conditions at some locations based on prevailing conditions at other locations. For example, the problem of predicting climate variables such as air temperature and precipitation on several points on the land based on climatic conditions in the oceans has been studied \cite{chatterjee2012sparse}. \cite{chatterjee2012sparse} used a sparse group lasso approach that ensures that all climatic variables from an oceanic location contribute to the prediction at a land location. However, this approach does not ensure that adjacent spatial locations have similar contributions due to spatial autocorrelation in the covariates. 

% }

% \subsubsection{Input and Output: ST Points}

% {\color{green} Old Text:\\

% An example of this is to predict crop-yield based on a number of climate and environmental factors. 

% }

% {\color{red} $\bullet$ Discuss Yan Liu's work on spatio-temporal Granger causality models.}

% $\bullet$ {\color{red} Discuss the example of `impact of climate change on agricultural productivity'.}

\subsection{Frequent Pattern Mining}
%There are three basic types of patterns that we discuss in this paper: (a) frequent patterns, (b) infrequent patterns (anomalies), and (c) changes.
Frequent pattern mining is the process of discovering patterns in a data set that occur frequently over multiple instances in a data set, e.g., frequently bought groups of items in market-basket transactions. Given the rich variety of data types and instances in ST applications, there are several categories of frequent pattern mining problems that can be formulated in the presence of spatial and temporal components, as described in the following. 
% This includes the detection of frequently co-occurring ST event types, sequential patterns in ST events, frequent sub-sequences in trajectory data and time-series data, and frequent patterns in network-based representations of ST rasters, as discussed below.

\subsubsection{Co-occurrence Patterns in ST Points}
Given a collection of points such as ST events of varying types, a common pattern of interest is co-occurrence patterns, which are basically subsets of ST event types that occur in close spatial and temporal proximity of each other. For example, given a data set of crime and other events in a city, we may be interested in finding ST event types that occur together (e.g., bar closing and drunk driving). Co-occurrence patterns has been studied in the context of spatial datasets as `co-location' patterns for more than a decade \cite{huang2004discovering,xiao2008density,wang2013finding}. In the spatial statistics literature, co-locations have been studied using measures such as the Ripley's cross-$K$ function \cite{dixon2002ripley}, which capture statistical patterns of attraction or repulsion among pairs of spatial point types.

% A natural extension of co-location patterns to spatio-temporal data sets, particularly the ST point-type data, is a co-occurrence pattern that is defined as a subset of point types that occur in close spatial and temporal proximity \cite{pillai2013filter}. 
One of the first approaches for discovering co-occurrence patterns include an Apriori-principle based approach developed by \cite{pillai2012spatio}. This approach relies on an interestingness measured termed as \emph{spatio-temporal co-occurrence co-efficient} to capture co-occurrence in spatial and temporal dimensions. A filter and refine approach was later proposed in \cite{pillai2013filter} to tackle the problem of discovering co-occurrence patterns in large event databases. Constructing rules from co-occurrence patterns have also been studied in \cite{pillai2014spatiotemporal}. Spatio-temporal extensions of the cross-$K$ function have been proposed in \cite{lynch2008spatiotemporal}, where the ST neighborhood of ST points is used for identifying attraction or repulsion patterns among ST point types in both space and time.

\begin{comment}
\begin{itemize}
\item Cross-K, co-location patterns, and its ST extensions (frequent occurrences of point event types)
\item Trajectory Patterns (finding common sub-sequences in multiple trajectories)
\item Time series Motifs
\end{itemize}
\end{comment}

\subsubsection{Sequential Patterns in ST Points}
Sequential patterns have been studied in the context of ST event data, where the occurrence of ST events of a particular type can trigger a sequence of ST events of other types. For example, a car accident on a freeway could trigger a traffic jam in its ST neighborhood.
Sequential patterns have been originally defined in the context of market-basket transactions where sequences of transactions from every customer are available and the goal is to discover ordered list of items appearing with high frequency \cite{agrawal1995mining}.  An approach for discovering sequential patterns of ST event types was presented in \cite{huang2008framework}. This approach was able to discover ordered lists of event types such as $f_1 \rightarrow f_2 \rightarrow \ldots \rightarrow f_n$, where events belonging to type $f_1$ trigger events of type $f_2$, that further triggers events of type $f_3$ and a series of events resulting in events of type $f_k$. They developed an Slicing-STS-Miner approach to efficiently discover statistically significant sequential patterns of events. A partially-ordered subsets of event types, referred to as cascading spatial temporal patterns \cite{mohan2012cascading,mohan2010cascading}, have also been studied to capture event sequences whose instance are located together and occur in successive stages. Some of the key challenges in mining ST sequential patterns include defining interesting measures that capture meaningful non-spurious patterns and developing efficient approaches to discover interesting patterns from an exponentially large space of candidate patterns.

\subsubsection{Sequential Patterns in Trajectories}
\label{sec:trajectory_pattern}

Sequential patterns in trajectories are sequences of spatial locations or regions that are visited by multiple moving objects in the same order, and are thus covered by multiple trajectory instances. For example, given the trajectories of tourists visiting a city, we can find sequential patterns that represent the typical behavior of tourists as they visit an attraction and move to another. 
% Note that in this case, the individual trajectories that support a trajectory pattern can happen at very different time points.
% Trajectories that represent paths travelled by moving objects such as taxis are also akin to sequences of spatial locations on their corresponding paths. Discovering commonly occurring sequences in these paths has gained interest in these application domains \cite{giannotti2007trajectory,hwang2005mining,jeung2008discovery,liu2012mining,li2013mining,vieira2009line}. 
Although trajectories can be treated as multi-dimensional sequences where the spatial identifiers (e.g., latitude and longitude) of visited locations are used as dimensions, traditional sequential pattern mining approaches are unsuitable for finding sequential patterns of trajectories because of their inability to handle the spatial nature of information contained in trajectories. For example, the locations visited by moving objects may not exactly match but be in close spatial vicinity to be considered as part of the same pattern. Moreover, trajectories belonging to a common sequential pattern are expected to follow similar spatial path but not exactly the same path. To handle the spatial aspect of trajectories in the process of mining sequential patterns, \cite{tsoukatos2001efficient} used ordered sequences of spatial elements such as rectangular regions to define sequential patterns. A trajectory is then considered to be part of a sequential pattern if every line segment joining consecutive locations of the trajectory are enclosed within the rectangular regions of the pattern. A similar approach for sequential pattern mining was developed in \cite{cao2005mining} where pattern elements were defined as spatial regions around frequent line segments of trajectories. Spatio-temporal extensions of association rules, termed as spatio-temporal association rules (STARs) \cite{verhein2006mining}, have also been explored to find frequent patterns of geographic regions visited by moving objects \cite{verhein2008mining}. Approaches for mining periodic patterns in trajectories, where moving objects periodically visit sequences of locations in time,  have been developed in \cite{li2014mining}.
Approaches to handle uncertainty have also been studied to handle the incompleteness of the data and artificial noise introduced in privacy-sensitive applications \cite{chen2016mining,li2013mining}.

% \begin{figure}[t]
% \centering
% \includegraphics[width=0.48\textwidth]{figures/examples/patterns_trajectories-eps-converted-to.pdf}
% \caption{Example of a sequential patterns found in GPS traces of trucks in Athens, Greece. The pattern A $\xrightarrow{\Delta t1}$ B$^\prime$ $\xrightarrow{\Delta t2}$ B$^{\prime \prime}$ is the finer pattern of a coarse pattern A $\xrightarrow{\Delta t1}$ B $\xrightarrow{\Delta t2}$ B that captures round trips taken by trucks. (Figure taken from \cite{giannotti2007trajectory})}
% \label{fig:patterns_trajectories}
% % \vspace{-0.25cm}
% \end{figure}

% \begin{figure}
% % \vspace{-0.3in}
% \centering
% \includegraphics[width=2.7in]{figures/examples/patterns_trajectories-eps-converted-to.pdf}
% \caption{Example of a sequential patterns found in GPS traces of trucks in Athens, Greece. (Figure taken from \cite{giannotti2007trajectory})}
% \label{fig:patterns_trajectories}
% % \vspace{-0.1in}
% \end{figure}

\begin{wrapfigure}{r}{2.5in}
\vspace{-0.2in}
\centering
\includegraphics[width=2.5in]{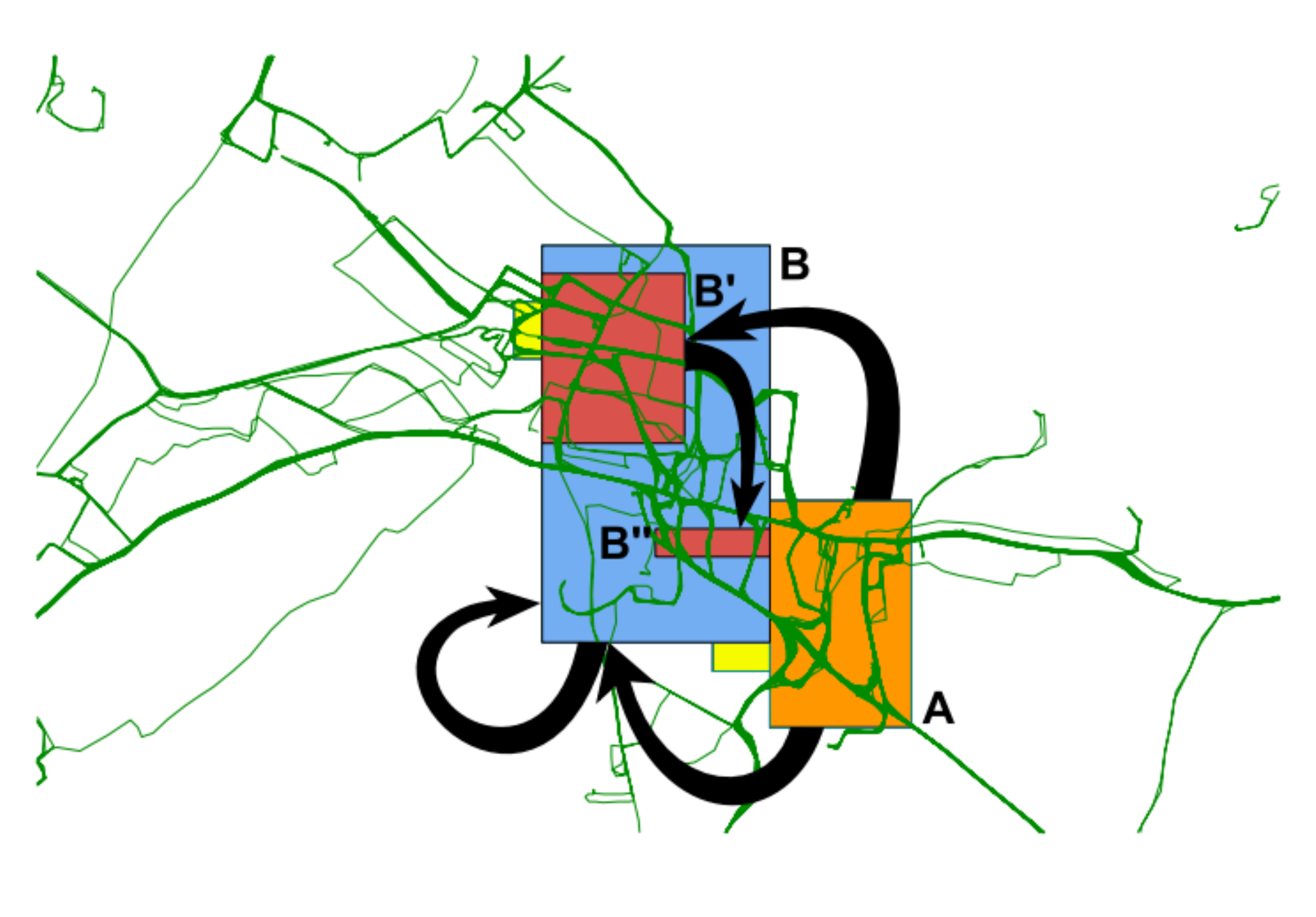}
\caption{Example of a sequential patterns found in GPS traces of trucks in Athens, Greece. (Figure taken from \cite{giannotti2007trajectory})}
\label{fig:patterns_trajectories}
\vspace{-0.1in}
\end{wrapfigure}

Another aspect of trajectories that is important to consider while mining sequential patterns in some applications is the presence of timing constraints among visited locations. For example, we may be interested in finding sequential patterns where the time duration between consecutive locations are similar across the trajectory instances, in addition to the similarity of paths followed by the moving objects \cite{giannotti2007trajectory}. This approach was used to discover sequential patterns in GPS traces of trucks in Athens, Greece. One resultant pattern A $\xrightarrow{\Delta t1}$ B$^\prime$ $\xrightarrow{\Delta t2}$ B$^{\prime \prime}$ is shown in Figure~\ref{fig:patterns_trajectories} that captures how the trucks visit region A first, then visit region B$^\prime$, and then visit region B$^{\prime \prime}$, with fixed time gaps between the locations. Another way of introducing timing constraints is to consider sequential patterns of trajectories that are in close temporal proximity of each other, i.e., the moving objects visit a sequence of locations at similar points in time. Such patterns are referred to as `flock patterns' \cite{vieira2009line} or `convoys' \cite{jeung2008discovery} that are used for finding groups of objects moving together.

%Example: When a subject is not involved in any task, the human brain is known to exhibit a `default mode activation'. Discovering such frequent activation patterns are of interest in neuroscience.

\subsubsection{Motif Patterns in Time-series}

In the context of ST raster data, time-series motifs represent repeated temporal measurements observed across multiple spatial locations. 
For example, in vegetation time series data of agricultural farms, the harvesting cycles of crops can be discovered as time-series motifs that recur at multiple farm locations belonging to the same crop type. 
% Time-series motifs are temporal sequences of measurements that frequently recur in time-series data sets \cite{patel2002mining}. 
% A real-world example of a time series motif is that of a repeated sequence of sounds made by a family of birds or insects, which can be extracted from a large data base of time-series of bird sounds. 
Time-series motifs have been extensively studied in applications such as ecology, medicine, finance, and production industry \cite{mueen2014time,torkamani2017survey}.
While the na\"{\i}ve approach for discovering motifs is quadratic in the length of the time series, approximate approaches \cite{chiu2003probabilistic,meng2008mining,minnen2007discovering} have also been designed. 
% Exact approaches \cite{mueen2009exact} have also been explored that are practically efficient even though they are worst case quadratic. 
Recently, the development of efficient approaches for constructing `matrix profile' \cite{yeh2016matrix}, a vector comprised of minimum non-trivial distance for each subsequence in time series $A$ with subsequences in time seies $B$, have enabled an efficient approach for motif discovery \cite{zhumatrix}. At the heart of this approach for computing the matrix profile is a fast Fourier transform based algorithm for efficiently computing z-normalized Euclidean distance between two time series subsequences. Using this approach, subsequence similarity in any given time series can be efficiently computed and the subsequences that are highly similar correspond to motifs.

While traditional formulations for motif discovery assume independence among the time-series, in ST applications, there is a strong spatial auto-correlation among the time-series at nearby locations, limiting the usefulness of existing formulations. This is because any time-series motif observed over a set of locations may also show up in the neighborhood of those locations with minor variations, resulting in a number of redundant patterns that look almost similar.
One approach to address this challenge is to enforce motif discovery algorithms to identify recurring temporal patterns that appear synchronously in spatially coherent regions, in contrast to isolated points in space. 
This would result in the discovery of physically relevant processes and events that span both space and time, e.g., droughts or floods affecting an ST region in climate data sets. We can call such ST motifs as `spatially coherent time-series motifs.' It may also be useful to explore a slightly modified version of this problem where locations belonging to the same region do not synchronously show a common pattern but instead there is a gradual evolution of temporal activity across them. Such `evolving ST motifs' can help us detect frequent ST phenomena that gradually moves across locations in a region.

An active research direction is to discover structural patterns in ST data that elucidate complex spatial and temporal dynamics.
For example, \cite{zhou2015predicting} developed approaches to discover ST dynamics in ambulance demand data for ambulance fleet management. A major challenge in this direction is to handle the heterogeneity in the presentation of different patterns in space and time. Sparsity of the underlying data also poses a challenge to capture patterns of interest. While early efforts in using tensor-based factorization \cite{takahashi2017autocyclone}, multi-task learning \cite{zhao2015multi}, and spatio-temporal kernel density estimation \cite{zhou2015predicting} approaches have shown promise, their effectiveness in handling heterogeneity and sparsity is yet to be investigated.

\subsubsection{Network Patterns in ST Rasters}

Given a collection of ST rasters, we are interested in discovering patterns of structure in network-based representations of ST data (described in Section \ref{sec:raster_instance}), which occur frequently in multiple ST rasters. 
A commonly studied type of network structure is the arrangement of \emph{communities} in the network, where a community represents a group of entities (nodes or links) that interact among themselves more than they do with the rest of the entities \cite{leskovec2010empirical}. 
Another type of network structure is network motifs \cite{milo2002network}, which are the basic building blocks of the connectivity patterns that frequently occur across multiple network instances. The use of network motifs has been explored in several domains, e.g., bioinformatics \cite{shen2002network} and social networks \cite{milo2002network}. 

In ST applications, network motif discovery has been used to identify the inherent ST structures of brain fMRI data \cite{sporns2004motifs}. Network communities in ST rasters can reveal interesting high-level patterns. For example, in climate data, a community can represent a set of distant locations that are experiencing similar climatic conditions and show consistent temporal activity. While the discovery of network patterns can prove to be a valuable tool in the analysis of ST raster data sets, care must be taken while using them as the spatial auto-correlation in the data can result in a number of spurious edges that can lead to misleading results. Hence, during the evaluation of patterns in network-based representations of ST rasters, it is important to filter out patterns arising from spatially neighboring locations for the discovery of long-range dependencies between distant locations.

%$\bullet$ {\color{red} Cover recent work related to motifs in ST data.}

\subsection{Anomaly Detection}

Anomalies are traditionally defined as instances that are remarkably different  from the majority of instances in the data set. In ST applications, detecting anomalies (also referred to as outliers) can help us identify interesting but rare phenomena, e.g., an anomalous trajectory taken by a taxi vehicle or the changing frequencies of floods and droughts in certain areas due to climate change or human actions that change the ecology of a region. 
Given a method for computing pair-wise similarities among ST instances (discussed in Section \ref{sec:sim_instances}), anomalies can be identified as instances that are highly dissimilar from the rest of the instances. Detecting anomalies can also be seen as a by-product of clustering, since instances that do not easily become part of a cluster or form an isolated cluster that is quite far from the other clusters can be termed as anomalies. There are two areas of work that are relevant in the context of detecting anomalies in ST data. First, a number of techniques have been developed for identifying anomalies in time series data, which are also called as discords.
% Observations in a time-series are considered anomalous if they are significantly deviant from other observations in its local temporal context (at nearby time points), from previous observations at the same position in a periodic cycle (e.g., same month or season), or from the entire history of observations in the time-series. Time series anomalies are also referred to as discords. 
This includes approaches for univariate time series \cite{bishop1994novelty,keogh2002finding,wei2005assumption,keogh2005hot} as well as multivariate time series \cite{chen2008exploiting,galeano2006outlier,baragona2007outliers,takeishi2014anomaly}
A survey of anomaly detection approaches for temporal data is presented in \cite{gupta2014outlier}. 
Second, approaches for detecting anomalies in spatial data, also termed as spatial outliers, have also been developed \cite{shashi2003spatial}. Spatial outliers are defined as points whose non-spatial attributes are different from those that are in its close spatial neighborhood . 
% Note that a spatial outlier may not be anomalous when viewed in the global context of all data points. Hence, the notion of abnormality of every spatial point is determined only by the nature of points in its local spatial neighborhood, and any point that deviates from this locally-defined normal behavior is termed an anomaly.
% Hence, a spatial outlier detection algorithm generally has two basic ingredients: determining the notion of normal behavior in the local neighborhood of every point, and measuring the deviation of every point from this normal behavior.
Techniques for detecting spatial outliers vary in their choice of method for constructing local neighborhood and assigning anomaly scores, e.g., using spatial distance measures \cite{knorr1997unified,knorr2000distance,lu2003algorithms}, graphical-distance measures \cite{shekhar2001detecting,kou2007spatial}, and visualization tools such as the Moran's I scatter plot, pocket plots, and the Variogram Cloud plot \cite{anselin1994exploratory,haslett1991dynamic}. A review of techniques for spatial outlier detection is presented in \cite{aggarwal2017spatial}.
However, the joint presence of spatial and temporal aspects introduces novel ways of describing anomalies in ST data, resulting in unique anomaly detection problems, as described below.

% {\color{blue} Old Text from Event Detection section that seems relevant as anomaly detection:\\

% An example of an event is a raise and fall of oxygenation in fMRI data at a voxel or a set of voxels that lasts for about 10 seconds that arises due to an external or internal stimuli \cite{logothetis2004interpreting}. Another event could be a significant change in the vegetation at a particular location or a set of locations, indicating change in land use from farm-land  to a building or a park to a commercial building \cite{boriah2009detecting}. Another type of event that are of interest are discovering incidences of flu or calamities like earthquakes from social media data such as Google searches \cite{ginsberg2009detecting} or Twitter posts \cite{earle2012twitter,sakaki2010earthquake}. Detecting such events are of interest to understand the changes and the time at which they are happening in the spatial field. Typically every event is characterized by a unique temporal signature and so techniques designed for detection of one event does not necessarily suit the detection of other events.

% }

\subsubsection{ST Point Anomalies} 
\label{sec:point_anomalies}

% \textbf{[Discuss about spatial outlier detection methods such as the use of Moran's I scatter plot and Variogram clouds.]}

% detecting spatial outliers. While some of these  approaches make use of Euclidean distance based proximity measures 

% can be classified based on several dimensions: distance measure (Euclidean distance based proximity \cite{knorr1997unified,knorr2000distance} vs. depth based \cite{ramaswamy2000efficient} vs. graph based proximity \cite{shekhar2001detecting}), span (local \cite{breunig1999optics,chawla2006slom} vs. global \cite{ramaswamy2000efficient} outliers), and type of approaches (metric space based \cite{knorr2000distance} vs. clustering based \cite{ng1994efficient} vs. graph based approaches \cite{kou2007spatial}).

The concept of spatial outliers can be easily extended to spatio-temporal domains, where the neighborhood of an ST point is defined with respect to both space and time. Thus, a spatio-temporal outlier is basically an ST point that breaks the natural ST auto-correlation structure of the normal points. One approach to find ST outliers is to first cluster the normal points using approaches such as ST-DBSCAN, and then report points that did not conform well to the discovered clusters \cite{kut2006spatio}. Another category of approaches presented in \cite{cheng2004hybrid,cheng2006multiscale} is to aggregate ST clusters at a coarser scale so that the effect of outliers on the clustering is reduced. ST outliers are then detected by comparing the original clustering with the coarsened clustering. Note that most ST outlier detection algorithms assume homogeneity in neighborhood properties across space and time, which can be violated in the presence of ST heterogeneity. This can be handled by methods that model the variance of normal instances in the local neighborhood of every point, along with its expected value \cite{sun2004local}.

% This definition can be extended to spatio-temporal problems, where the context of every ST reference point (i.e., its ST neighborhood) is used to define 

% Outliers in spatial datasets are analogous to anomalies in time series data, where non-spatial attributes of objects located in close proximity are expected to be similar and those that violate this expectation are considered as spatial outliers  

\subsubsection{Trajectory Anomalies}

% There are two basic categories of approaches for detecting anomalous trajectories of moving objects. The first category considers a trajectory to be anomalous if its spatial features (e.g., shape, distance, or angle)  are very distinct from the remaining trajectories in the data. The second category focuses on the temporal information of trajectories to identify anomalous trajectories that deviate from their historical behavior. In the following, we briefly describe approaches in both these categories for detecting anomalous trajectories.

There are several ways we can identify a trajectory to be anomalous. A common approach is to compute pair-wise similarities among trajectories (discussed in Section \ref{sec:trajectory_sim}) and identify trajectories that are spatially distant from the others. For example, \cite{lee2008trajectory} explored varying notions of distances between trajectory sub-sequences (such as perpendicular distance, parallel distance, and angle distance) and identified a trajectory to be anomalous if it had very few neighbors within a certain threshold. Distance based methods for detecting anomalous trajectories have also been explored in \cite{bu2009efficient}, where trajectories that appear in distant spatial regions than the rest of the trajectories are considered anomalous. Another class of approaches is to aggregate the spatial shapes of trajectories using a coarse spatial grid, and then compute useful characteristics of a trajectory in every spatial grid cell. For example, we can compute the average direction of trajectories in every spatial grid and identify a trajectory as anomalous if it deviates from the expected behavior in a number of grid cells that it covers, as proposed in \cite{ge2010top}. A graph-based approach for detecting outliers in traffic data streams was developed in \cite{liu2011discovering}, where nodes are regions and edge weights represent the traffic flow between regions. Traffic outliers were discovered as edge anomalies in this graph, which were subsequently analyzed for causal interactions using causal outlier trees. Supervised methods for detecting anomalous trajectory shapes (or motifs) have also been explored in \cite{li2006motion,li2007roam}.

While the techniques discussed above for identifying trajectory anomalies focused on the spatial pattern (e.g., shape, distance, or angle) of trajectories, we can also consider a trajectory to be anomalous if it deviates from its local neighbors as it switches from one cohort to other, e.g., a zebra moving from one group to another. To identify such anomalies, the approach presented in \cite{li2009temporal} computed the spatial neighbors of every moving object at any given time stamp (using trajectory information in the immediate history) and designed an anomaly score based on deviations from the trajectories of neighbors. Such an approach can enable the discovery of trajectories that are contextually anomalous in their local neighborhoods, and can thus be robust to aggregate changes such as population shifts. It can also be useful for detecting anomalies in an on-line fashion, as trajectories that begin to drift apart from their local neighbors will start to accumulate large anomaly scores in a short number of time stamps.

\subsubsection{Group Anomalies in ST Rasters}

While techniques described in Section \ref{sec:point_anomalies} are well-suited for discovering ST point anomalies, it is often the case that anomalies appear in ST raster data as spatially contiguous groups of locations  (regions) that show anomalous values consistently for a short duration of time stamps. Some examples of such group anomalies in ST raster data include rare events such as cyclones, floods, and droughts that result in abnormally high or low precipitation in a given region for a certain duration of time, or an abnormal number of tweets or emergency calls from a spatial region in a small time window. 
% The detection of such anomalous objects in ST raster data require novel methods that can incorporate the spatial as well as temporal characteristics of the anomalous object.

Most approaches for detecting group anomalies in ST raster data decompose the anomaly detection problem by first treating the spatial and temporal properties of the outliers independently, which are then merged together in a post-processing step. For example, the approach presented in 
\cite{wu2010spatio} made use of the spatial scan statistic to find the top-$k$ contiguous groups of locations that showed anomalous activity at every time stamp. These groups of locations were then stitched together in time to find ST groups of anomalous values. A similar line of work has been explored in \cite{lu2007detecting}, where the anomalous spatial regions were discovered using an image segmentation algorithm and  a regression technique was applied to track the movement of the centers of the anomalous regions across consecutive time stamps.
Another approach for detecting anomalies in ST raster data was explored in \cite{faghmous2013parameter} for discovering ocean eddies in climate data. Ocean eddies are revolving masses of water in the ocean that appear as local extremas (depressions or elevations) in a snapshot of sea surface height data on any given time stamp, and these extremas keep moving over time. The approach presented in \cite{faghmous2013parameter} identified eddies as  local extremas in space using a bottom-up thresholding scheme, which could then be stitched in time using multiple hypothesis object tracking procedures \cite{faghmous2013multiple}. 
An orthogonal approach to detecting anomalies in ST rasters is to find anomalies in time series data that can then be merged across space \cite{faghmous2012novel}.

% Similar approaches have been explored for tracking cyclones as local minimas in sea level pressure data \cite{stolorz1995fast}.

Approaches for detecting anomalies in ST raster that jointly use information about the spatial and temporal aspects of the data have also been developed. For example, in the application of detecting abnormal activities in crowded scenes from surveillance camera videos, anomaly detection approaches using models of normal activity have been developed in \cite{li2014anomaly,kratz2009anomaly}. In this problem, a major difficulty is to come up with a notion of normal motion activity that can be differentiated from anomalous activities, which can be quite complex in crowded atmospheres. To address this challenge, \cite{li2014anomaly} used a mixture of dynamic textures models to capture the spatial and temporal salience of normal activities, borrowing from the vast literature on activity recognition in the area of computer vision. Spatial and temporal anomaly maps are then constructed at multiple spatial scales, which are then integrated together using a conditional random field framework. On the other hand, the approach presented in \cite{kratz2009anomaly} model the steady-state motion behavior of normal space-time volumes that can capture the variations in the ST data and can compactly represent the overall video volume. These models are then used to detect unusual motion activities as statistical deviations from the normal model.

A special type of group anomaly in ST rasters is \emph{bursts} of activity in the time series of locations, detected for short time periods. For example, Twitter queries involving specific terms such as `earthquake' can be discovered as spatio-temporal burst events related to natural disasters. One of the early works on detecting ST burst events includes the work by \cite{lappas2012spatiotemporal}, where both the spatial and temporal nature of burst patterns were jointly taken into consideration. A system for mining ST burst events, termed as the Spatio-TEmporal Miner (STEM), has been developed in \cite{lappas2013stem}.

% \subsubsection{ST Raster Network Anomalies}

\subsection{Change Detection}

The problem of change detection involves identifying the time point when the behavior of a system undergoes a significant deviation from its past behavior. Detecting changes in ST data is useful in several applications, e.g., discovering transitions from an El Nino phase to a La Nina phase in climate science \cite{miralles2014nino} or understanding the switching of the state of the brain from planning to cognition in brain imaging data \cite{tang2012neural}. 
% The time points at which the underlying system switches from one phase to another needs to be discovered from the data.
% An alternate view of this problem is one of determining segments of time where the system exhibits a cohesive pattern. 
% In the following, we first describe methods for identifying changes in time-series at individual locations of an ST raster, and then discuss approaches for identifying changes in both space and time.

%In this survey, we refer the objective of identifying `change' whose signature or presentation is not known in advance. When the presentation of event is known in advance we refer to the problem as that of event detection. For example, we refer to the objective of discovering changes in the brain state where we do not know the characterization of each state or the change in advance as change detection. On the other hand we refer to the objective of determining the event of forest fire, that is expected to result in a dramatic decrease in the vegetation at a location, as event detection.

% \subsubsection{Changes in Time Series}

%Example: Climate phenomenon switches from El Nina to La Nina every few years; Human brain switches from one task to another. Discovering such changes in state of the system in question can shed light on the state transitions of the system.

Change detection has been extensively studied  in the context of time-series data, where the objective is to determine time intervals (segments) that exhibit homogeneous properties  \cite{keogh2004segmenting,liu2008novel,mithal2012time,huang2013hinging}. 
% One application of this segmentation work is to determine time intervals where distinct activities were performed by a user wearing sensors \cite{chamroukhi2013joint}. 
Different types of homogeneity in segments such as mean  \cite{bernaola2001scale,horvath2001change}, variance \cite{whitcher2000multiscale,andreou2002detecting,inclan1994use}, and distribution statistics \cite{carpena1999statistical,grosse2002analysis} have been used for identifying time series segments and their changes. 
% While much work has been done in the area of time series segmentation for univariate time series \cite{keogh2004segmenting,keogh2001online,liu2008novel,boriah2008land,yuan2005graph,huang2013hinging,himberg2001time}, some of these approaches have also been extended to multivariate time series \cite{guo2012adaptive,fox2014joint,chamroukhi2013joint}. 
Apart from time series segmentation, other approaches has also been explored for identifying changes in time-series. For example, variational approaches for switching state space models have been used to discover abnormal breathing episodes during sleep (also known as sleep apnea) based on chest volume measurements \cite{ghahramani2000variational}. Another category of approaches is change detection methods for periodic or semi-periodic data. 
% Moving average based change detection is shown to be promising for discovering disturbances in vegetation amidst challenges such as seasonal patterns and noise in the data \cite{boriah2009detecting}. 
\cite{boriah2010comparative} defined an iterative bottom-up approach for change detection in periodic time series for identifying forest fires, where adjoining annual segments are merged if they are highly similar, after accounting for the noise and seasonal variability in the data. Recently, Chandola et al. \cite{chandola2011scalable} defined different types of changes in `periodic' time series (which exhibits annual cycles) and proposed a Gaussian Process based approach for discovering such changes. 
% The cost of each merge is determined using the Manhattan distance between the two annual segments. The last merge in their approach is treated as the incidence of forest fire. 
% This approach is effective in handling the seasonality in the data, but it cannot handle noise and variability in the data. To handle this problem a variation of this approach that takes variability into account is proposed \cite{}, where a new merge cost is determined as the ratio of the original merge cost to the minimum merge cost of all the merges. 
Other techniques for time series change detection have been explored in remote sensing applications \cite{karpatne2016monitoring}, e.g., using anomaly detection approaches \cite{lunetta2006land,mithalCIDU2011}, forecasting-based approaches \cite{chandola2011scalable,Liang2014}, and sub-sequence pattern matching approaches \cite{Salmon2011,Zhu2012}. 

% A wide range of algorithms have been proposed to identify changes in time-series for remote sensing applications , e.g., using segmentation-based approaches, 

In ST applications involving raster data, it is important to consider the spatial context of time-series at every location to identify changes in both space and time. 
A recent work by \cite{chen2013contextual} defined a new type of change called \textit{contextual change}, where a time series is considered to be changing if it deviates from other time series in its local context. The context of a time series can be defined in a number of ways, e.g., by considering the group of time series that are similar to the given time series for a period of time, or the time series observed at locations in close spatial vicinity. 
Another generalization of change detection problems in ST applications is to determine the spatial extent and temporal window in an ST raster where a change has likely manifested. For example, the set of geographic locations and time intervals where loss in vegetation occurred due to deforestation in a forest area is of interest to ecologists. Recent work has explored some efficient approaches to capture specific type of events in global vegetation index data. An efficient approach that enumerates and prunes candidate spatio-temporal windows has been proposed by \cite{zhou2013discovering}. This approach is referred to as a space-time window enumeration and pruning (SWEP) approach and was shown to be efficient for discovering the space and time subspaces where there is a consistent decrease in vegetation. \cite{zhou2011discovering} defined change detection in the context of a path (i.e., a highway or a longitude), where their goal is to determine sub-paths where abrupt changes are seen. They proposed a a sub-path enumeration and pruning (SEP) approach and have shown its promise in discovering abrupt changes in vegetation across longitudes in Africa.   

%For the problem of discovering paths (e.g., longitudinal) along which there is a significant change in a spatio-temporal variable (e.g., longitudinal paths with abrupt changes in vegetation),  was shown to be efficient . 

Approaches that are relevant to spatio-temporal change detection problem have also been studied in the domain of image analysis where the objective is to determine objects and segments of videos where they are present  \cite{lhermitte2008hierarchical,grundmann2010efficient,moscheni1998spatio}. Suitability of these approaches to spatio-temporal data such as remote sensing and magnetic resonance imaging is yet to be explored. 
A taxonomy of possible generalization of `change' in spatio-temporal data is provided by \cite{zhou2014spatiotemporal}, where they discuss applications that require the spatial extent of change to be defined as a point, line-segment, polygon or a network, and the temporal extent of changed to be defined as a time point or a time interval.

\subsection{Relationship Mining}

\begin{figure}
% \vspace{-0.35in}
\centering
\includegraphics[width=0.6\textwidth]{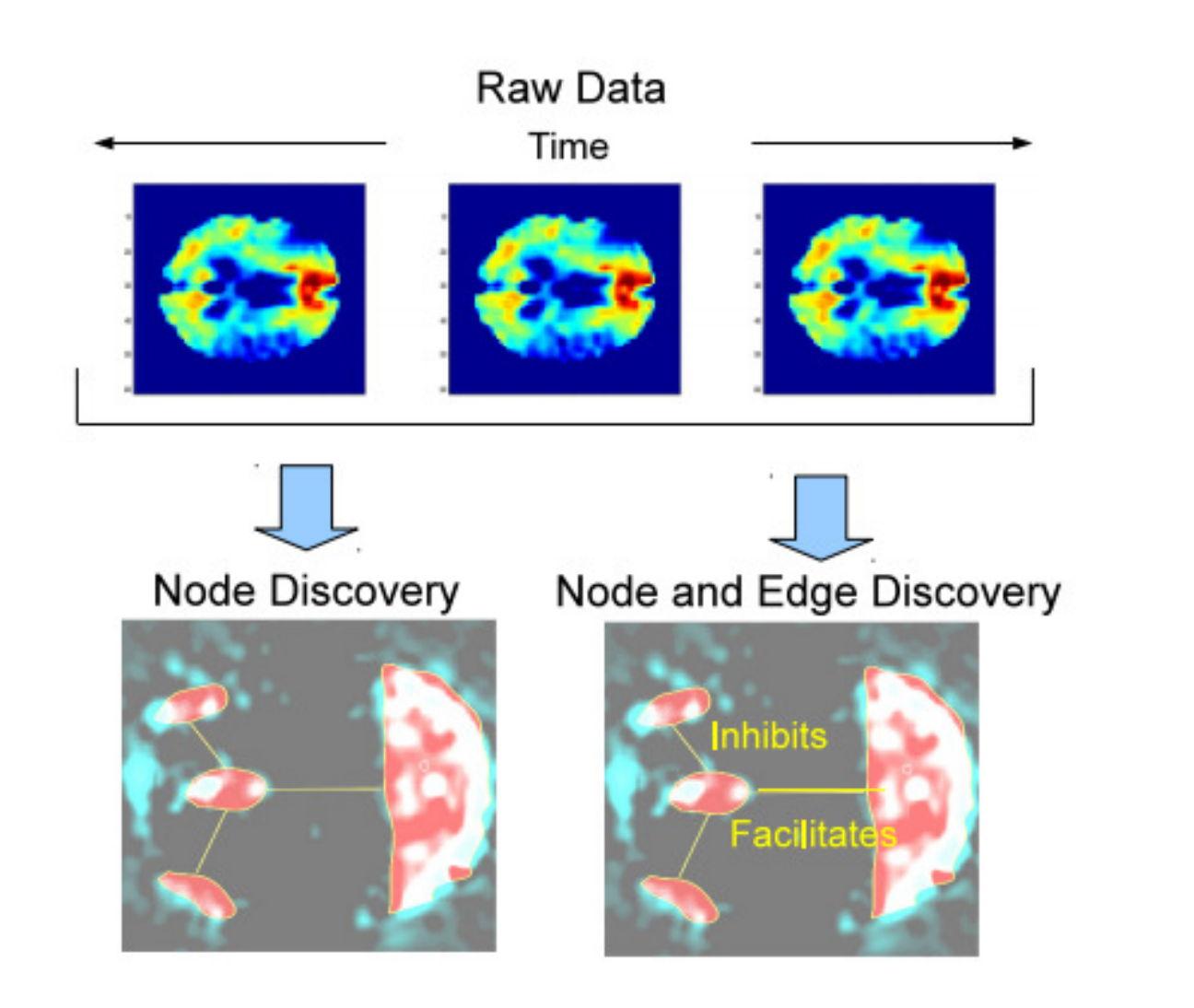}
\caption{Example of a set of nodes and edges that are simultaneously discovered using a tensor decomposition of spatio-temporal data proposed in \cite{davidson2013network}. (Figure taken from \cite{davidson2013network})}
\label{fig:net_discovery}
% \vspace{-0.1in}
\end{figure}

In the domain of time series data mining, relationships among pairs of time series can be discovered using any of the similarity measures defined over time series instances as described in Section \ref{sec:ts_sim}. 
However, in ST applications where the time series are also associated with a spatial information about the location of their observation, relationships in ST data can be defined in numerous ways. One type of relationship is defined as a similarity/dissimilarity between two distant groups of contiguous locations. 
Such long-range relationships are of interest in a number of applications such as climate science and brain imaging. For example, while positive correlations have generally been studied in brain imaging \cite{bassett2006small}, negative correlations (also referred to as teleconnections) among distant locations have been studied in climate datasets \cite{yang2012systematic}.
% However, due to the spatial autocorrelation in the data, a set of contiguous locations tends to exhibit similarity with another distant set of contiguous locations. This will result in multiple redundant `long-distance similarities'. 
In such problems, the goal is to determine both the sets of contiguous locations, not just a pair of locations, that participate in a relationship. 
Hence, it is important to discover the pairs of regions as well as their relationships simultaneously.
% For example, in the analysis of brain fMRI data, time series from locations in prefrontal and visual areas located at the front and the back of the head, respectively, are known to be highly correlated when the subject is resting. 
This is because if the regions are discovered independently of the relationships between them, e.g., using a clustering algorithm, we may miss a number of relationships whose one or both regions fall at the boundary of the clusters.  \cite{davidson2013network} designed a tensor-based approach to discover regions and relationships among them simultaneously, with applications in brain fMRI Data. Figure~\ref{fig:net_discovery} shows the network discovered from the resting-state fMRI data of a healthy subject, which appears to be highly similar to a commonly found and widely studied network known as the \emph{default-mode network} in resting subjects' fMRI scans \cite{raichle2007default}.

\begin{figure}[t]
\centering
\includegraphics[width=0.7\textwidth]{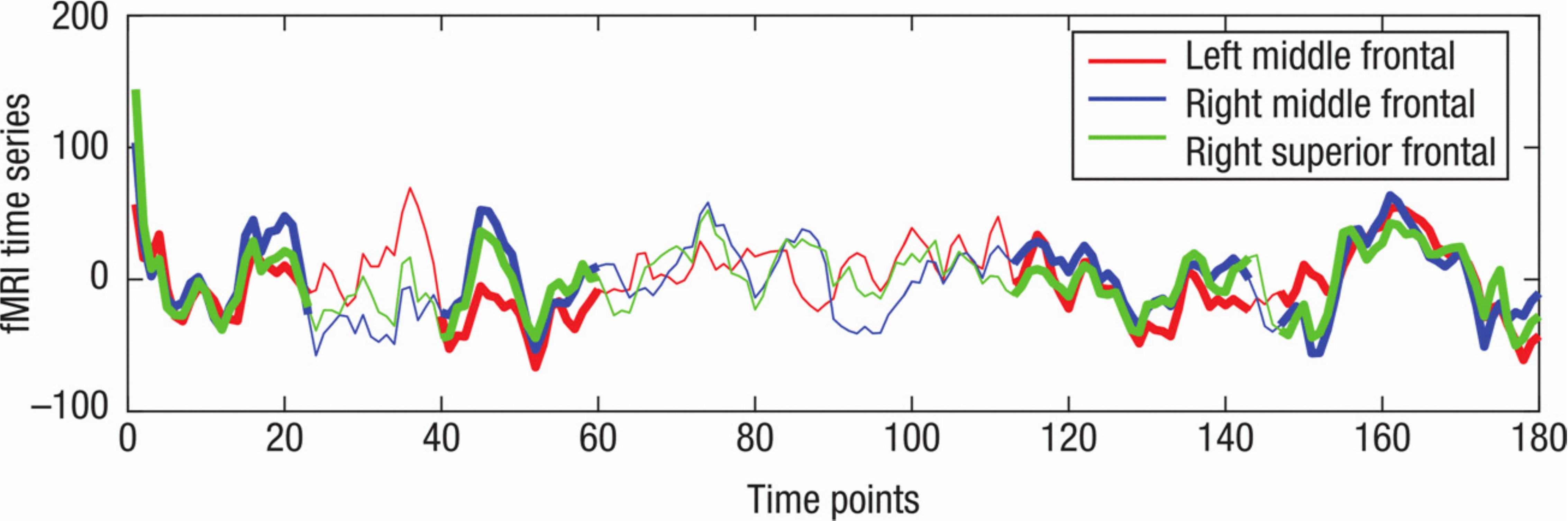}
\caption{Example of a time series pattern that shows three different brain regions exhibiting high similarity in different time intervals. This patterns is discovered using the pattern mining approach proposed in \cite{atluri2014discovering}. (Figure taken from \cite{atluri2016brain})}
\label{fig:ts_patterns}
% \vspace{-0.25cm}
\end{figure}

An important consideration when mining relationships in ST data is that the strength of relationships among pairs of regions may vary with time. For example, \cite{handwerker2012periodic} demonstrated that the correlation between time series from the Posterior Cingulate region and other locations in the brain change with time. Due to this, it becomes necessary to determine the pair of interacting regions as well as the time window within which they interact. 
% There can be two scenarios here: i) The similarity within each set is invariant with time, but the similarity across the two sets varies with time. ii) The similarity within each set varies with time and so the sets themselves are not valid for studying long distance similarity. In the first case, the sets need to be discovered once for a given data set and the varying long distance similarity needs to be studied with respect to time. 
\cite{atluri2014discovering} designed a pattern mining based approach to study dynamic relationships among time series from different regions (with homogeneous time series). This approach was used in the context of brain fMRI data to discover sets of intermittently synergistic brain regions. An example of such a set is shown in Figure~\ref{fig:ts_patterns}. Another approach was explored in \cite{kawale2011discovering} using graph-based methods for climate science applications, where a new graph was constructed for each time interval and the relationships within a time interval are discovered from the corresponding graph.

%A long-distance similarity could either appear and disappear (scenario (i)) or shift in location (scenario (ii)) as time progresses. In the context of scenario (ii), associating similarity between successive time intervals is needed. In both the scenarios, the evolution of the similarity needs to be captured and patterns in time varying long-distance similarities needs to be discovered to shed light on the principles that govern the underlying systems. 
Another important consideration when discovering relationships in ST data is that the relationships among time-series could exist with time lags. This can be because one region has an influence on the other and the lag is due to the time it takes to conduct the influence through the system. Such relationships are referred to as lagged relationships and they have been studied in several applications such as climate science \cite{chen2011forecasting,lu2016exploring}. A more complex version of lagged relationships is dynamic lagged relationships, where the lagged relationships lasts for a small interval rather than the entire time. In this case the goal is to determine the two sets of locations as well as the intervals in each of them that exhibit the desired similarity. A central challenge in discovering lagged relationships is the increase in the number of degrees of freedom available for finding relationships, which can result in spurious detections unless statistical corrections for multiple hypothesis testing have been carefully performed \cite{bland1995multiple}.

While lagged relationships suggest some type of causal association between the two regions, causal relationships that have been defined in time series data are known to capture truly causal associations \cite{granger1969investigating}. Clive Granger who received Nobel prize for this work defined causality based on the notion that a cause ($x$) can predict the effect ($y$) significantly better than an auto-regressive model of the effect ($y$) itself. This work can be generalized for multivariate time series data where the effect of all variables  on a target variable can be studied. Such an approach does not take into account the causal relationships that could exist between the causal factors themselves. This observation motivated the need to develop Granger causal maps for multivariate time series where the causal relationships between all variables are assessed \cite{eichler2013causal}. These are further extended to capture sparsity \cite{arnold2007temporal} and group structure \cite{lozano2009grouped} in the underlying model. The framework of Pearl causality has also been explored for identifying causal interactions in climate science \cite{ED:2012,ED:2017,causal_attribution}, where probabilistic graphical models were constructed using causal edges among the nodes.

\section{Conclusions and Future Work}

In this article, we attempted to provide a broad overview of the field of spatio-temporal data mining that covers research spanning more than several decades. However, given the vast literature on this topic and space limitations, we have only been able to cover a small fraction of work in this fast growing area of research. Still our hope is that this article provides the right foundation for consolidating the rich and diverse literature on STDM research, that has been explored in different application contexts for different types of ST data, under a common over-arching structure. The organization of STDM research presented in this article can be used as a stepping stone for future reviews on this topic that provide detailed empirical comparisons of different STDM methods, e.g., in terms of accuracy, efficiency, and scalability. While such evaluations have been performed in limited contexts for specific topics, e.g., for trajectory pattern mining \cite{mamoulis2004mining,giannotti2007trajectory} and time-series data mining \cite{ding2008querying,esling2012time}, an extensive evaluation covering the full breadth of STDM problems and methods is missing.

We also hope that this article helps the reader in identifying the relevant set of questions that can be posed for a given ST data set, and the right choice of methods for pursuing a research question in STDM. This can be particularly helpful in translating STDM research ideas developed in one community to another, so that new problems in novel application domains can be solved. For example, in flow dynamics, data pertaining to the breakup of liquid droplets is generated using simulations where droplets are present in some spatial locations in arbitrary shapes. The goal in these studies is to study droplet dynamics, i.e., changes in shape and position of the droplets with time under various flow and ambient conditions. Such patterns can be discovered using approaches for finding moving clusters of migratory animals, discussed in Section \ref{sec:clus_trajectories} in the context of clustering trajectories.
% Similarly in aerospace engineering applications \cite{dong2016smoothed}, data from different locations within the propulsion system are collected over time to model the wear and tear of the system.  

The variety of data types, problems, and methods, and the range of emerging application areas where ST data is being increasingly collected and new scientific questions explored  makes spatio-temporal data mining a quintessential melting pot for new research in data mining. 
One of the major emerging themes in STDM research is of studying novel representations of ST raster data sets. 
% In some application domains such as brain imaging and climate science there has been tremendous interest in transforming the spatio-temporal raster data into a network \cite{atluri2016brain,feldhoff2015complex}. Particularly in the case of brain imaging, where the data is collected from a large neural network, the brain, a network representation is naturally suited. 
Most recent work in this direction involves defining novel types of edges or relationships between spatial entities and developing effective approaches for discovering them \cite{agrawal2017tripoles}. Much of the work in this area focuses on capturing `static' edges, while there is increasing recognition that `dynamic' edges are better suited for representing ST raster data. Approaches for effectively capturing such edges in spatial-temporal varying graphs have been explored in \cite{chen2010learning} for applications in climate science. 
% Studying effective representation schemes is also relevant beyond ST raster data. 
% In the context of trajectory datasets, different schemes such as semantic trajectories \cite{bogorny2014constant}, symbolic trajectories \cite{guting2015symbolic}, and spatio-textual trajectories \cite{damiani2016spatial} are recently being explored. 

Another emerging direction where there is tremendous interest is to integratively mine multi-modal spatio-temporal datasets. Multi-modal ST datasets are available in domains such as neuroimaging and climate science. In neuroimaging, fMRI and ECG capture the same underlying brain activity using very different technologies that offer different spatial and temporal resolution. On the other hand, in climate science, different variables such as temperature, pressure, humidity, and precipitation are available at the same spatial and temporal resolution. In the former case, the objective of interest is to construct a best possible image of brain activity based on the images from different modalities. Recent work in the form of BrainZoom \cite{fu2017brainzoom} is one of the first efforts in this direction where they developed an optimization based approach to construct the best possible brain activity image based on fMRI and MEG images. In the latter case, the objective of interest is to learn a model of the climate system by taking into account the different climate variables and their inter-relationships \cite{tye2016simulating}. 

Yet another direction that is most relevant in particular to ST data involves tackling the problem of determining the level of granularity or resolution at which a phenomenon of interest needs to be searched for, be it clusters, patterns or anomalies. Most of the existing work in ST data mining overlooks this granularity problem. A closely related contribution is made in the context of network science where it was found that existing community detection approaches might miss important substructures in the network as they over-partition or under-partition the network \cite{fortunato2007resolution}. They also noted that the definition of modularity that these techniques rely upon can have an inherent resolution limit and argued for the need of developing approaches without such limits \cite{fortunato2007resolution}. To address this problem, \cite{delvenne2010stability} proposed an approach to discover clusters that are relevant at different granularities. Similarly, in the context of ST data, there is a need for understanding the inherent limitations of existing approaches in tackling the granularity problem and for developing new approaches that can overcome such limitations. 

Since many problems in physical sciences involve the study of processes that are spatio-temporal in nature, e.g., the dynamics of turbulent flow or the evolution of climate states and weather patterns, there is a growing interest to incorporate the scientific basis of such processes in the STDM framework. While traditional STDM methods typically rely only on the information contained in the data, the complex nature of problems and the paucity of observations in scientific applications requires a principled way of integrating data science methods with the wealth of scientific knowledge, often encoded as physics (or theory)-based models. This is the paradigm of theory-guided data science \cite{karpatne2016theory} that is gaining attention in several disciplines for accelerating scientific discovery from data, and is particularly relevant for STDM applications. Overall, considering the emerging problems and promising research directions, we anticipate the relatively young field of STDM to grow significantly in the next decade.

% Acknowledgments
%\begin{acks}
%This research is supported by 
%\end{acks}

% Bibliography

\bibliographystyle{ACM-Reference-Format}
\bibliography{ref,ref2,ref_anuj}


\begin{thebibliography}{00}

%%% the Web address.

\ifx \showCODEN    \undefined \def \showCODEN     #1{\unskip}     \fi
\ifx \showDOI      \undefined \def \showDOI       #1{#1}\fi
\ifx \showISBNx    \undefined \def \showISBNx     #1{\unskip}     \fi
\ifx \showISBNxiii \undefined \def \showISBNxiii  #1{\unskip}     \fi
\ifx \showISSN     \undefined \def \showISSN      #1{\unskip}     \fi
\ifx \showLCCN     \undefined \def \showLCCN      #1{\unskip}     \fi
\ifx \shownote     \undefined \def \shownote      #1{#1}          \fi
\ifx \showarticletitle \undefined \def \showarticletitle #1{#1}   \fi
\ifx \showURL      \undefined \def \showURL       {\relax}        \fi
% invisible to TeX
\providecommand\bibfield[2]{#2}
\providecommand\bibinfo[2]{#2}
\providecommand\natexlab[1]{#1}
\providecommand\showeprint[2][]{arXiv:#2}

\bibitem[\protect\citeauthoryear{Abdelhaq, Sengstock, and Gertz}{Abdelhaq
  et~al\mbox{.}}{2013}]%
        {abdelhaq2013eventweet}
\bibfield{author}{\bibinfo{person}{H. Abdelhaq} {}et al.}
  \bibinfo{year}{2013}\natexlab{}.
\newblock \showarticletitle{Eventweet: Online localized event detection from
  twitter}.
\newblock \bibinfo{journal}{{\em VLDB\/}} \bibinfo{volume}{6},
  \bibinfo{number}{12}, \bibinfo{pages}{1326--1329}.
\newblock


\bibitem[\protect\citeauthoryear{Agarwal, McGregor, Phillips,
  Venkatasubramanian, and Zhu}{Agarwal et~al\mbox{.}}{2006}]%
        {agarwal2006spatial}
\bibfield{author}{\bibinfo{person}{D. Agarwal} {}et al.}
  \bibinfo{year}{2006}\natexlab{}.
\newblock \showarticletitle{Spatial scan statistics: approximations and
  performance study}. In \bibinfo{booktitle}{{\em SIGKDD}}. ACM,
  \bibinfo{pages}{24--33}.
\newblock


\bibitem[\protect\citeauthoryear{Aggarwal}{Aggarwal}{2015}]%
        {aggarwal2015mining}
\bibfield{author}{\bibinfo{person}{C.~C. Aggarwal}.}
  \bibinfo{year}{2015}\natexlab{}.
\newblock \showarticletitle{Mining Spatial Data}. In \bibinfo{booktitle}{{\em
  Data Mining}}. Springer, \bibinfo{pages}{531--555}.
\newblock


\bibitem[\protect\citeauthoryear{Aggarwal}{Aggarwal}{2017}]%
        {aggarwal2017spatial}
\bibfield{author}{\bibinfo{person}{C.~C. Aggarwal}.}
  \bibinfo{year}{2017}\natexlab{}.
\newblock \showarticletitle{Spatial Outlier Detection}.
\newblock In \bibinfo{booktitle}{{\em Outlier Analysis}}.
  \bibinfo{publisher}{Springer}, \bibinfo{pages}{345--368}.
\newblock


\bibitem[\protect\citeauthoryear{Agrawal and Srikant}{Agrawal and
  Srikant}{1995}]%
        {agrawal1995mining}
\bibfield{author}{\bibinfo{person}{R. Agrawal} {}et al.}
  \bibinfo{year}{1995}\natexlab{}.
\newblock \showarticletitle{Mining sequential patterns}. In
  \bibinfo{booktitle}{{\em ICDE}}. IEEE, \bibinfo{pages}{3--14}.
\newblock


\bibitem[\protect\citeauthoryear{Agrawal, Atluri, Karpatne, Haltom, Liess,
  Chatterjee, and Kumar}{Agrawal et~al\mbox{.}}{2017}]%
        {agrawal2017tripoles}
\bibfield{author}{\bibinfo{person}{S. Agrawal} {}et al.}
  \bibinfo{year}{2017}\natexlab{}.
\newblock \showarticletitle{Tripoles: A New Class of Relationships in Time
  Series Data}. In \bibinfo{booktitle}{{\em KDD}}. ACM,
  \bibinfo{pages}{697--706}.
\newblock


\bibitem[\protect\citeauthoryear{Al-Naymat, Chawla, and Gudmundsson}{Al-Naymat
  et~al\mbox{.}}{2007}]%
        {al2007dimensionality}
\bibfield{author}{\bibinfo{person}{G. Al-Naymat} {}et al.}
  \bibinfo{year}{2007}\natexlab{}.
\newblock \showarticletitle{Dimensionality reduction for long duration and
  complex spatio-temporal queries}. In \bibinfo{booktitle}{{\em Symposium on
  Applied Computing}}. ACM, \bibinfo{pages}{393--397}.
\newblock


\bibitem[\protect\citeauthoryear{Alon, Sclaroff, Kollios, and Pavlovic}{Alon
  et~al\mbox{.}}{2003}]%
        {alon2003discovering}
\bibfield{author}{\bibinfo{person}{J. Alon} {}et al.}
  \bibinfo{year}{2003}\natexlab{}.
\newblock \showarticletitle{Discovering clusters in motion time-series data}.
  In \bibinfo{booktitle}{{\em CVPR}}, Vol.~\bibinfo{volume}{1}. IEEE,
  \bibinfo{pages}{375--381}.
\newblock


\bibitem[\protect\citeauthoryear{Alt and Godau}{Alt and Godau}{1995}]%
        {alt1995computing}
\bibfield{author}{\bibinfo{person}{H. Alt} {}et al.}
  \bibinfo{year}{1995}\natexlab{}.
\newblock \showarticletitle{Computing the Fr{\'e}chet distance between two
  polygonal curves}.
\newblock \bibinfo{journal}{{\em International Journal of Computational
  Geometry \& Applications\/}} \bibinfo{volume}{5}, \bibinfo{number}{01n02}
  (\bibinfo{year}{1995}), \bibinfo{pages}{75--91}.
\newblock


\bibitem[\protect\citeauthoryear{Andreou and Ghysels}{Andreou and
  Ghysels}{2002}]%
        {andreou2002detecting}
\bibfield{author}{\bibinfo{person}{E. Andreou} {}et al.}
  \bibinfo{year}{2002}\natexlab{}.
\newblock \showarticletitle{Detecting multiple breaks in financial market
  volatility dynamics}.
\newblock \bibinfo{journal}{{\em Journal of Applied Econometrics\/}}
  \bibinfo{volume}{17}, \bibinfo{number}{5} (\bibinfo{year}{2002}),
  \bibinfo{pages}{579--600}.
\newblock


\bibitem[\protect\citeauthoryear{Anselin}{Anselin}{1994}]%
        {anselin1994exploratory}
\bibfield{author}{\bibinfo{person}{L. Anselin}.}
  \bibinfo{year}{1994}\natexlab{}.
\newblock \showarticletitle{Exploratory spatial data analysis and geographic
  information systems}.
\newblock \bibinfo{journal}{{\em New tools for spatial analysis\/}}
  \bibinfo{volume}{17} (\bibinfo{year}{1994}), \bibinfo{pages}{45--54}.
\newblock


\bibitem[\protect\citeauthoryear{Anselin}{Anselin}{1995}]%
        {anselin1995local}
\bibfield{author}{\bibinfo{person}{L. Anselin}.}
  \bibinfo{year}{1995}\natexlab{}.
\newblock \showarticletitle{Local indicators of spatial association—LISA}.
\newblock \bibinfo{journal}{{\em Geographical analysis\/}}
  \bibinfo{volume}{27}, \bibinfo{number}{2} (\bibinfo{year}{1995}),
  \bibinfo{pages}{93--115}.
\newblock


\bibitem[\protect\citeauthoryear{Aoki}{Aoki}{2013}]%
        {aoki2013state}
\bibfield{author}{\bibinfo{person}{M. Aoki}.} \bibinfo{year}{2013}\natexlab{}.
\newblock \bibinfo{booktitle}{{\em State space modeling of time series}}.
\newblock \bibinfo{publisher}{Springer Science \& Business Media}.
\newblock


\bibitem[\protect\citeauthoryear{Arnold, Liu, and Abe}{Arnold
  et~al\mbox{.}}{2007}]%
        {arnold2007temporal}
\bibfield{author}{\bibinfo{person}{A. Arnold} {}et al.}
  \bibinfo{year}{2007}\natexlab{}.
\newblock \showarticletitle{Temporal causal modeling with graphical granger
  methods}. In \bibinfo{booktitle}{{\em SIGKDD}}. ACM, \bibinfo{pages}{66--75}.
\newblock


\bibitem[\protect\citeauthoryear{Atluri, MacDonald~III, Lim, and Kumar}{Atluri
  et~al\mbox{.}}{2016}]%
        {atluri2016brain}
\bibfield{author}{\bibinfo{person}{G. Atluri} {}et al.}
  \bibinfo{year}{2016}\natexlab{}.
\newblock \showarticletitle{The Brain-Network Paradigm: Using Functional
  Imaging Data to Study How the Brain Works}.
\newblock \bibinfo{journal}{{\em Computer\/}} \bibinfo{volume}{49},
  \bibinfo{number}{10} (\bibinfo{year}{2016}), \bibinfo{pages}{65--71}.
\newblock


\bibitem[\protect\citeauthoryear{Atluri, Padmanabhan, Fang, Steinbach,
  Petrella, Lim, MacDonald, Samatova, Doraiswamy, and Kumar}{Atluri
  et~al\mbox{.}}{2013}]%
        {atluri2013complex}
\bibfield{author}{\bibinfo{person}{G. Atluri} {}et al.}
  \bibinfo{year}{2013}\natexlab{}.
\newblock \showarticletitle{Complex biomarker discovery in neuroimaging data{:}
  Finding a needle in a haystack}.
\newblock \bibinfo{journal}{{\em NeuroImage: Clinical\/}}  \bibinfo{volume}{3}
  (\bibinfo{year}{2013}), \bibinfo{pages}{123--131}.
\newblock


\bibitem[\protect\citeauthoryear{Atluri, Steinbach, Lim, MacDonald~III, and
  Kumar}{Atluri et~al\mbox{.}}{2014}]%
        {atluri2014discovering}
\bibfield{author}{\bibinfo{person}{G. Atluri} {}et al.}
  \bibinfo{year}{2014}\natexlab{}.
\newblock \showarticletitle{Discovering Groups of Time Series with Similar
  Behavior in Multiple Small Intervals of Time}. In \bibinfo{booktitle}{{\em
  SIAM International Conference on Data Mining}}.
\newblock


\bibitem[\protect\citeauthoryear{Atluri, Steinbach, Lim, Kumar, and
  MacDonald}{Atluri et~al\mbox{.}}{2015}]%
        {atluri2015connectivity}
\bibfield{author}{\bibinfo{person}{G. Atluri} {}et al.}
  \bibinfo{year}{2015}\natexlab{}.
\newblock \showarticletitle{Connectivity cluster analysis for discovering
  discriminative subnetworks in schizophrenia}.
\newblock \bibinfo{journal}{{\em Human brain mapping\/}} \bibinfo{volume}{36},
  \bibinfo{number}{2} (\bibinfo{year}{2015}), \bibinfo{pages}{756--767}.
\newblock


\bibitem[\protect\citeauthoryear{Bahadori, Yu, and Liu}{Bahadori
  et~al\mbox{.}}{2014}]%
        {bahadori2014fast}
\bibfield{author}{\bibinfo{person}{M. Bahadori} {}et al.}
  \bibinfo{year}{2014}\natexlab{}.
\newblock \showarticletitle{Fast multivariate spatio-temporal analysis via low
  rank tensor learning}. In \bibinfo{booktitle}{{\em NIPS}}.
  \bibinfo{pages}{3491--3499}.
\newblock


\bibitem[\protect\citeauthoryear{Baragona and Battaglia}{Baragona and
  Battaglia}{2007}]%
        {baragona2007outliers}
\bibfield{author}{\bibinfo{person}{R. Baragona} {}et al.}
  \bibinfo{year}{2007}\natexlab{}.
\newblock \showarticletitle{Outliers detection in multivariate time series by
  independent component analysis}.
\newblock \bibinfo{journal}{{\em Neural computation\/}} \bibinfo{volume}{19},
  \bibinfo{number}{7} (\bibinfo{year}{2007}), \bibinfo{pages}{1962--1984}.
\newblock


\bibitem[\protect\citeauthoryear{Bassett and Bullmore}{Bassett and
  Bullmore}{2006}]%
        {bassett2006small}
\bibfield{author}{\bibinfo{person}{D. Bassett} {}et al.}
  \bibinfo{year}{2006}\natexlab{}.
\newblock \showarticletitle{Small-world brain networks}.
\newblock \bibinfo{journal}{{\em Neuroscientist\/}} \bibinfo{volume}{12},
  \bibinfo{number}{6} (\bibinfo{year}{2006}), \bibinfo{pages}{512--523}.
\newblock


\bibitem[\protect\citeauthoryear{Bellec, Perlbarg, Jbabdi,
  P{\'e}l{\'e}grini-Issac, Anton, Doyon, and Benali}{Bellec
  et~al\mbox{.}}{2006}]%
        {bellec2006identification}
\bibfield{author}{\bibinfo{person}{P. Bellec} {}et al.}
  \bibinfo{year}{2006}\natexlab{}.
\newblock \showarticletitle{Identification of large-scale networks in the brain
  using fMRI}.
\newblock \bibinfo{journal}{{\em Neuroimage\/}} \bibinfo{volume}{29},
  \bibinfo{number}{4} (\bibinfo{year}{2006}), \bibinfo{pages}{1231--1243}.
\newblock


\bibitem[\protect\citeauthoryear{Berg and L{\"a}ssig}{Berg and
  L{\"a}ssig}{2006}]%
        {berg2006cross}
\bibfield{author}{\bibinfo{person}{J. Berg} {}et al.}
  \bibinfo{year}{2006}\natexlab{}.
\newblock \showarticletitle{Cross-species analysis of biological networks by
  Bayesian alignment}.
\newblock \bibinfo{journal}{{\em PNAS\/}} \bibinfo{volume}{103},
  \bibinfo{number}{29}, \bibinfo{pages}{10967--10972}.
\newblock


\bibitem[\protect\citeauthoryear{Bernaola-Galv{\'a}n, Ivanov, Amaral, and
  Stanley}{Bernaola-Galv{\'a}n et~al\mbox{.}}{2001}]%
        {bernaola2001scale}
\bibfield{author}{\bibinfo{person}{P. Bernaola-Galv{\'a}n} {}et al.}
  \bibinfo{year}{2001}\natexlab{}.
\newblock \showarticletitle{Scale invariance in the nonstationarity of human
  heart rate}.
\newblock \bibinfo{journal}{{\em PRL\/}} \bibinfo{volume}{87},
  \bibinfo{number}{16} (\bibinfo{year}{2001}), \bibinfo{pages}{168105}.
\newblock


\bibitem[\protect\citeauthoryear{Birant and Kut}{Birant and Kut}{2007}]%
        {birant2007st}
\bibfield{author}{\bibinfo{person}{D. Birant} {}et al.}
  \bibinfo{year}{2007}\natexlab{}.
\newblock \showarticletitle{ST-DBSCAN: An algorithm for clustering
  spatial--temporal data}.
\newblock \bibinfo{journal}{{\em Data and Knowledge Engineering\/}}
  \bibinfo{volume}{60}, \bibinfo{number}{1} (\bibinfo{year}{2007}),
  \bibinfo{pages}{208--221}.
\newblock


\bibitem[\protect\citeauthoryear{Bishop}{Bishop}{1994}]%
        {bishop1994novelty}
\bibfield{author}{\bibinfo{person}{C. Bishop}.}
  \bibinfo{year}{1994}\natexlab{}.
\newblock \showarticletitle{Novelty detection and neural network validation}.
  In \bibinfo{booktitle}{{\em Vision, Image and Signal Processing}},
  Vol.~\bibinfo{volume}{141}. IET, \bibinfo{pages}{217--222}.
\newblock


\bibitem[\protect\citeauthoryear{Bland and Altman}{Bland and Altman}{1995}]%
        {bland1995multiple}
\bibfield{author}{\bibinfo{person}{J.~M. Bland} {}et al.}
  \bibinfo{year}{1995}\natexlab{}.
\newblock \showarticletitle{Multiple significance tests: the Bonferroni
  method}.
\newblock \bibinfo{journal}{{\em Bmj\/}} \bibinfo{volume}{310},
  \bibinfo{number}{6973} (\bibinfo{year}{1995}), \bibinfo{pages}{170}.
\newblock


\bibitem[\protect\citeauthoryear{Blumensath, Jbabdi, Glasser, Van~Essen,
  Ugurbil, Behrens, and Smith}{Blumensath et~al\mbox{.}}{2013}]%
        {blumensath2013spatially}
\bibfield{author}{\bibinfo{person}{T. Blumensath} {}et al.}
  \bibinfo{year}{2013}\natexlab{}.
\newblock \showarticletitle{Spatially constrained hierarchical parcellation of
  the brain with resting-state fMRI}.
\newblock \bibinfo{journal}{{\em Neuroimage\/}}  \bibinfo{volume}{76}
  (\bibinfo{year}{2013}), \bibinfo{pages}{313--324}.
\newblock


\bibitem[\protect\citeauthoryear{Bogorny, Renso, Aquino, Lucca~Siqueira, and
  Alvares}{Bogorny et~al\mbox{.}}{2014}]%
        {bogorny2014constant}
\bibfield{author}{\bibinfo{person}{V. Bogorny} {}et al.}
  \bibinfo{year}{2014}\natexlab{}.
\newblock \showarticletitle{Constant--a conceptual data model for semantic
  trajectories of moving objects}.
\newblock \bibinfo{journal}{{\em Transactions in GIS\/}} \bibinfo{volume}{18},
  \bibinfo{number}{1} (\bibinfo{year}{2014}), \bibinfo{pages}{66--88}.
\newblock


\bibitem[\protect\citeauthoryear{Boriah, Mithal, Garg, Kumar, Steinbach,
  Potter, and Klooster}{Boriah et~al\mbox{.}}{2010}]%
        {boriah2010comparative}
\bibfield{author}{\bibinfo{person}{S. Boriah} {}et al.}
  \bibinfo{year}{2010}\natexlab{}.
\newblock \showarticletitle{A Comparative Study Of Algorithms For Land Cover
  Change}. In \bibinfo{booktitle}{{\em CIDU}}. \bibinfo{pages}{175--188}.
\newblock


\bibitem[\protect\citeauthoryear{Box and Jenkins}{Box and Jenkins}{1976}]%
        {box1976time}
\bibfield{author}{\bibinfo{person}{G.~E. Box} {}et al.}
  \bibinfo{year}{1976}\natexlab{}.
\newblock \bibinfo{booktitle}{{\em Time series analysis: forecasting and
  control}}.
\newblock \bibinfo{publisher}{Holden-Day}.
\newblock


\bibitem[\protect\citeauthoryear{Brunsdon, Fotheringham, and Charlton}{Brunsdon
  et~al\mbox{.}}{1998}]%
        {brunsdon1998geographically}
\bibfield{author}{\bibinfo{person}{C. Brunsdon} {}et al.}
  \bibinfo{year}{1998}\natexlab{}.
\newblock \showarticletitle{Geographically weighted regression}.
\newblock \bibinfo{journal}{{\em JRSS: D\/}} \bibinfo{volume}{47},
  \bibinfo{number}{3} (\bibinfo{year}{1998}), \bibinfo{pages}{431--443}.
\newblock


\bibitem[\protect\citeauthoryear{Bu, Chen, Fu, and Liu}{Bu
  et~al\mbox{.}}{2009}]%
        {bu2009efficient}
\bibfield{author}{\bibinfo{person}{Y. Bu} {}et al.}
  \bibinfo{year}{2009}\natexlab{}.
\newblock \showarticletitle{Efficient anomaly monitoring over moving object
  trajectory streams}. In \bibinfo{booktitle}{{\em KDD}}.
  \bibinfo{pages}{159--168}.
\newblock


\bibitem[\protect\citeauthoryear{Cao, Mamoulis, and Cheung}{Cao
  et~al\mbox{.}}{2005}]%
        {cao2005mining}
\bibfield{author}{\bibinfo{person}{H. Cao} {}et al.}
  \bibinfo{year}{2005}\natexlab{}.
\newblock \showarticletitle{Mining frequent spatio-temporal sequential
  patterns}. In \bibinfo{booktitle}{{\em ICDM}}. IEEE, \bibinfo{pages}{82--89}.
\newblock


\bibitem[\protect\citeauthoryear{Carney}{Carney}{2016}]%
        {carney2016all}
\bibfield{author}{\bibinfo{person}{N. Carney}.}
  \bibinfo{year}{2016}\natexlab{}.
\newblock \showarticletitle{All Lives Matter, but so does race: Black Lives
  Matter and the evolving role of social media}.
\newblock \bibinfo{journal}{{\em Humanity \& Society\/}} \bibinfo{volume}{40},
  \bibinfo{number}{2} (\bibinfo{year}{2016}), \bibinfo{pages}{180--199}.
\newblock


\bibitem[\protect\citeauthoryear{Carpena and Bernaola-Galv{\'a}n}{Carpena and
  Bernaola-Galv{\'a}n}{1999}]%
        {carpena1999statistical}
\bibfield{author}{\bibinfo{person}{P. Carpena} {}et al.}
  \bibinfo{year}{1999}\natexlab{}.
\newblock \showarticletitle{Statistical characterization of the mobility edge
  of vibrational states in disordered materials}.
\newblock \bibinfo{journal}{{\em Physical Review B\/}} \bibinfo{volume}{60},
  \bibinfo{number}{1} (\bibinfo{year}{1999}), \bibinfo{pages}{201}.
\newblock


\bibitem[\protect\citeauthoryear{Castro, Zhang, Chen, Li, and Pan}{Castro
  et~al\mbox{.}}{2013}]%
        {castro2013taxi}
\bibfield{author}{\bibinfo{person}{P.~S. Castro} {}et al.}
  \bibinfo{year}{2013}\natexlab{}.
\newblock \showarticletitle{From taxi GPS traces to social and community
  dynamics: A survey}.
\newblock \bibinfo{journal}{{\em ACM Computing Surveys (CSUR)\/}}
  \bibinfo{volume}{46}, \bibinfo{number}{2} (\bibinfo{year}{2013}),
  \bibinfo{pages}{17}.
\newblock


\bibitem[\protect\citeauthoryear{Chandola and Vatsavai}{Chandola and
  Vatsavai}{2011}]%
        {chandola2011scalable}
\bibfield{author}{\bibinfo{person}{V. Chandola} {}et al.}
  \bibinfo{year}{2011}\natexlab{}.
\newblock \showarticletitle{A scalable gaussian process analysis algorithm for
  biomass monitoring}.
\newblock \bibinfo{journal}{{\em Statistical Analysis and Data Mining: The ASA
  Data Science Journal\/}} \bibinfo{volume}{4}, \bibinfo{number}{4}
  (\bibinfo{year}{2011}), \bibinfo{pages}{430--445}.
\newblock


\bibitem[\protect\citeauthoryear{Chandola, Vatsavai, Kumar, and
  Ganguly}{Chandola et~al\mbox{.}}{2015}]%
        {chandola2015analyzing}
\bibfield{author}{\bibinfo{person}{V. Chandola} {}et al.}
  \bibinfo{year}{2015}\natexlab{}.
\newblock \showarticletitle{Analyzing big spa-tial and big spatiotemporal data:
  a case study of methods and ap-plications}.
\newblock \bibinfo{journal}{{\em Big Data Analytics\/}}  \bibinfo{volume}{33}
  (\bibinfo{year}{2015}), \bibinfo{pages}{239}.
\newblock


\bibitem[\protect\citeauthoryear{Chen, Cheng, Jiang, and Yoshihira}{Chen
  et~al\mbox{.}}{2008}]%
        {chen2008exploiting}
\bibfield{author}{\bibinfo{person}{H. Chen} {}et al.}
  \bibinfo{year}{2008}\natexlab{}.
\newblock \showarticletitle{Exploiting local and global invariants for the
  management of large scale information systems}. In \bibinfo{booktitle}{{\em
  ICDM}}. IEEE, \bibinfo{pages}{113--122}.
\newblock


\bibitem[\protect\citeauthoryear{Chen, Liu, Liu, and Carbonell}{Chen
  et~al\mbox{.}}{2010}]%
        {chen2010learning}
\bibfield{author}{\bibinfo{person}{X. Chen} {}et al.}
  \bibinfo{year}{2010}\natexlab{}.
\newblock \showarticletitle{Learning Spatial-Temporal Varying Graphs with
  Applications to Climate Data Analysis.}. In \bibinfo{booktitle}{{\em AAAI}}.
\newblock


\bibitem[\protect\citeauthoryear{Chen, Faghmous, Khandelwal, and Kumar}{Chen
  et~al\mbox{.}}{2015}]%
        {chen2015clustering}
\bibfield{author}{\bibinfo{person}{X.~C. Chen} {}et al.}
  \bibinfo{year}{2015}\natexlab{}.
\newblock \showarticletitle{Clustering Dynamic Spatio-Temporal Patterns in The
  Presence of Noise and Missing Data.}. In \bibinfo{booktitle}{{\em IJCAI}}.
  \bibinfo{pages}{2575--2581}.
\newblock


\bibitem[\protect\citeauthoryear{Chen, Steinhaeuser, Boriah, Chatterjee, and
  Kumar}{Chen et~al\mbox{.}}{2013}]%
        {chen2013contextual}
\bibfield{author}{\bibinfo{person}{X.~C. Chen} {}et al.}
  \bibinfo{year}{2013}\natexlab{}.
\newblock \showarticletitle{Contextual Time Series Change Detection}. In
  \bibinfo{booktitle}{{\em SDM}}. SIAM, \bibinfo{pages}{503--511}.
\newblock


\bibitem[\protect\citeauthoryear{Chen, Randerson, Morton, DeFries, Collatz,
  Kasibhatla, Giglio, Jin, and Marlier}{Chen et~al\mbox{.}}{2011}]%
        {chen2011forecasting}
\bibfield{author}{\bibinfo{person}{Y. Chen} {}et al.}
  \bibinfo{year}{2011}\natexlab{}.
\newblock \showarticletitle{Forecasting fire season severity in South America
  using sea surface temperature anomalies}.
\newblock \bibinfo{journal}{{\em Science\/}} \bibinfo{volume}{334},
  \bibinfo{number}{6057} (\bibinfo{year}{2011}), \bibinfo{pages}{787--791}.
\newblock


\bibitem[\protect\citeauthoryear{Chen, Wang, and Chen}{Chen
  et~al\mbox{.}}{2016}]%
        {chen2016mining}
\bibfield{author}{\bibinfo{person}{Y.~C. Chen} {}et al.}
  \bibinfo{year}{2016}\natexlab{}.
\newblock \showarticletitle{Mining User Trajectories from Smartphone Data
  Considering Data Uncertainty}. In \bibinfo{booktitle}{{\em International
  Conference on Big Data Analytics and Knowledge Discovery}}. Springer,
  \bibinfo{pages}{51--67}.
\newblock


\bibitem[\protect\citeauthoryear{Cheng, Haworth, Anbaroglu, Tanaksaranond, and
  Wang}{Cheng et~al\mbox{.}}{2014}]%
        {cheng2014spatiotemporal}
\bibfield{author}{\bibinfo{person}{T. Cheng} {}et al.}
  \bibinfo{year}{2014}\natexlab{}.
\newblock \showarticletitle{Spatiotemporal data mining}.
\newblock In \bibinfo{booktitle}{{\em Handbook of Regional Science}}.
  \bibinfo{pages}{1173--1193}.
\newblock


\bibitem[\protect\citeauthoryear{Cheng and Li}{Cheng and Li}{2004}]%
        {cheng2004hybrid}
\bibfield{author}{\bibinfo{person}{T. Cheng} {}et al.}
  \bibinfo{year}{2004}\natexlab{}.
\newblock \showarticletitle{A hybrid approach to detect spatial-temporal
  outliers}. In \bibinfo{booktitle}{{\em Intl. Conf. on Geoinformatics
  Research}}. \bibinfo{pages}{173--178}.
\newblock


\bibitem[\protect\citeauthoryear{Cheng and Li}{Cheng and Li}{2006}]%
        {cheng2006multiscale}
\bibfield{author}{\bibinfo{person}{T. Cheng} {}et al.}
  \bibinfo{year}{2006}\natexlab{}.
\newblock \showarticletitle{A multiscale approach for spatio-temporal outlier
  detection}.
\newblock \bibinfo{journal}{{\em TGIS\/}} \bibinfo{volume}{10},
  \bibinfo{number}{2} (\bibinfo{year}{2006}), \bibinfo{pages}{253--263}.
\newblock


\bibitem[\protect\citeauthoryear{Cheng and Wicks}{Cheng and Wicks}{2014}]%
        {cheng2014event}
\bibfield{author}{\bibinfo{person}{T. Cheng} {}et al.}
  \bibinfo{year}{2014}\natexlab{}.
\newblock \showarticletitle{Event detection using Twitter: a spatio-temporal
  approach}.
\newblock \bibinfo{journal}{{\em PloS one\/}} \bibinfo{volume}{9},
  \bibinfo{number}{6} (\bibinfo{year}{2014}), \bibinfo{pages}{e97807}.
\newblock


\bibitem[\protect\citeauthoryear{Chierichetti, Kleinberg, Kumar, Mahdian, and
  Pandey}{Chierichetti et~al\mbox{.}}{2014}]%
        {chierichetti2014event}
\bibfield{author}{\bibinfo{person}{F. Chierichetti} {}et al.}
  \bibinfo{year}{2014}\natexlab{}.
\newblock \showarticletitle{Event Detection via Communication Pattern
  Analysis.}. In \bibinfo{booktitle}{{\em ICWSM}}.
\newblock


\bibitem[\protect\citeauthoryear{Chiu, Keogh, and Lonardi}{Chiu
  et~al\mbox{.}}{2003}]%
        {chiu2003probabilistic}
\bibfield{author}{\bibinfo{person}{B. Chiu} {}et al.}
  \bibinfo{year}{2003}\natexlab{}.
\newblock \showarticletitle{Probabilistic discovery of time series motifs}. In
  \bibinfo{booktitle}{{\em SIGKDD}}. ACM, \bibinfo{pages}{493--498}.
\newblock


\bibitem[\protect\citeauthoryear{Chudova, Gaffney, Mjolsness, and
  Smyth}{Chudova et~al\mbox{.}}{2003}]%
        {chudova2003translation}
\bibfield{author}{\bibinfo{person}{D. Chudova} {}et al.}
  \bibinfo{year}{2003}\natexlab{}.
\newblock \showarticletitle{Translation-invariant mixture models for curve
  clustering}. In \bibinfo{booktitle}{{\em KDD}}. \bibinfo{pages}{79--88}.
\newblock


\bibitem[\protect\citeauthoryear{Craddock, James, Holtzheimer, Hu, and
  Mayberg}{Craddock et~al\mbox{.}}{2012}]%
        {craddock2012whole}
\bibfield{author}{\bibinfo{person}{R.~C. Craddock} {}et al.}
  \bibinfo{year}{2012}\natexlab{}.
\newblock \showarticletitle{A whole brain fMRI atlas generated via spatially
  constrained spectral clustering}.
\newblock \bibinfo{journal}{{\em Human brain mapping\/}} \bibinfo{volume}{33},
  \bibinfo{number}{8} (\bibinfo{year}{2012}), \bibinfo{pages}{1914--1928}.
\newblock


\bibitem[\protect\citeauthoryear{Cressie and Wikle}{Cressie and Wikle}{2015}]%
        {cressie2015statistics}
\bibfield{author}{\bibinfo{person}{N. Cressie} {}et al.}
  \bibinfo{year}{2015}\natexlab{}.
\newblock \bibinfo{booktitle}{{\em Statistics for spatio-temporal data}}.
\newblock \bibinfo{publisher}{John Wiley \& Sons}.
\newblock


\bibitem[\protect\citeauthoryear{Culotta}{Culotta}{2010}]%
        {culotta2010towards}
\bibfield{author}{\bibinfo{person}{A. Culotta}.}
  \bibinfo{year}{2010}\natexlab{}.
\newblock \showarticletitle{Towards detecting influenza epidemics by analyzing
  Twitter messages}. In \bibinfo{booktitle}{{\em Proceedings of the first
  workshop on social media analytics}}. ACM, \bibinfo{pages}{115--122}.
\newblock


\bibitem[\protect\citeauthoryear{Damiani}{Damiani}{2016}]%
        {damiani2016spatial}
\bibfield{author}{\bibinfo{person}{M.~L. Damiani}.}
  \bibinfo{year}{2016}\natexlab{}.
\newblock \showarticletitle{Spatial trajectories segmentation: trends and
  challenges}. In \bibinfo{booktitle}{{\em Proceedings of the 5th ACM
  SIGSPATIAL International Workshop on Mobile Geographic Information Systems}}.
  ACM, \bibinfo{pages}{1--1}.
\newblock


\bibitem[\protect\citeauthoryear{Davidson, Gilpin, Carmichael, and
  Walker}{Davidson et~al\mbox{.}}{2013}]%
        {davidson2013network}
\bibfield{author}{\bibinfo{person}{I. Davidson} {}et al.}
  \bibinfo{year}{2013}\natexlab{}.
\newblock \showarticletitle{Network discovery via constrained tensor analysis
  of fmri data}. In \bibinfo{booktitle}{{\em KDD}}. \bibinfo{pages}{194--202}.
\newblock


\bibitem[\protect\citeauthoryear{De~Martino, Gentile, Esposito, Balsi,
  Di~Salle, Goebel, and Formisano}{De~Martino et~al\mbox{.}}{2007}]%
        {de2007classification}
\bibfield{author}{\bibinfo{person}{F. De~Martino} {}et al.}
  \bibinfo{year}{2007}\natexlab{}.
\newblock \showarticletitle{Classification of fMRI independent components using
  IC-fingerprints and support vector machine classifiers}.
\newblock \bibinfo{journal}{{\em Neuroimage\/}} \bibinfo{volume}{34},
  \bibinfo{number}{1} (\bibinfo{year}{2007}), \bibinfo{pages}{177--194}.
\newblock


\bibitem[\protect\citeauthoryear{DeFries and Chan}{DeFries and Chan}{2000}]%
        {defries2000multiple}
\bibfield{author}{\bibinfo{person}{R. DeFries} {}et al.}
  \bibinfo{year}{2000}\natexlab{}.
\newblock \showarticletitle{Multiple criteria for evaluating machine learning
  algorithms for land cover classification from satellite data}.
\newblock \bibinfo{journal}{{\em Remote Sensing of Environment\/}}
  \bibinfo{volume}{74}, \bibinfo{number}{3} (\bibinfo{year}{2000}),
  \bibinfo{pages}{503--515}.
\newblock


\bibitem[\protect\citeauthoryear{Delvenne, Yaliraki, and Barahona}{Delvenne
  et~al\mbox{.}}{2010}]%
        {delvenne2010stability}
\bibfield{author}{\bibinfo{person}{J.-C. Delvenne} {}et al.}
  \bibinfo{year}{2010}\natexlab{}.
\newblock \showarticletitle{Stability of graph communities across time scales}.
\newblock \bibinfo{journal}{{\em Proceedings of the National Academy of
  Sciences\/}} \bibinfo{volume}{107}, \bibinfo{number}{29}
  (\bibinfo{year}{2010}), \bibinfo{pages}{12755--12760}.
\newblock


\bibitem[\protect\citeauthoryear{Deng, Shahabi, Demiryurek, Zhu, Yu, and
  Liu}{Deng et~al\mbox{.}}{2016}]%
        {deng2016latent}
\bibfield{author}{\bibinfo{person}{D. Deng} {}et al.}
  \bibinfo{year}{2016}\natexlab{}.
\newblock \showarticletitle{Latent space model for road networks to predict
  time-varying traffic}.
\newblock \bibinfo{journal}{{\em arXiv preprint arXiv:1602.04301\/}}
  (\bibinfo{year}{2016}).
\newblock


\bibitem[\protect\citeauthoryear{Diggle}{Diggle}{2013}]%
        {diggle2013statistical}
\bibfield{author}{\bibinfo{person}{P.~J. Diggle}.}
  \bibinfo{year}{2013}\natexlab{}.
\newblock \bibinfo{booktitle}{{\em Statistical analysis of spatial and
  spatio-temporal point patterns}}.
\newblock \bibinfo{publisher}{CRC Press}.
\newblock


\bibitem[\protect\citeauthoryear{Ding, Trajcevski, Scheuermann, Wang, and
  Keogh}{Ding et~al\mbox{.}}{2008}]%
        {ding2008querying}
\bibfield{author}{\bibinfo{person}{H. Ding} {}et al.}
  \bibinfo{year}{2008}\natexlab{}.
\newblock \showarticletitle{Querying and mining of time series data:
  experimental comparison of representations and distance measures}.
\newblock \bibinfo{journal}{{\em Proceedings of the VLDB Endowment\/}}
  \bibinfo{volume}{1}, \bibinfo{number}{2} (\bibinfo{year}{2008}),
  \bibinfo{pages}{1542--1552}.
\newblock


\bibitem[\protect\citeauthoryear{Dixon}{Dixon}{2002}]%
        {dixon2002ripley}
\bibfield{author}{\bibinfo{person}{P.~M. Dixon}.}
  \bibinfo{year}{2002}\natexlab{}.
\newblock \showarticletitle{Ripley's K function}.
\newblock \bibinfo{journal}{{\em Encyclopedia of environmetrics\/}}
  (\bibinfo{year}{2002}).
\newblock


\bibitem[\protect\citeauthoryear{Dodge, Weibel, and Lautensch{\"u}tz}{Dodge
  et~al\mbox{.}}{2008}]%
        {dodge2008towards}
\bibfield{author}{\bibinfo{person}{S. Dodge} {}et al.}
  \bibinfo{year}{2008}\natexlab{}.
\newblock \showarticletitle{Towards a taxonomy of movement patterns}.
\newblock \bibinfo{journal}{{\em Info Vis\/}} \bibinfo{volume}{7},
  \bibinfo{number}{3-4} (\bibinfo{year}{2008}), \bibinfo{pages}{240--252}.
\newblock


\bibitem[\protect\citeauthoryear{Ebert-Uphoff and Deng}{Ebert-Uphoff and
  Deng}{2012}]%
        {ED:2012}
\bibfield{author}{\bibinfo{person}{I. Ebert-Uphoff} {}et al.}
  \bibinfo{year}{2012}\natexlab{}.
\newblock \showarticletitle{Causal discovery for climate research using
  graphical models}.
\newblock \bibinfo{journal}{{\em Journal of Climate\/}} \bibinfo{volume}{25},
  \bibinfo{number}{17} (\bibinfo{year}{2012}), \bibinfo{pages}{5648--5665}.
\newblock


\bibitem[\protect\citeauthoryear{Ebert-Uphoff and Deng}{Ebert-Uphoff and
  Deng}{2017}]%
        {ED:2017}
\bibfield{author}{\bibinfo{person}{I. Ebert-Uphoff} {}et al.}
  \bibinfo{year}{2017}\natexlab{}.
\newblock \showarticletitle{Causal Discovery in the geosciences - Using
  synthetic data to learn how to interpret results}.
\newblock \bibinfo{journal}{{\em Computer \& Geosciences\/}}
  \bibinfo{volume}{99} (\bibinfo{date}{February} \bibinfo{year}{2017}),
  \bibinfo{pages}{50--60}.
\newblock


\bibitem[\protect\citeauthoryear{Eftelioglu, Shekhar, Kang, and
  Farah}{Eftelioglu et~al\mbox{.}}{2016}]%
        {eftelioglu2016ring}
\bibfield{author}{\bibinfo{person}{E. Eftelioglu} {}et al.}
  \bibinfo{year}{2016}\natexlab{}.
\newblock \showarticletitle{Ring-Shaped Hotspot Detection}.
\newblock \bibinfo{journal}{{\em IEEE Transactions on Knowledge and Data
  Engineering\/}} \bibinfo{volume}{28}, \bibinfo{number}{12}
  (\bibinfo{year}{2016}), \bibinfo{pages}{3367--3381}.
\newblock


\bibitem[\protect\citeauthoryear{Eftelioglu, Shekhar, Oliver, Zhou, Evans, Xie,
  Kang, Laubscher, and Farah}{Eftelioglu et~al\mbox{.}}{2014}]%
        {eftelioglu2014ring}
\bibfield{author}{\bibinfo{person}{E. Eftelioglu} {}et al.}
  \bibinfo{year}{2014}\natexlab{}.
\newblock \showarticletitle{Ring-Shaped Hotspot Detection: A Summary of
  Results}. In \bibinfo{booktitle}{{\em ICDM}}. IEEE,
  \bibinfo{pages}{815--820}.
\newblock


\bibitem[\protect\citeauthoryear{Eichler}{Eichler}{2013}]%
        {eichler2013causal}
\bibfield{author}{\bibinfo{person}{M. Eichler}.}
  \bibinfo{year}{2013}\natexlab{}.
\newblock \showarticletitle{Causal inference with multiple time series:
  principles and problems}.
\newblock \bibinfo{journal}{{\em Phil. Trans. of the Royal Society of London A:
  Math., Phys. and Eng. Sciences\/}} \bibinfo{volume}{371},
  \bibinfo{number}{1997} (\bibinfo{year}{2013}), \bibinfo{pages}{20110613}.
\newblock


\bibitem[\protect\citeauthoryear{Eklund, Nichols, and Knutsson}{Eklund
  et~al\mbox{.}}{2016}]%
        {eklund2016cluster}
\bibfield{author}{\bibinfo{person}{A. Eklund} {}et al.}
  \bibinfo{year}{2016}\natexlab{}.
\newblock \showarticletitle{Cluster failure: why fMRI inferences for spatial
  extent have inflated false-positive rates}.
\newblock \bibinfo{journal}{{\em Proceedings of the National Academy of
  Sciences\/}} (\bibinfo{year}{2016}), \bibinfo{pages}{201602413}.
\newblock


\bibitem[\protect\citeauthoryear{Esling and Agon}{Esling and Agon}{2012}]%
        {esling2012time}
\bibfield{author}{\bibinfo{person}{P. Esling} {}et al.}
  \bibinfo{year}{2012}\natexlab{}.
\newblock \showarticletitle{Time-series data mining}.
\newblock \bibinfo{journal}{{\em ACM Computing Surveys (CSUR)\/}}
  \bibinfo{volume}{45}, \bibinfo{number}{1} (\bibinfo{year}{2012}),
  \bibinfo{pages}{12}.
\newblock


\bibitem[\protect\citeauthoryear{Ester, Kriegel, and Sander}{Ester
  et~al\mbox{.}}{1997}]%
        {ester1997spatial}
\bibfield{author}{\bibinfo{person}{M. Ester} {}et al.}
  \bibinfo{year}{1997}\natexlab{}.
\newblock \showarticletitle{Spatial data mining: A database approach}. In
  \bibinfo{booktitle}{{\em ISSD}}. Springer, \bibinfo{pages}{47--66}.
\newblock


\bibitem[\protect\citeauthoryear{Ester, Kriegel, Sander, Xu,
  et~al\mbox{.}}{Ester et~al\mbox{.}}{1996}]%
        {ester1996density}
\bibfield{author}{\bibinfo{person}{M. Ester} {}et al.}
  \bibinfo{year}{1996}\natexlab{}.
\newblock \showarticletitle{A density-based algorithm for discovering clusters
  in large spatial databases with noise}. In \bibinfo{booktitle}{{\em SIGKDD}},
  Vol.~\bibinfo{volume}{96}. \bibinfo{pages}{226--231}.
\newblock


\bibitem[\protect\citeauthoryear{Faghmous, Chamber, Boriah, Vikeb{\o}, Liess,
  dos Santos~Mesquita, and Kumar}{Faghmous et~al\mbox{.}}{2012}]%
        {faghmous2012novel}
\bibfield{author}{\bibinfo{person}{J.~H. Faghmous} {}et al.}
  \bibinfo{year}{2012}\natexlab{}.
\newblock \showarticletitle{A Novel and Scalable Spatio-Temporal Technique for
  Ocean Eddy Monitoring.}. In \bibinfo{booktitle}{{\em AAAI}}.
\newblock


\bibitem[\protect\citeauthoryear{Faghmous, Le, Uluyol, Kumar, and
  Chatterjee}{Faghmous et~al\mbox{.}}{2013a}]%
        {faghmous2013parameter}
\bibfield{author}{\bibinfo{person}{J.~H. Faghmous} {}et al.}
  \bibinfo{year}{2013}\natexlab{a}.
\newblock \showarticletitle{A parameter-free spatio-temporal pattern mining
  model to catalog global ocean dynamics}. In \bibinfo{booktitle}{{\em Data
  Mining (ICDM), 2013 IEEE 13th International Conference on}}. IEEE,
  \bibinfo{pages}{151--160}.
\newblock


\bibitem[\protect\citeauthoryear{Faghmous, Uluyol, Styles, Le, Mithal, Boriah,
  and Kumar}{Faghmous et~al\mbox{.}}{2013b}]%
        {faghmous2013multiple}
\bibfield{author}{\bibinfo{person}{J.~H. Faghmous} {}et al.}
  \bibinfo{year}{2013}\natexlab{b}.
\newblock \showarticletitle{Multiple Hypothesis Object Tracking For
  Unsupervised Self-Learning: An Ocean Eddy Tracking Application}. In
  \bibinfo{booktitle}{{\em AAAI}}.
\newblock


\bibitem[\protect\citeauthoryear{Farine, Strandburg-Peshkin, Berger-Wolf,
  Ziebart, Brugere, Li, and Crofoot}{Farine et~al\mbox{.}}{2016}]%
        {farine2016both}
\bibfield{author}{\bibinfo{person}{D.~R. Farine} {}et al.}
  \bibinfo{year}{2016}\natexlab{}.
\newblock \showarticletitle{Both nearest neighbours and long-term affiliates
  predict individual locations during collective movement in wild baboons}.
\newblock \bibinfo{journal}{{\em Scientific reports\/}}  \bibinfo{volume}{6}
  (\bibinfo{year}{2016}).
\newblock


\bibitem[\protect\citeauthoryear{Feldhoff, Lange, Volkholz, Donges, Kurths, and
  Gerstengarbe}{Feldhoff et~al\mbox{.}}{2015}]%
        {feldhoff2015complex}
\bibfield{author}{\bibinfo{person}{J.~H. Feldhoff} {}et al.}
  \bibinfo{year}{2015}\natexlab{}.
\newblock \showarticletitle{Complex networks for climate model evaluation with
  application to statistical versus dynamical modeling of South American
  climate}.
\newblock \bibinfo{journal}{{\em Climate dynamics\/}} \bibinfo{volume}{44},
  \bibinfo{number}{5-6} (\bibinfo{year}{2015}), \bibinfo{pages}{1567--1581}.
\newblock


\bibitem[\protect\citeauthoryear{Feng, Zhang, Zhang, Han, Wang, Aggarwal, and
  Huang}{Feng et~al\mbox{.}}{2015}]%
        {feng2015streamcube}
\bibfield{author}{\bibinfo{person}{W. Feng} {}et al.}
  \bibinfo{year}{2015}\natexlab{}.
\newblock \showarticletitle{STREAMCUBE: hierarchical spatio-temporal hashtag
  clustering for event exploration over the twitter stream}. In
  \bibinfo{booktitle}{{\em Data Engineering (ICDE), 2015 IEEE 31st
  International Conference on}}. IEEE, \bibinfo{pages}{1561--1572}.
\newblock


\bibitem[\protect\citeauthoryear{Fortunato and Barth{\'e}lemy}{Fortunato and
  Barth{\'e}lemy}{2007}]%
        {fortunato2007resolution}
\bibfield{author}{\bibinfo{person}{S. Fortunato} {}et al.}
  \bibinfo{year}{2007}\natexlab{}.
\newblock \showarticletitle{Resolution limit in community detection}.
\newblock \bibinfo{journal}{{\em Proceedings of the National Academy of
  Sciences\/}} \bibinfo{volume}{104}, \bibinfo{number}{1}
  (\bibinfo{year}{2007}), \bibinfo{pages}{36--41}.
\newblock


\bibitem[\protect\citeauthoryear{Fu}{Fu}{2011}]%
        {fu2011review}
\bibfield{author}{\bibinfo{person}{T.-c. Fu}.} \bibinfo{year}{2011}\natexlab{}.
\newblock \showarticletitle{A review on time series data mining}.
\newblock \bibinfo{journal}{{\em Eng. App. of AI\/}} \bibinfo{volume}{24},
  \bibinfo{number}{1} (\bibinfo{year}{2011}), \bibinfo{pages}{164--181}.
\newblock


\bibitem[\protect\citeauthoryear{Fu, Huang, Stretcu, Song, Papalexakis,
  Talukdar, Mitchell, Sidiropoulo, Faloutsos, and Poczos}{Fu
  et~al\mbox{.}}{2017}]%
        {fu2017brainzoom}
\bibfield{author}{\bibinfo{person}{X. Fu} {}et al.}
  \bibinfo{year}{2017}\natexlab{}.
\newblock \showarticletitle{BrainZoom: High Resolution Reconstruction from
  Multi-modal Brain Signals}. In \bibinfo{booktitle}{{\em Proceedings of the
  2017 SIAM International Conference on Data Mining}}. SIAM,
  \bibinfo{pages}{216--227}.
\newblock


\bibitem[\protect\citeauthoryear{Gaffney and Smyth}{Gaffney and Smyth}{1999}]%
        {gaffney1999trajectory}
\bibfield{author}{\bibinfo{person}{S. Gaffney} {}et al.}
  \bibinfo{year}{1999}\natexlab{}.
\newblock \showarticletitle{Trajectory clustering with mixtures of regression
  models}. In \bibinfo{booktitle}{{\em SIGKDD}}. ACM, \bibinfo{pages}{63--72}.
\newblock


\bibitem[\protect\citeauthoryear{Galeano, Pe{\~n}a, and Tsay}{Galeano
  et~al\mbox{.}}{2006}]%
        {galeano2006outlier}
\bibfield{author}{\bibinfo{person}{P. Galeano} {}et al.}
  \bibinfo{year}{2006}\natexlab{}.
\newblock \showarticletitle{Outlier detection in multivariate time series by
  projection pursuit}.
\newblock \bibinfo{journal}{{\it J. Amer. Statist. Assoc.}}
  \bibinfo{volume}{101}, \bibinfo{number}{474} (\bibinfo{year}{2006}),
  \bibinfo{pages}{654--669}.
\newblock


\bibitem[\protect\citeauthoryear{Gardner}{Gardner}{2006}]%
        {gardner2006exponential}
\bibfield{author}{\bibinfo{person}{E.~S. Gardner}.}
  \bibinfo{year}{2006}\natexlab{}.
\newblock \showarticletitle{Exponential smoothing: The state of the art--Part
  II}.
\newblock \bibinfo{journal}{{\em I J Forecasting\/}} \bibinfo{volume}{22},
  \bibinfo{number}{4}, \bibinfo{pages}{637--666}.
\newblock


\bibitem[\protect\citeauthoryear{Gatrell, Bailey, Diggle, and
  Rowlingson}{Gatrell et~al\mbox{.}}{1996}]%
        {gatrell1996spatial}
\bibfield{author}{\bibinfo{person}{A.~C. Gatrell} {}et al.}
  \bibinfo{year}{1996}\natexlab{}.
\newblock \showarticletitle{Spatial point pattern analysis and its application
  in geographical epidemiology}.
\newblock \bibinfo{journal}{{\em Transactions of the Institute of British
  geographers\/}} (\bibinfo{year}{1996}), \bibinfo{pages}{256--274}.
\newblock


\bibitem[\protect\citeauthoryear{Ge, Xiong, Zhou, Ozdemir, Yu, and Lee}{Ge
  et~al\mbox{.}}{2010}]%
        {ge2010top}
\bibfield{author}{\bibinfo{person}{Y. Ge} {}et al.}
  \bibinfo{year}{2010}\natexlab{}.
\newblock \showarticletitle{Top-eye: Top-k evolving trajectory outlier
  detection}. In \bibinfo{booktitle}{{\em CIKM}}. ACM,
  \bibinfo{pages}{1733--1736}.
\newblock


\bibitem[\protect\citeauthoryear{Ghahramani and Hinton}{Ghahramani and
  Hinton}{2000}]%
        {ghahramani2000variational}
\bibfield{author}{\bibinfo{person}{Z. Ghahramani} {}et al.}
  \bibinfo{year}{2000}\natexlab{}.
\newblock \showarticletitle{Variational learning for switching state-space
  models}.
\newblock \bibinfo{journal}{{\em Neural computation\/}} \bibinfo{volume}{12},
  \bibinfo{number}{4} (\bibinfo{year}{2000}), \bibinfo{pages}{831--864}.
\newblock


\bibitem[\protect\citeauthoryear{Ghosh and Deuser}{Ghosh and Deuser}{1995}]%
        {ghosh1995classification}
\bibfield{author}{\bibinfo{person}{J. Ghosh} {}et al.}
  \bibinfo{year}{1995}\natexlab{}.
\newblock \showarticletitle{Classification of spatio-temporal patterns with
  applications to recognition of sonar sequences}.
\newblock \bibinfo{journal}{{\em Neural Representation of Temporal Patterns\/}}
  (\bibinfo{year}{1995}), \bibinfo{pages}{221--250}.
\newblock


\bibitem[\protect\citeauthoryear{Giannotti, Nanni, Pinelli, and
  Pedreschi}{Giannotti et~al\mbox{.}}{2007}]%
        {giannotti2007trajectory}
\bibfield{author}{\bibinfo{person}{F. Giannotti} {}et al.}
  \bibinfo{year}{2007}\natexlab{}.
\newblock \showarticletitle{Trajectory pattern mining}. In
  \bibinfo{booktitle}{{\em SIGKDD}}. ACM, \bibinfo{pages}{330--339}.
\newblock


\bibitem[\protect\citeauthoryear{Ginsberg, Mohebbi, Patel, Brammer, Smolinski,
  and Brilliant}{Ginsberg et~al\mbox{.}}{2009}]%
        {ginsberg2009detecting}
\bibfield{author}{\bibinfo{person}{J. Ginsberg} {}et al.}
  \bibinfo{year}{2009}\natexlab{}.
\newblock \showarticletitle{Detecting influenza epidemics using search engine
  query data}.
\newblock \bibinfo{journal}{{\em Nature\/}} \bibinfo{volume}{457},
  \bibinfo{number}{7232} (\bibinfo{year}{2009}), \bibinfo{pages}{1012--1014}.
\newblock


\bibitem[\protect\citeauthoryear{Glatman-Freedman, Kaufman, Kopel, Bassal,
  Taran, Valinsky, Agmon, Shpriz, Cohen, Anis, et~al\mbox{.}}{Glatman-Freedman
  et~al\mbox{.}}{2016}]%
        {glatman2016near}
\bibfield{author}{\bibinfo{person}{A. Glatman-Freedman} {}et al.}
  \bibinfo{year}{2016}\natexlab{}.
\newblock \showarticletitle{Near real-time space-time cluster analysis for
  detection of enteric disease outbreaks in a community setting}.
\newblock \bibinfo{journal}{{\em Journal of Infection\/}} \bibinfo{volume}{73},
  \bibinfo{number}{2} (\bibinfo{year}{2016}), \bibinfo{pages}{99--106}.
\newblock


\bibitem[\protect\citeauthoryear{Gomide, Veloso, Meira~Jr, Almeida, Benevenuto,
  Ferraz, and Teixeira}{Gomide et~al\mbox{.}}{2011}]%
        {gomide2011dengue}
\bibfield{author}{\bibinfo{person}{J. Gomide} {}et al.}
  \bibinfo{year}{2011}\natexlab{}.
\newblock \showarticletitle{Dengue surveillance based on a computational model
  of spatio-temporal locality of Twitter}. In \bibinfo{booktitle}{{\em
  Proceedings of the 3rd international web science conference}}. ACM,
  \bibinfo{pages}{3}.
\newblock


\bibitem[\protect\citeauthoryear{Goutte, Toft, Rostrup, Nielsen, and
  Hansen}{Goutte et~al\mbox{.}}{1999}]%
        {goutte1999clustering}
\bibfield{author}{\bibinfo{person}{C. Goutte} {}et al.}
  \bibinfo{year}{1999}\natexlab{}.
\newblock \showarticletitle{On clustering fMRI time series}.
\newblock \bibinfo{journal}{{\em NeuroImage\/}} \bibinfo{volume}{9},
  \bibinfo{number}{3} (\bibinfo{year}{1999}), \bibinfo{pages}{298--310}.
\newblock


\bibitem[\protect\citeauthoryear{Granger}{Granger}{1969}]%
        {granger1969investigating}
\bibfield{author}{\bibinfo{person}{C.~W. Granger}.}
  \bibinfo{year}{1969}\natexlab{}.
\newblock \showarticletitle{Investigating causal relations by econometric
  models and cross-spectral methods}.
\newblock \bibinfo{journal}{{\em Econometrica: Journal of the Econometric
  Society\/}} (\bibinfo{year}{1969}), \bibinfo{pages}{424--438}.
\newblock


\bibitem[\protect\citeauthoryear{Graves and Schmidhuber}{Graves and
  Schmidhuber}{2009}]%
        {graves2009offline}
\bibfield{author}{\bibinfo{person}{A. Graves} {}et al.}
  \bibinfo{year}{2009}\natexlab{}.
\newblock \showarticletitle{Offline handwriting recognition with
  multidimensional recurrent neural networks}. In \bibinfo{booktitle}{{\em
  Advances in neural information processing systems}}.
  \bibinfo{pages}{545--552}.
\newblock


\bibitem[\protect\citeauthoryear{Grosse, Bernaola-Galv{\'a}n, Carpena,
  Rom{\'a}n-Rold{\'a}n, Oliver, and Stanley}{Grosse et~al\mbox{.}}{2002}]%
        {grosse2002analysis}
\bibfield{author}{\bibinfo{person}{I. Grosse} {}et al.}
  \bibinfo{year}{2002}\natexlab{}.
\newblock \showarticletitle{Analysis of symbolic sequences using the
  Jensen-Shannon divergence}.
\newblock \bibinfo{journal}{{\em Physical Review E\/}} \bibinfo{volume}{65},
  \bibinfo{number}{4} (\bibinfo{year}{2002}), \bibinfo{pages}{041905}.
\newblock


\bibitem[\protect\citeauthoryear{Grundmann, Kwatra, Han, and Essa}{Grundmann
  et~al\mbox{.}}{2010}]%
        {grundmann2010efficient}
\bibfield{author}{\bibinfo{person}{M. Grundmann} {}et al.}
  \bibinfo{year}{2010}\natexlab{}.
\newblock \showarticletitle{Efficient hierarchical graph-based video
  segmentation}. In \bibinfo{booktitle}{{\em CVPR}}.
  \bibinfo{pages}{2141--2148}.
\newblock


\bibitem[\protect\citeauthoryear{Gupta, Gao, Aggarwal, and Han}{Gupta
  et~al\mbox{.}}{2014}]%
        {gupta2014outlier}
\bibfield{author}{\bibinfo{person}{M. Gupta} {}et al.}
  \bibinfo{year}{2014}\natexlab{}.
\newblock \showarticletitle{Outlier detection for temporal data: A survey}.
\newblock \bibinfo{journal}{{\em TKDE\/}} \bibinfo{volume}{26},
  \bibinfo{number}{9} (\bibinfo{year}{2014}), \bibinfo{pages}{2250--2267}.
\newblock


\bibitem[\protect\citeauthoryear{G{\"u}ting, Vald{\'e}s, and
  Damiani}{G{\"u}ting et~al\mbox{.}}{2015}]%
        {guting2015symbolic}
\bibfield{author}{\bibinfo{person}{R.~H. G{\"u}ting} {}et al.}
  \bibinfo{year}{2015}\natexlab{}.
\newblock \showarticletitle{Symbolic trajectories}.
\newblock \bibinfo{journal}{{\em ACM Transactions on Spatial Algorithms and
  Systems\/}} \bibinfo{volume}{1}, \bibinfo{number}{2} (\bibinfo{year}{2015}),
  \bibinfo{pages}{7}.
\newblock


\bibitem[\protect\citeauthoryear{Handwerker, Roopchansingh, Gonzalez-Castillo,
  and Bandettini}{Handwerker et~al\mbox{.}}{2012}]%
        {handwerker2012periodic}
\bibfield{author}{\bibinfo{person}{D.~A. Handwerker} {}et al.}
  \bibinfo{year}{2012}\natexlab{}.
\newblock \showarticletitle{Periodic changes in fMRI connectivity}.
\newblock \bibinfo{journal}{{\em Neuroimage\/}} \bibinfo{volume}{63},
  \bibinfo{number}{3}, \bibinfo{pages}{1712--1719}.
\newblock


\bibitem[\protect\citeauthoryear{Hannart, Pearl, Otto, Naveau, and
  Ghil}{Hannart et~al\mbox{.}}{2016}]%
        {causal_attribution}
\bibfield{author}{\bibinfo{person}{A. Hannart} {}et al.}
  \bibinfo{year}{2016}\natexlab{}.
\newblock \showarticletitle{Causal Counterfactual Theory for the Attribution of
  Weather and Climate-Related Events}.
\newblock \bibinfo{journal}{{\em Bulletin of the American Meteorological
  Society\/}} (\bibinfo{year}{2016}).
\newblock


\bibitem[\protect\citeauthoryear{Haralick and Shapiro}{Haralick and
  Shapiro}{1985}]%
        {haralick1985image}
\bibfield{author}{\bibinfo{person}{R.~M. Haralick} {}et al.}
  \bibinfo{year}{1985}\natexlab{}.
\newblock \showarticletitle{Image segmentation techniques}.
\newblock \bibinfo{journal}{{\em CVGIP\/}} \bibinfo{volume}{29},
  \bibinfo{number}{1} (\bibinfo{year}{1985}), \bibinfo{pages}{100--132}.
\newblock


\bibitem[\protect\citeauthoryear{Hardisty and Klippel}{Hardisty and
  Klippel}{2010}]%
        {hardisty2010analysing}
\bibfield{author}{\bibinfo{person}{F. Hardisty} {}et al.}
  \bibinfo{year}{2010}\natexlab{}.
\newblock \showarticletitle{Analysing spatio-temporal autocorrelation with
  LISTA-Viz}.
\newblock \bibinfo{journal}{{\em IJGIS\/}} \bibinfo{volume}{24},
  \bibinfo{number}{10}, \bibinfo{pages}{1515--1526}.
\newblock


\bibitem[\protect\citeauthoryear{Haritaoglu, Harwood, and Davis}{Haritaoglu
  et~al\mbox{.}}{2000}]%
        {haritaoglu2000w}
\bibfield{author}{\bibinfo{person}{I. Haritaoglu} {}et al.}
  \bibinfo{year}{2000}\natexlab{}.
\newblock \showarticletitle{W 4: Real-time surveillance of people and their
  activities}.
\newblock \bibinfo{journal}{{\em TPAMI\/}} \bibinfo{volume}{22},
  \bibinfo{number}{8}, \bibinfo{pages}{809--830}.
\newblock


\bibitem[\protect\citeauthoryear{Harvey}{Harvey}{1990}]%
        {harvey1990forecasting}
\bibfield{author}{\bibinfo{person}{A.~C. Harvey}.}
  \bibinfo{year}{1990}\natexlab{}.
\newblock \bibinfo{booktitle}{{\em Forecasting, structural time series models
  and the Kalman filter}}.
\newblock \bibinfo{publisher}{Cambridge U Press}.
\newblock


\bibitem[\protect\citeauthoryear{Haslett, Bradley, Craig, Unwin, and
  Wills}{Haslett et~al\mbox{.}}{1991}]%
        {haslett1991dynamic}
\bibfield{author}{\bibinfo{person}{J. Haslett} {}et al.}
  \bibinfo{year}{1991}\natexlab{}.
\newblock \showarticletitle{Dynamic graphics for exploring spatial data with
  application to locating global and local anomalies}.
\newblock \bibinfo{journal}{{\em The American Statistician\/}}
  \bibinfo{volume}{45}, \bibinfo{number}{3} (\bibinfo{year}{1991}),
  \bibinfo{pages}{234--242}.
\newblock


\bibitem[\protect\citeauthoryear{Heller, Stanley, Yekutieli, Rubin, and
  Benjamini}{Heller et~al\mbox{.}}{2006}]%
        {heller2006cluster}
\bibfield{author}{\bibinfo{person}{R. Heller} {}et al.}
  \bibinfo{year}{2006}\natexlab{}.
\newblock \showarticletitle{Cluster-based analysis of FMRI data}.
\newblock \bibinfo{journal}{{\em NeuroImage\/}} \bibinfo{volume}{33},
  \bibinfo{number}{2} (\bibinfo{year}{2006}), \bibinfo{pages}{599--608}.
\newblock


\bibitem[\protect\citeauthoryear{Hills, Lines, Baranauskas, Mapp, and
  Bagnall}{Hills et~al\mbox{.}}{2014}]%
        {hills2014classification}
\bibfield{author}{\bibinfo{person}{J. Hills} {}et al.}
  \bibinfo{year}{2014}\natexlab{}.
\newblock \showarticletitle{Classification of time series by shapelet
  transformation}.
\newblock \bibinfo{journal}{{\em Data Mining and Knowledge Discovery\/}}
  \bibinfo{volume}{28}, \bibinfo{number}{4} (\bibinfo{year}{2014}),
  \bibinfo{pages}{851--881}.
\newblock


\bibitem[\protect\citeauthoryear{Horton, Gudmundsson, Chawla, and
  Estephan}{Horton et~al\mbox{.}}{2014}]%
        {horton2014classification}
\bibfield{author}{\bibinfo{person}{M. Horton} {}et al.}
  \bibinfo{year}{2014}\natexlab{}.
\newblock \showarticletitle{Classification of passes in football matches using
  spatiotemporal data}.
\newblock \bibinfo{journal}{{\em arXiv preprint arXiv:1407.5093\/}}
  (\bibinfo{year}{2014}).
\newblock


\bibitem[\protect\citeauthoryear{Horv{\'a}th}{Horv{\'a}th}{2001}]%
        {horvath2001change}
\bibfield{author}{\bibinfo{person}{L. Horv{\'a}th}.}
  \bibinfo{year}{2001}\natexlab{}.
\newblock \bibinfo{title}{Change-point detection in long-memory processes}.
\newblock   (\bibinfo{year}{2001}), \bibinfo{numpages}{218--234}~pages.
\newblock


\bibitem[\protect\citeauthoryear{Huang, Matijas, Suykens, et~al\mbox{.}}{Huang
  et~al\mbox{.}}{2013}]%
        {huang2013hinging}
\bibfield{author}{\bibinfo{person}{X. Huang} {}et al.}
  \bibinfo{year}{2013}\natexlab{}.
\newblock \showarticletitle{Hinging hyperplanes for time-series segmentation}.
\newblock \bibinfo{journal}{{\em TNLLS\/}} \bibinfo{volume}{24},
  \bibinfo{number}{8} (\bibinfo{year}{2013}), \bibinfo{pages}{1279--1291}.
\newblock


\bibitem[\protect\citeauthoryear{Huang, Shekhar, and Xiong}{Huang
  et~al\mbox{.}}{2004}]%
        {huang2004discovering}
\bibfield{author}{\bibinfo{person}{Y. Huang} {}et al.}
  \bibinfo{year}{2004}\natexlab{}.
\newblock \showarticletitle{Discovering colocation patterns from spatial data
  sets: a general approach}.
\newblock \bibinfo{journal}{{\em IEEE Transactions on Knowledge and Data
  Engineering\/}} \bibinfo{volume}{16}, \bibinfo{number}{12}
  (\bibinfo{year}{2004}), \bibinfo{pages}{1472--1485}.
\newblock


\bibitem[\protect\citeauthoryear{Huang, Zhang, and Zhang}{Huang
  et~al\mbox{.}}{2008}]%
        {huang2008framework}
\bibfield{author}{\bibinfo{person}{Y. Huang} {}et al.}
  \bibinfo{year}{2008}\natexlab{}.
\newblock \showarticletitle{A framework for mining sequential patterns from
  spatio-temporal event data sets}.
\newblock \bibinfo{journal}{{\em IEEE Transactions on Knowledge and data
  engineering\/}} \bibinfo{volume}{20}, \bibinfo{number}{4}
  (\bibinfo{year}{2008}), \bibinfo{pages}{433--448}.
\newblock


\bibitem[\protect\citeauthoryear{Hurlburt, Cheung, Schrijver, Chang, Freeland,
  Green, Heck, Jaffey, Kobashi, Schiff, et~al\mbox{.}}{Hurlburt
  et~al\mbox{.}}{2010}]%
        {hurlburt2010heliophysics}
\bibfield{author}{\bibinfo{person}{N. Hurlburt} {}et al.}
  \bibinfo{year}{2010}\natexlab{}.
\newblock \showarticletitle{Heliophysics event knowledgebase for the Solar
  Dynamics Observatory (SDO) and beyond}.
\newblock In \bibinfo{booktitle}{{\em The Solar Dynamics Observatory}}.
  \bibinfo{publisher}{Springer}, \bibinfo{pages}{67--78}.
\newblock


\bibitem[\protect\citeauthoryear{Ihler, Hutchins, and Smyth}{Ihler
  et~al\mbox{.}}{2006}]%
        {ihler2006adaptive}
\bibfield{author}{\bibinfo{person}{A. Ihler} {}et al.}
  \bibinfo{year}{2006}\natexlab{}.
\newblock \showarticletitle{Adaptive event detection with time-varying poisson
  processes}. In \bibinfo{booktitle}{{\em SIGKDD}}. \bibinfo{pages}{207--216}.
\newblock


\bibitem[\protect\citeauthoryear{Inclan and Tiao}{Inclan and Tiao}{1994}]%
        {inclan1994use}
\bibfield{author}{\bibinfo{person}{C. Inclan} {}et al.}
  \bibinfo{year}{1994}\natexlab{}.
\newblock \showarticletitle{Use of cumulative sums of squares for retrospective
  detection of changes of variance}.
\newblock \bibinfo{journal}{{\it J. Amer. Statist. Assoc.}}
  \bibinfo{volume}{89}, \bibinfo{number}{427} (\bibinfo{year}{1994}),
  \bibinfo{pages}{913--923}.
\newblock


\bibitem[\protect\citeauthoryear{Jain, Zamir, Savarese, and Saxena}{Jain
  et~al\mbox{.}}{2016}]%
        {jain2016structural}
\bibfield{author}{\bibinfo{person}{A. Jain} {}et al.}
  \bibinfo{year}{2016}\natexlab{}.
\newblock \showarticletitle{Structural-RNN: Deep learning on spatio-temporal
  graphs}. In \bibinfo{booktitle}{{\em Proceedings of the IEEE Conference on
  Computer Vision and Pattern Recognition}}. \bibinfo{pages}{5308--5317}.
\newblock


\bibitem[\protect\citeauthoryear{Jeung, Yiu, Zhou, Jensen, and Shen}{Jeung
  et~al\mbox{.}}{2008}]%
        {jeung2008discovery}
\bibfield{author}{\bibinfo{person}{H. Jeung} {}et al.}
  \bibinfo{year}{2008}\natexlab{}.
\newblock \showarticletitle{Discovery of convoys in trajectory databases}.
\newblock \bibinfo{journal}{{\em VLDB\/}} \bibinfo{volume}{1},
  \bibinfo{number}{1} (\bibinfo{year}{2008}), \bibinfo{pages}{1068--1080}.
\newblock


\bibitem[\protect\citeauthoryear{Jia, Khandelwal, Nayak, Gerber, Carlson, West,
  and Kumar}{Jia et~al\mbox{.}}{2017a}]%
        {jia2017incremental}
\bibfield{author}{\bibinfo{person}{X. Jia} {}et al.}
  \bibinfo{year}{2017}\natexlab{a}.
\newblock \showarticletitle{Incremental Dual-memory LSTM in Land Cover
  Prediction}. In \bibinfo{booktitle}{{\em Proceedings of the 23rd ACM SIGKDD
  International Conference on Knowledge Discovery and Data Mining}}. ACM,
  \bibinfo{pages}{867--876}.
\newblock


\bibitem[\protect\citeauthoryear{Jia, Khandelwal, Nayak, Gerber, Carlson, West,
  and Kumar}{Jia et~al\mbox{.}}{2017b}]%
        {jia2017predict}
\bibfield{author}{\bibinfo{person}{X. Jia} {}et al.}
  \bibinfo{year}{2017}\natexlab{b}.
\newblock \showarticletitle{Predict land covers with transition modeling and
  incremental learning}. In \bibinfo{booktitle}{{\em Proceedings of the 2017
  SIAM International Conference on Data Mining}}. SIAM,
  \bibinfo{pages}{171--179}.
\newblock


\bibitem[\protect\citeauthoryear{Jiang, Shekhar, Zhou, Knight, and
  Corcoran}{Jiang et~al\mbox{.}}{2015}]%
        {jiang2015focal}
\bibfield{author}{\bibinfo{person}{Z. Jiang} {}et al.}
  \bibinfo{year}{2015}\natexlab{}.
\newblock \showarticletitle{Focal-test-based spatial decision tree learning}.
\newblock \bibinfo{journal}{{\em IEEE Transactions on Knowledge and Data
  Engineering\/}} \bibinfo{volume}{27}, \bibinfo{number}{6}
  (\bibinfo{year}{2015}), \bibinfo{pages}{1547--1559}.
\newblock


\bibitem[\protect\citeauthoryear{Jun and Ghosh}{Jun and Ghosh}{2011}]%
        {jun2011spatially}
\bibfield{author}{\bibinfo{person}{G. Jun} {}et al.}
  \bibinfo{year}{2011}\natexlab{}.
\newblock \showarticletitle{Spatially adaptive classification of land cover
  with remote sensing data}.
\newblock \bibinfo{journal}{{\em IEEE Transactions on Geoscience and Remote
  Sensing\/}} \bibinfo{volume}{49}, \bibinfo{number}{7} (\bibinfo{year}{2011}),
  \bibinfo{pages}{2662--2673}.
\newblock


\bibitem[\protect\citeauthoryear{Kalnis, Mamoulis, and Bakiras}{Kalnis
  et~al\mbox{.}}{2005}]%
        {kalnis2005discovering}
\bibfield{author}{\bibinfo{person}{P. Kalnis} {}et al.}
  \bibinfo{year}{2005}\natexlab{}.
\newblock \showarticletitle{On discovering moving clusters in spatio-temporal
  data}. In \bibinfo{booktitle}{{\em ISSTD}}. Springer,
  \bibinfo{pages}{364--381}.
\newblock


\bibitem[\protect\citeauthoryear{Karpathy, Toderici, Shetty, Leung, Sukthankar,
  and Fei-Fei}{Karpathy et~al\mbox{.}}{2014}]%
        {karpathy2014large}
\bibfield{author}{\bibinfo{person}{A. Karpathy} {}et al.}
  \bibinfo{year}{2014}\natexlab{}.
\newblock \showarticletitle{Large-scale video classification with convolutional
  neural networks}. In \bibinfo{booktitle}{{\em Proceedings of the IEEE
  conference on Computer Vision and Pattern Recognition}}.
  \bibinfo{pages}{1725--1732}.
\newblock


\bibitem[\protect\citeauthoryear{Karpatne, Atluri, Faghmous, Steinbach,
  Banerjee, Ganguly, Shekhar, Samatova, and Kumar}{Karpatne
  et~al\mbox{.}}{2017}]%
        {karpatne2016theory}
\bibfield{author}{\bibinfo{person}{A. Karpatne} {}et al.}
  \bibinfo{year}{2017}\natexlab{}.
\newblock \showarticletitle{Theory-guided Data Science: A New Paradigm for
  Scientific Discovery}.
\newblock \bibinfo{journal}{{\em TKDE\/}} (\bibinfo{year}{2017}).
\newblock


\bibitem[\protect\citeauthoryear{Karpatne, Faghmous, Kawale, Styles, Blank,
  Mithal, Chen, Khandelwal, Boriah, Steinhaeuser, et~al\mbox{.}}{Karpatne
  et~al\mbox{.}}{2013}]%
        {karpatne2013earth}
\bibfield{author}{\bibinfo{person}{A. Karpatne} {}et al.}
  \bibinfo{year}{2013}\natexlab{}.
\newblock \showarticletitle{Earth science applications of sensor data}.
\newblock In \bibinfo{booktitle}{{\em Managing and Mining Sensor Data}}.
  \bibinfo{publisher}{Springer}, \bibinfo{pages}{505--530}.
\newblock


\bibitem[\protect\citeauthoryear{Karpatne, Jiang, Vatsavai, Shekhar, and
  Kumar}{Karpatne et~al\mbox{.}}{2016a}]%
        {karpatne2016monitoring}
\bibfield{author}{\bibinfo{person}{A. Karpatne} {}et al.}
  \bibinfo{year}{2016}\natexlab{a}.
\newblock \showarticletitle{Monitoring Land-Cover Changes: A Machine-Learning
  Perspective}.
\newblock \bibinfo{journal}{{\em IEEE Geoscience and Remote Sensing
  Magazine\/}} \bibinfo{volume}{4}, \bibinfo{number}{2} (\bibinfo{year}{2016}),
  \bibinfo{pages}{8--21}.
\newblock


\bibitem[\protect\citeauthoryear{Karpatne, Khandelwal, Chen, Mithal, Faghmous,
  and Kumar}{Karpatne et~al\mbox{.}}{2016b}]%
        {karpatne2016global}
\bibfield{author}{\bibinfo{person}{A. Karpatne} {}et al.}
  \bibinfo{year}{2016}\natexlab{b}.
\newblock \showarticletitle{Global monitoring of inland water dynamics:
  state-of-the-art, challenges, and opportunities}.
\newblock In \bibinfo{booktitle}{{\em Computational Sustainability}}.
  \bibinfo{publisher}{Springer}, \bibinfo{pages}{121--147}.
\newblock


\bibitem[\protect\citeauthoryear{Kasetkasem and Varshney}{Kasetkasem and
  Varshney}{2002}]%
        {kasetkasem2002image}
\bibfield{author}{\bibinfo{person}{T. Kasetkasem} {}et al.}
  \bibinfo{year}{2002}\natexlab{}.
\newblock \showarticletitle{An image change detection algorithm based on Markov
  random field models}.
\newblock \bibinfo{journal}{{\em Geoscience and Remote Sensing, IEEE
  Transactions on\/}} \bibinfo{volume}{40}, \bibinfo{number}{8}
  (\bibinfo{year}{2002}), \bibinfo{pages}{1815--1823}.
\newblock


\bibitem[\protect\citeauthoryear{Kawale, Steinbach, and Kumar}{Kawale
  et~al\mbox{.}}{2011}]%
        {kawale2011discovering}
\bibfield{author}{\bibinfo{person}{J. Kawale} {}et al.}
  \bibinfo{year}{2011}\natexlab{}.
\newblock \showarticletitle{Discovering Dynamic Dipoles in Climate Data.}. In
  \bibinfo{booktitle}{{\em SDM}}. SIAM, \bibinfo{pages}{107--118}.
\newblock


\bibitem[\protect\citeauthoryear{Kelejian and Prucha}{Kelejian and
  Prucha}{1999}]%
        {kelejian1999generalized}
\bibfield{author}{\bibinfo{person}{H.~H. Kelejian} {}et al.}
  \bibinfo{year}{1999}\natexlab{}.
\newblock \showarticletitle{A generalized moments estimator for the
  autoregressive parameter in a spatial model}.
\newblock \bibinfo{journal}{{\em International economic review\/}}
  \bibinfo{volume}{40}, \bibinfo{number}{2} (\bibinfo{year}{1999}),
  \bibinfo{pages}{509--533}.
\newblock


\bibitem[\protect\citeauthoryear{Keogh, Chu, Hart, and Pazzani}{Keogh
  et~al\mbox{.}}{2004}]%
        {keogh2004segmenting}
\bibfield{author}{\bibinfo{person}{E. Keogh} {}et al.}
  \bibinfo{year}{2004}\natexlab{}.
\newblock \showarticletitle{Segmenting time series: A survey and novel
  approach}.
\newblock \bibinfo{journal}{{\em DMTSD\/}}  \bibinfo{volume}{57}
  (\bibinfo{year}{2004}), \bibinfo{pages}{1--22}.
\newblock


\bibitem[\protect\citeauthoryear{Keogh and Kasetty}{Keogh and Kasetty}{2003}]%
        {keogh2003need}
\bibfield{author}{\bibinfo{person}{E. Keogh} {}et al.}
  \bibinfo{year}{2003}\natexlab{}.
\newblock \showarticletitle{On the need for time series data mining benchmarks:
  a survey and empirical demonstration}.
\newblock \bibinfo{journal}{{\em Data Mining and knowledge discovery\/}}
  \bibinfo{volume}{7}, \bibinfo{number}{4} (\bibinfo{year}{2003}),
  \bibinfo{pages}{349--371}.
\newblock


\bibitem[\protect\citeauthoryear{Keogh, Lin, and Fu}{Keogh
  et~al\mbox{.}}{2005}]%
        {keogh2005hot}
\bibfield{author}{\bibinfo{person}{E. Keogh} {}et al.}
  \bibinfo{year}{2005}\natexlab{}.
\newblock \showarticletitle{Hot sax: Efficiently finding the most unusual time
  series subsequence}. In \bibinfo{booktitle}{{\em ICDM}}.
  \bibinfo{pages}{8--pp}.
\newblock


\bibitem[\protect\citeauthoryear{Keogh, Lonardi, and Chiu}{Keogh
  et~al\mbox{.}}{2002}]%
        {keogh2002finding}
\bibfield{author}{\bibinfo{person}{E. Keogh} {}et al.}
  \bibinfo{year}{2002}\natexlab{}.
\newblock \showarticletitle{Finding surprising patterns in a time series
  database in linear time and space}. In \bibinfo{booktitle}{{\em SIGKDD}}.
  ACM, \bibinfo{pages}{550--556}.
\newblock


\bibitem[\protect\citeauthoryear{Keogh and Ratanamahatana}{Keogh and
  Ratanamahatana}{2005}]%
        {keogh2005exact}
\bibfield{author}{\bibinfo{person}{E. Keogh} {}et al.}
  \bibinfo{year}{2005}\natexlab{}.
\newblock \showarticletitle{Exact indexing of dynamic time warping}.
\newblock \bibinfo{journal}{{\em Knowledge and information systems\/}}
  \bibinfo{volume}{7}, \bibinfo{number}{3} (\bibinfo{year}{2005}),
  \bibinfo{pages}{358--386}.
\newblock


\bibitem[\protect\citeauthoryear{Khandelwal, Karpatne, Marlier, Kim,
  Lettenmaier, and Kumar}{Khandelwal et~al\mbox{.}}{2017}]%
        {rse}
\bibfield{author}{\bibinfo{person}{A. Khandelwal} {}et al.}
  \bibinfo{year}{2017}\natexlab{}.
\newblock \showarticletitle{An Approach for Global Monitoring of Surface Water
  Extent Variations Using MODIS Data}. In \bibinfo{booktitle}{{\em Remote
  Sensing of Environment}}.
\newblock


\bibitem[\protect\citeauthoryear{Kisilevich, Mansmann, Nanni, and
  Rinzivillo}{Kisilevich et~al\mbox{.}}{2010}]%
        {kisilevich2010spatio}
\bibfield{author}{\bibinfo{person}{S. Kisilevich} {}et al.}
  \bibinfo{year}{2010}\natexlab{}.
\newblock \showarticletitle{Spatio-temporal clustering}.
\newblock \bibinfo{journal}{{\em DM \& KD Handbook\/}}  \bibinfo{volume}{1},
  \bibinfo{pages}{855}.
\newblock


\bibitem[\protect\citeauthoryear{Kistler, Collins, Saha, White, Woollen,
  Kalnay, Chelliah, Ebisuzaki, Kanamitsu, Kousky, et~al\mbox{.}}{Kistler
  et~al\mbox{.}}{2001}]%
        {kistler2001ncep}
\bibfield{author}{\bibinfo{person}{R. Kistler} {}et al.}
  \bibinfo{year}{2001}\natexlab{}.
\newblock \showarticletitle{The NCEP--NCAR 50--year reanalysis: Monthly means
  CD--ROM and documentation}.
\newblock \bibinfo{journal}{{\em Bulletin of the American Meteorological
  society\/}} \bibinfo{volume}{82}, \bibinfo{number}{2} (\bibinfo{year}{2001}),
  \bibinfo{pages}{247--267}.
\newblock


\bibitem[\protect\citeauthoryear{Knorr and Ng}{Knorr and Ng}{1997}]%
        {knorr1997unified}
\bibfield{author}{\bibinfo{person}{E.~M. Knorr} {}et al.}
  \bibinfo{year}{1997}\natexlab{}.
\newblock \showarticletitle{A Unified Notion of Outliers: Properties and
  Computation}. In \bibinfo{booktitle}{{\em KDD}}. \bibinfo{pages}{219--222}.
\newblock


\bibitem[\protect\citeauthoryear{Knorr, Ng, and Tucakov}{Knorr
  et~al\mbox{.}}{2000}]%
        {knorr2000distance}
\bibfield{author}{\bibinfo{person}{E.~M. Knorr} {}et al.}
  \bibinfo{year}{2000}\natexlab{}.
\newblock \showarticletitle{Distance-based outliers: algorithms and
  applications}.
\newblock \bibinfo{journal}{{\em VLDB\/}} \bibinfo{volume}{8},
  \bibinfo{number}{3-4} (\bibinfo{year}{2000}), \bibinfo{pages}{237--253}.
\newblock


\bibitem[\protect\citeauthoryear{Kou, Lu, and Dos~Santos}{Kou
  et~al\mbox{.}}{2007}]%
        {kou2007spatial}
\bibfield{author}{\bibinfo{person}{Y. Kou} {}et al.}
  \bibinfo{year}{2007}\natexlab{}.
\newblock \showarticletitle{Spatial outlier detection: a graph-based approach}.
  In \bibinfo{booktitle}{{\em ICTAI}}, Vol.~\bibinfo{volume}{1}. IEEE,
  \bibinfo{pages}{281--288}.
\newblock


\bibitem[\protect\citeauthoryear{Kratz and Nishino}{Kratz and Nishino}{2009}]%
        {kratz2009anomaly}
\bibfield{author}{\bibinfo{person}{L. Kratz} {}et al.}
  \bibinfo{year}{2009}\natexlab{}.
\newblock \showarticletitle{Anomaly detection in extremely crowded scenes using
  spatio-temporal motion pattern models}. In \bibinfo{booktitle}{{\em CVPR}}.
  IEEE, \bibinfo{pages}{1446--1453}.
\newblock


\bibitem[\protect\citeauthoryear{Krizhevsky, Sutskever, and Hinton}{Krizhevsky
  et~al\mbox{.}}{2012}]%
        {krizhevsky2012imagenet}
\bibfield{author}{\bibinfo{person}{A. Krizhevsky} {}et al.}
  \bibinfo{year}{2012}\natexlab{}.
\newblock \showarticletitle{Imagenet classification with deep convolutional
  neural networks}. In \bibinfo{booktitle}{{\em NIPS}}.
  \bibinfo{pages}{1097--1105}.
\newblock


\bibitem[\protect\citeauthoryear{Ku, Gretton, Macke, and Logothetis}{Ku
  et~al\mbox{.}}{2008}]%
        {ku2008comparison}
\bibfield{author}{\bibinfo{person}{S.-p. Ku} {}et al.}
  \bibinfo{year}{2008}\natexlab{}.
\newblock \showarticletitle{Comparison of pattern recognition methods in
  classifying high-resolution BOLD signals obtained at high magnetic field in
  monkeys}.
\newblock \bibinfo{journal}{{\em MRI\/}} \bibinfo{volume}{26},
  \bibinfo{number}{7} (\bibinfo{year}{2008}), \bibinfo{pages}{1007--1014}.
\newblock


\bibitem[\protect\citeauthoryear{Kulldorff}{Kulldorff}{1997}]%
        {kulldorff1997spatial}
\bibfield{author}{\bibinfo{person}{M. Kulldorff}.}
  \bibinfo{year}{1997}\natexlab{}.
\newblock \showarticletitle{A spatial scan statistic}.
\newblock \bibinfo{journal}{{\em Comm. in Stat.-Theory and methods\/}}
  \bibinfo{volume}{26}, \bibinfo{number}{6} (\bibinfo{year}{1997}),
  \bibinfo{pages}{1481--1496}.
\newblock


\bibitem[\protect\citeauthoryear{Kulldorff}{Kulldorff}{2001}]%
        {kulldorff2001prospective}
\bibfield{author}{\bibinfo{person}{M. Kulldorff}.}
  \bibinfo{year}{2001}\natexlab{}.
\newblock \showarticletitle{Prospective time periodic geographical disease
  surveillance using a scan statistic}.
\newblock \bibinfo{journal}{{\em Journal of the Royal Statistical Society:
  Series A (Statistics in Society)\/}} \bibinfo{volume}{164},
  \bibinfo{number}{1} (\bibinfo{year}{2001}), \bibinfo{pages}{61--72}.
\newblock


\bibitem[\protect\citeauthoryear{Kulldorff, Heffernan, Hartman, Assun{\c{c}}ao,
  and Mostashari}{Kulldorff et~al\mbox{.}}{2005}]%
        {kulldorff2005space}
\bibfield{author}{\bibinfo{person}{M. Kulldorff} {}et al.}
  \bibinfo{year}{2005}\natexlab{}.
\newblock \showarticletitle{A space--time permutation scan statistic for
  disease outbreak detection}.
\newblock \bibinfo{journal}{{\em Plos med\/}} \bibinfo{volume}{2},
  \bibinfo{number}{3} (\bibinfo{year}{2005}), \bibinfo{pages}{e59}.
\newblock


\bibitem[\protect\citeauthoryear{Kut and Birant}{Kut and Birant}{2006}]%
        {kut2006spatio}
\bibfield{author}{\bibinfo{person}{A. Kut} {}et al.}
  \bibinfo{year}{2006}\natexlab{}.
\newblock \showarticletitle{Spatio-temporal outlier detection in large
  databases}.
\newblock \bibinfo{journal}{{\em CIT\/}} \bibinfo{volume}{14},
  \bibinfo{number}{4} (\bibinfo{year}{2006}), \bibinfo{pages}{291--297}.
\newblock


\bibitem[\protect\citeauthoryear{Lappas, Vieira, Gunopulos, and Tsotras}{Lappas
  et~al\mbox{.}}{2012}]%
        {lappas2012spatiotemporal}
\bibfield{author}{\bibinfo{person}{T. Lappas} {}et al.}
  \bibinfo{year}{2012}\natexlab{}.
\newblock \showarticletitle{On the spatiotemporal burstiness of terms}.
\newblock \bibinfo{journal}{{\em Proceedings of the VLDB Endowment\/}}
  \bibinfo{volume}{5}, \bibinfo{number}{9} (\bibinfo{year}{2012}),
  \bibinfo{pages}{836--847}.
\newblock


\bibitem[\protect\citeauthoryear{Lappas, Vieira, Gunopulos, and Tsotras}{Lappas
  et~al\mbox{.}}{2013}]%
        {lappas2013stem}
\bibfield{author}{\bibinfo{person}{T. Lappas} {}et al.}
  \bibinfo{year}{2013}\natexlab{}.
\newblock \showarticletitle{STEM: A spatio-temporal miner for bursty activity}.
  In \bibinfo{booktitle}{{\em Proceedings of the 2013 ACM SIGMOD International
  Conference on Management of Data}}. ACM, \bibinfo{pages}{1021--1024}.
\newblock


\bibitem[\protect\citeauthoryear{LeCun and Bengio}{LeCun and Bengio}{1995}]%
        {lecun1995convolutional}
\bibfield{author}{\bibinfo{person}{Y. LeCun} {}et al.}
  \bibinfo{year}{1995}\natexlab{}.
\newblock \showarticletitle{Convolutional networks for images, speech, and time
  series}.
\newblock \bibinfo{journal}{{\em The handbook of brain theory and neural
  networks\/}} \bibinfo{volume}{3361}, \bibinfo{number}{10}
  (\bibinfo{year}{1995}), \bibinfo{pages}{1995}.
\newblock


\bibitem[\protect\citeauthoryear{Lee, Han, and Li}{Lee et~al\mbox{.}}{2008}]%
        {lee2008trajectory}
\bibfield{author}{\bibinfo{person}{J.-G. Lee} {}et al.}
  \bibinfo{year}{2008}\natexlab{}.
\newblock \showarticletitle{Trajectory outlier detection: A
  partition-and-detect framework}. In \bibinfo{booktitle}{{\em ICDE}}.
  \bibinfo{pages}{140--149}.
\newblock


\bibitem[\protect\citeauthoryear{Lee, Han, and Whang}{Lee
  et~al\mbox{.}}{2007}]%
        {lee2007trajectory}
\bibfield{author}{\bibinfo{person}{J.-G. Lee} {}et al.}
  \bibinfo{year}{2007}\natexlab{}.
\newblock \showarticletitle{Trajectory clustering: a partition-and-group
  framework}. In \bibinfo{booktitle}{{\em SIGMOD}}. ACM,
  \bibinfo{pages}{593--604}.
\newblock


\bibitem[\protect\citeauthoryear{Leskovec, Lang, and Mahoney}{Leskovec
  et~al\mbox{.}}{2010}]%
        {leskovec2010empirical}
\bibfield{author}{\bibinfo{person}{J. Leskovec} {}et al.}
  \bibinfo{year}{2010}\natexlab{}.
\newblock \showarticletitle{Empirical comparison of algorithms for network
  community detection}. In \bibinfo{booktitle}{{\em Proceedings of the 19th
  international conference on World wide web}}. ACM, \bibinfo{pages}{631--640}.
\newblock


\bibitem[\protect\citeauthoryear{Lhermitte, Verbesselt, Jonckheere, Nackaerts,
  van Aardt, Verstraeten, and Coppin}{Lhermitte et~al\mbox{.}}{2008}]%
        {lhermitte2008hierarchical}
\bibfield{author}{\bibinfo{person}{S. Lhermitte} {}et al.}
  \bibinfo{year}{2008}\natexlab{}.
\newblock \showarticletitle{Hierarchical image segmentation based on similarity
  of NDVI time series}.
\newblock \bibinfo{journal}{{\em Remote Sensing of Environment\/}}
  \bibinfo{volume}{112}, \bibinfo{number}{2} (\bibinfo{year}{2008}),
  \bibinfo{pages}{506--521}.
\newblock


\bibitem[\protect\citeauthoryear{Li, Calder, and Cressie}{Li
  et~al\mbox{.}}{2007a}]%
        {li2007beyond}
\bibfield{author}{\bibinfo{person}{H. Li} {}et al.}
  \bibinfo{year}{2007}\natexlab{a}.
\newblock \showarticletitle{Beyond Moran's I: testing for spatial dependence
  based on the spatial autoregressive model}.
\newblock \bibinfo{journal}{{\em Geographical Analysis\/}}
  \bibinfo{volume}{39}, \bibinfo{number}{4} (\bibinfo{year}{2007}),
  \bibinfo{pages}{357--375}.
\newblock


\bibitem[\protect\citeauthoryear{Li, Zang, Zhang, Li, Wu, et~al\mbox{.}}{Li
  et~al\mbox{.}}{2014b}]%
        {li2014review}
\bibfield{author}{\bibinfo{person}{M. Li} {}et al.}
  \bibinfo{year}{2014}\natexlab{b}.
\newblock \showarticletitle{A review of remote sensing image classification
  techniques: The role of spatio-contextual information}.
\newblock \bibinfo{journal}{{\em European Journal of Remote Sensing\/}}
  \bibinfo{volume}{47}, \bibinfo{number}{2014} (\bibinfo{year}{2014}),
  \bibinfo{pages}{389--411}.
\newblock


\bibitem[\protect\citeauthoryear{Li, Mahadevan, and Vasconcelos}{Li
  et~al\mbox{.}}{2014a}]%
        {li2014anomaly}
\bibfield{author}{\bibinfo{person}{W. Li} {}et al.}
  \bibinfo{year}{2014}\natexlab{a}.
\newblock \showarticletitle{Anomaly detection and localization in crowded
  scenes}.
\newblock \bibinfo{journal}{{\em TPAMI\/}} \bibinfo{volume}{36},
  \bibinfo{number}{1} (\bibinfo{year}{2014}), \bibinfo{pages}{18--32}.
\newblock


\bibitem[\protect\citeauthoryear{Li and Yang}{Li and Yang}{2009}]%
        {li2009comparing}
\bibfield{author}{\bibinfo{person}{W. Li} {}et al.}
  \bibinfo{year}{2009}\natexlab{}.
\newblock \showarticletitle{Comparing networks from a data analysis
  perspective}. In \bibinfo{booktitle}{{\em CCS}}. Springer,
  \bibinfo{pages}{1907--1916}.
\newblock


\bibitem[\protect\citeauthoryear{Li, Han, and Kim}{Li et~al\mbox{.}}{2006}]%
        {li2006motion}
\bibfield{author}{\bibinfo{person}{X. Li} {}et al.}
  \bibinfo{year}{2006}\natexlab{}.
\newblock \showarticletitle{Motion-alert: automatic anomaly detection in
  massive moving objects}. In \bibinfo{booktitle}{{\em International Conference
  on Intelligence and Security Informatics}}. Springer,
  \bibinfo{pages}{166--177}.
\newblock


\bibitem[\protect\citeauthoryear{Li, Han, Kim, and Gonzalez}{Li
  et~al\mbox{.}}{2007b}]%
        {li2007roam}
\bibfield{author}{\bibinfo{person}{X. Li} {}et al.}
  \bibinfo{year}{2007}\natexlab{b}.
\newblock \showarticletitle{Roam: Rule-and motif-based anomaly detection in
  massive moving object data sets}. In \bibinfo{booktitle}{{\em Proceedings of
  the 2007 SIAM International Conference on Data Mining}}. SIAM,
  \bibinfo{pages}{273--284}.
\newblock


\bibitem[\protect\citeauthoryear{Li, Li, Han, and Lee}{Li
  et~al\mbox{.}}{2009}]%
        {li2009temporal}
\bibfield{author}{\bibinfo{person}{X. Li} {}et al.}
  \bibinfo{year}{2009}\natexlab{}.
\newblock \showarticletitle{Temporal outlier detection in vehicle traffic
  data}. In \bibinfo{booktitle}{{\em ICDE}}. IEEE, \bibinfo{pages}{1319--1322}.
\newblock


\bibitem[\protect\citeauthoryear{Li, Bailey, Kulik, and Pei}{Li
  et~al\mbox{.}}{2013}]%
        {li2013mining}
\bibfield{author}{\bibinfo{person}{Y. Li} {}et al.}
  \bibinfo{year}{2013}\natexlab{}.
\newblock \showarticletitle{Mining probabilistic frequent spatio-temporal
  sequential patterns with gap constraints from uncertain databases}. In
  \bibinfo{booktitle}{{\em ICDM}}. IEEE, \bibinfo{pages}{448--457}.
\newblock


\bibitem[\protect\citeauthoryear{Li, Li, Gunopulos, and Guibas}{Li
  et~al\mbox{.}}{2016}]%
        {li2016knowledge}
\bibfield{author}{\bibinfo{person}{Y. Li} {}et al.}
  \bibinfo{year}{2016}\natexlab{}.
\newblock \showarticletitle{Knowledge-based trajectory completion from sparse
  GPS samples}. In \bibinfo{booktitle}{{\em Proceedings of the 24th ACM
  SIGSPATIAL International Conference on Advances in Geographic Information
  Systems}}. ACM, \bibinfo{pages}{33}.
\newblock


\bibitem[\protect\citeauthoryear{Li}{Li}{2014}]%
        {li2014spatiotemporal}
\bibfield{author}{\bibinfo{person}{Z. Li}.} \bibinfo{year}{2014}\natexlab{}.
\newblock \showarticletitle{Spatiotemporal pattern mining: algorithms and
  applications}.
\newblock In \bibinfo{booktitle}{{\em Frequent Pattern Mining}}.
  \bibinfo{publisher}{Springer}, \bibinfo{pages}{283--306}.
\newblock


\bibitem[\protect\citeauthoryear{Li}{Li}{2017}]%
        {li2017semantic}
\bibfield{author}{\bibinfo{person}{Z. Li}.} \bibinfo{year}{2017}\natexlab{}.
\newblock \showarticletitle{Semantic Understanding of Spatial Trajectories}. In
  \bibinfo{booktitle}{{\em International Symposium on Spatial and Temporal
  Databases}}. Springer, Cham, \bibinfo{pages}{398--401}.
\newblock


\bibitem[\protect\citeauthoryear{Li, Ding, Han, and Kays}{Li
  et~al\mbox{.}}{2010}]%
        {li2010swarm}
\bibfield{author}{\bibinfo{person}{Z. Li} {}et al.}
  \bibinfo{year}{2010}\natexlab{}.
\newblock \showarticletitle{Swarm: Mining relaxed temporal moving object
  clusters}.
\newblock \bibinfo{journal}{{\em VLDB\/}} \bibinfo{volume}{3},
  \bibinfo{number}{1-2} (\bibinfo{year}{2010}), \bibinfo{pages}{723--734}.
\newblock


\bibitem[\protect\citeauthoryear{Li and Han}{Li and Han}{2014}]%
        {li2014mining}
\bibfield{author}{\bibinfo{person}{Z. Li} {}et al.}
  \bibinfo{year}{2014}\natexlab{}.
\newblock \showarticletitle{Mining periodicity from dynamic and incomplete
  spatiotemporal data}.
\newblock In \bibinfo{booktitle}{{\em Data Mining and Knowledge Discovery for
  Big Data}}. \bibinfo{publisher}{Springer}, \bibinfo{pages}{41--81}.
\newblock


\bibitem[\protect\citeauthoryear{Liang, Chen, Hawbaker, Zhu, and Gong}{Liang
  et~al\mbox{.}}{2014}]%
        {Liang2014}
\bibfield{author}{\bibinfo{person}{L. Liang} {}et al.}
  \bibinfo{year}{2014}\natexlab{}.
\newblock \showarticletitle{{Mapping mountain pine beetle mortality through
  growth trend analysis of time-series Landsat data}}.
\newblock \bibinfo{journal}{{\em Remote Sensing\/}} (\bibinfo{year}{2014}).
\newblock
\showURL{%
\url{http://www.mdpi.com/2072-4292/6/6/5696/htm}}


\bibitem[\protect\citeauthoryear{Liao}{Liao}{2005}]%
        {liao2005clustering}
\bibfield{author}{\bibinfo{person}{T.~W. Liao}.}
  \bibinfo{year}{2005}\natexlab{}.
\newblock \showarticletitle{Clustering of time series data—a survey}.
\newblock \bibinfo{journal}{{\em Pattern recognition\/}} \bibinfo{volume}{38},
  \bibinfo{number}{11} (\bibinfo{year}{2005}), \bibinfo{pages}{1857--1874}.
\newblock


\bibitem[\protect\citeauthoryear{Liu, Liu, Ni, Fan, and Li}{Liu
  et~al\mbox{.}}{2010}]%
        {liu2010towards}
\bibfield{author}{\bibinfo{person}{S. Liu} {}et al.}
  \bibinfo{year}{2010}\natexlab{}.
\newblock \showarticletitle{Towards mobility-based clustering}. In
  \bibinfo{booktitle}{{\em SIGKDD}}. ACM, \bibinfo{pages}{919--928}.
\newblock


\bibitem[\protect\citeauthoryear{Liu, Zheng, Chawla, Yuan, and Xing}{Liu
  et~al\mbox{.}}{2011}]%
        {liu2011discovering}
\bibfield{author}{\bibinfo{person}{W. Liu} {}et al.}
  \bibinfo{year}{2011}\natexlab{}.
\newblock \showarticletitle{Discovering spatio-temporal causal interactions in
  traffic data streams}. In \bibinfo{booktitle}{{\em Proceedings of the 17th
  ACM SIGKDD international conference on Knowledge discovery and data mining}}.
  ACM, \bibinfo{pages}{1010--1018}.
\newblock


\bibitem[\protect\citeauthoryear{Liu, Chang, and Duyn}{Liu
  et~al\mbox{.}}{2013}]%
        {liu2013decomposition}
\bibfield{author}{\bibinfo{person}{X. Liu} {}et al.}
  \bibinfo{year}{2013}\natexlab{}.
\newblock \showarticletitle{Decomposition of spontaneous brain activity into
  distinct fMRI co-activation patterns}.
\newblock \bibinfo{journal}{{\em Frontiers in systems neuroscience\/}}
  \bibinfo{volume}{7} (\bibinfo{year}{2013}).
\newblock


\bibitem[\protect\citeauthoryear{Liu and Duyn}{Liu and Duyn}{2013}]%
        {liu2013time}
\bibfield{author}{\bibinfo{person}{X. Liu} {}et al.}
  \bibinfo{year}{2013}\natexlab{}.
\newblock \showarticletitle{Time-varying functional network information
  extracted from brief instances of spontaneous brain activity}.
\newblock \bibinfo{journal}{{\em Proceedings of the National Academy of
  Sciences\/}} \bibinfo{volume}{110}, \bibinfo{number}{11}
  (\bibinfo{year}{2013}), \bibinfo{pages}{4392--4397}.
\newblock


\bibitem[\protect\citeauthoryear{Liu, Lin, and Wang}{Liu et~al\mbox{.}}{2008}]%
        {liu2008novel}
\bibfield{author}{\bibinfo{person}{X. Liu} {}et al.}
  \bibinfo{year}{2008}\natexlab{}.
\newblock \showarticletitle{Novel online methods for time series segmentation}.
\newblock \bibinfo{journal}{{\em TKDE\/}} \bibinfo{volume}{20},
  \bibinfo{number}{12} (\bibinfo{year}{2008}), \bibinfo{pages}{1616--1626}.
\newblock


\bibitem[\protect\citeauthoryear{Lozano, Abe, Liu, and Rosset}{Lozano
  et~al\mbox{.}}{2009a}]%
        {lozano2009grouped}
\bibfield{author}{\bibinfo{person}{A.~C. Lozano} {}et al.}
  \bibinfo{year}{2009}\natexlab{a}.
\newblock \showarticletitle{Grouped graphical Granger modeling for gene
  expression regulatory networks discovery}.
\newblock \bibinfo{journal}{{\em Bioinformatics\/}} \bibinfo{volume}{25},
  \bibinfo{number}{12} (\bibinfo{year}{2009}), \bibinfo{pages}{i110--i118}.
\newblock


\bibitem[\protect\citeauthoryear{Lozano, Li, Niculescu-Mizil, Liu, Perlich,
  Hosking, and Abe}{Lozano et~al\mbox{.}}{2009b}]%
        {lozano2009spatial}
\bibfield{author}{\bibinfo{person}{A.~C. Lozano} {}et al.}
  \bibinfo{year}{2009}\natexlab{b}.
\newblock \showarticletitle{Spatial-temporal causal modeling for climate change
  attribution}. In \bibinfo{booktitle}{{\em KDD}}. \bibinfo{pages}{587--596}.
\newblock


\bibitem[\protect\citeauthoryear{Lu, Chen, and Kou}{Lu et~al\mbox{.}}{2003a}]%
        {lu2003algorithms}
\bibfield{author}{\bibinfo{person}{C.-T. Lu} {}et al.}
  \bibinfo{year}{2003}\natexlab{a}.
\newblock \showarticletitle{Algorithms for spatial outlier detection}. In
  \bibinfo{booktitle}{{\em ICDM}}. IEEE, \bibinfo{pages}{597--600}.
\newblock


\bibitem[\protect\citeauthoryear{Lu, Kou, Zhao, and Chen}{Lu
  et~al\mbox{.}}{2007}]%
        {lu2007detecting}
\bibfield{author}{\bibinfo{person}{C.-T. Lu} {}et al.}
  \bibinfo{year}{2007}\natexlab{}.
\newblock \showarticletitle{Detecting and tracking regional outliers in
  meteorological data}.
\newblock \bibinfo{journal}{{\em Information Sciences\/}}
  \bibinfo{volume}{177}, \bibinfo{number}{7} (\bibinfo{year}{2007}),
  \bibinfo{pages}{1609--1632}.
\newblock


\bibitem[\protect\citeauthoryear{Lu, Lall, Kawale, Liess, and Kumar}{Lu
  et~al\mbox{.}}{2016}]%
        {lu2016exploring}
\bibfield{author}{\bibinfo{person}{M. Lu} {}et al.}
  \bibinfo{year}{2016}\natexlab{}.
\newblock \showarticletitle{Exploring the Predictability of 30-Day Extreme
  Precipitation Occurrence Using a Global SST--SLP Correlation Network}.
\newblock \bibinfo{journal}{{\em Journal of Climate\/}} \bibinfo{volume}{29},
  \bibinfo{number}{3} (\bibinfo{year}{2016}), \bibinfo{pages}{1013--1029}.
\newblock


\bibitem[\protect\citeauthoryear{Lu, Jiang, and Zang}{Lu
  et~al\mbox{.}}{2003b}]%
        {lu2003region}
\bibfield{author}{\bibinfo{person}{Y. Lu} {}et al.}
  \bibinfo{year}{2003}\natexlab{b}.
\newblock \showarticletitle{Region growing method for the analysis of
  functional MRI data}.
\newblock \bibinfo{journal}{{\em NeuroImage\/}} \bibinfo{volume}{20},
  \bibinfo{number}{1} (\bibinfo{year}{2003}), \bibinfo{pages}{455--465}.
\newblock


\bibitem[\protect\citeauthoryear{Lunetta, Knight, Ediriwickrema, Lyon, and
  Worthy}{Lunetta et~al\mbox{.}}{2006}]%
        {lunetta2006land}
\bibfield{author}{\bibinfo{person}{R.~S. Lunetta} {}et al.}
  \bibinfo{year}{2006}\natexlab{}.
\newblock \showarticletitle{{Land-cover change detection using multi-temporal
  MODIS NDVI data}}.
\newblock \bibinfo{journal}{{\em Remote sensing of environment\/}}
  \bibinfo{volume}{105}, \bibinfo{number}{2} (\bibinfo{year}{2006}),
  \bibinfo{pages}{142--154}.
\newblock


\bibitem[\protect\citeauthoryear{Luo, Lu, Cheng, Valdes-Sosa, Wen, Ding, and
  Feng}{Luo et~al\mbox{.}}{2013}]%
        {luo2013spatio}
\bibfield{author}{\bibinfo{person}{Q. Luo} {}et al.}
  \bibinfo{year}{2013}\natexlab{}.
\newblock \showarticletitle{Spatio-temporal Granger causality: A new
  framework}.
\newblock \bibinfo{journal}{{\em NeuroImage\/}}  \bibinfo{volume}{79}
  (\bibinfo{year}{2013}), \bibinfo{pages}{241--263}.
\newblock


\bibitem[\protect\citeauthoryear{Lynall, Bassett, Kerwin, McKenna, Kitzbichler,
  Muller, and Bullmore}{Lynall et~al\mbox{.}}{2010}]%
        {lynall2010functional}
\bibfield{author}{\bibinfo{person}{M.-E. Lynall} {}et al.}
  \bibinfo{year}{2010}\natexlab{}.
\newblock \showarticletitle{Functional connectivity and brain networks in
  schizophrenia}.
\newblock \bibinfo{journal}{{\em The Journal of Neuroscience\/}}
  \bibinfo{volume}{30}, \bibinfo{number}{28} (\bibinfo{year}{2010}),
  \bibinfo{pages}{9477--9487}.
\newblock


\bibitem[\protect\citeauthoryear{Lynch and Moorcroft}{Lynch and
  Moorcroft}{2008}]%
        {lynch2008spatiotemporal}
\bibfield{author}{\bibinfo{person}{H.~J. Lynch} {}et al.}
  \bibinfo{year}{2008}\natexlab{}.
\newblock \showarticletitle{A spatiotemporal Ripley’s K-function to analyze
  interactions between spruce budworm and fire in British Columbia, Canada}.
\newblock \bibinfo{journal}{{\em CJFR\/}} \bibinfo{volume}{38},
  \bibinfo{number}{12} (\bibinfo{year}{2008}), \bibinfo{pages}{3112--3119}.
\newblock


\bibitem[\protect\citeauthoryear{Mahlein}{Mahlein}{2016}]%
        {mahlein2016plant}
\bibfield{author}{\bibinfo{person}{A.-K. Mahlein}.}
  \bibinfo{year}{2016}\natexlab{}.
\newblock \showarticletitle{Plant disease detection by imaging
  sensors--parallels and specific demands for precision agriculture and plant
  phenotyping}.
\newblock \bibinfo{journal}{{\em Plant Disease\/}} \bibinfo{volume}{100},
  \bibinfo{number}{2} (\bibinfo{year}{2016}), \bibinfo{pages}{241--251}.
\newblock


\bibitem[\protect\citeauthoryear{Mamoulis}{Mamoulis}{2009}]%
        {mamoulis2009spatio}
\bibfield{author}{\bibinfo{person}{N. Mamoulis}.}
  \bibinfo{year}{2009}\natexlab{}.
\newblock \showarticletitle{Spatio-temporal data mining}.
\newblock In \bibinfo{booktitle}{{\em Encyclopedia of DB Systems}}.
  \bibinfo{publisher}{Springer}, \bibinfo{pages}{2725--2730}.
\newblock


\bibitem[\protect\citeauthoryear{Mamoulis, Cao, Kollios, Hadjieleftheriou, Tao,
  and Cheung}{Mamoulis et~al\mbox{.}}{2004}]%
        {mamoulis2004mining}
\bibfield{author}{\bibinfo{person}{N. Mamoulis} {}et al.}
  \bibinfo{year}{2004}\natexlab{}.
\newblock \showarticletitle{Mining, indexing, and querying historical
  spatiotemporal data}. In \bibinfo{booktitle}{{\em Proceedings of the tenth
  ACM SIGKDD international conference on Knowledge discovery and data mining}}.
  ACM, \bibinfo{pages}{236--245}.
\newblock


\bibitem[\protect\citeauthoryear{Matsubara, Sakurai, Van~Panhuis, and
  Faloutsos}{Matsubara et~al\mbox{.}}{2014}]%
        {matsubara2014funnel}
\bibfield{author}{\bibinfo{person}{Y. Matsubara} {}et al.}
  \bibinfo{year}{2014}\natexlab{}.
\newblock \showarticletitle{FUNNEL: automatic mining of spatially coevolving
  epidemics}. In \bibinfo{booktitle}{{\em SIGKDD}}. ACM,
  \bibinfo{pages}{105--114}.
\newblock


\bibitem[\protect\citeauthoryear{Meng, Yuan, Hans, and Wu}{Meng
  et~al\mbox{.}}{2008}]%
        {meng2008mining}
\bibfield{author}{\bibinfo{person}{J. Meng} {}et al.}
  \bibinfo{year}{2008}\natexlab{}.
\newblock \showarticletitle{Mining motifs from human motion}. In
  \bibinfo{booktitle}{{\em Proc. of EUROGRAPHICS}}, Vol.~\bibinfo{volume}{8}.
\newblock


\bibitem[\protect\citeauthoryear{Mezer, Yovel, Pasternak, Gorfine, and
  Assaf}{Mezer et~al\mbox{.}}{2009}]%
        {mezer2009cluster}
\bibfield{author}{\bibinfo{person}{A. Mezer} {}et al.}
  \bibinfo{year}{2009}\natexlab{}.
\newblock \showarticletitle{Cluster analysis of resting-state fMRI time
  series}.
\newblock \bibinfo{journal}{{\em Neuroimage\/}} \bibinfo{volume}{45},
  \bibinfo{number}{4} (\bibinfo{year}{2009}), \bibinfo{pages}{1117--1125}.
\newblock


\bibitem[\protect\citeauthoryear{Mikolov, Karafi{\'a}t, Burget, Cernock{\`y},
  and Khudanpur}{Mikolov et~al\mbox{.}}{2010}]%
        {mikolov2010recurrent}
\bibfield{author}{\bibinfo{person}{T. Mikolov} {}et al.}
  \bibinfo{year}{2010}\natexlab{}.
\newblock \showarticletitle{Recurrent neural network based language model.}. In
  \bibinfo{booktitle}{{\em Interspeech}}, Vol.~\bibinfo{volume}{2}.
  \bibinfo{pages}{3}.
\newblock


\bibitem[\protect\citeauthoryear{Milo, Shen-Orr, Itzkovitz, Kashtan,
  Chklovskii, and Alon}{Milo et~al\mbox{.}}{2002}]%
        {milo2002network}
\bibfield{author}{\bibinfo{person}{R. Milo} {}et al.}
  \bibinfo{year}{2002}\natexlab{}.
\newblock \showarticletitle{Network motifs: simple building blocks of complex
  networks}.
\newblock \bibinfo{journal}{{\em Science\/}} \bibinfo{volume}{298},
  \bibinfo{number}{5594}, \bibinfo{pages}{824--827}.
\newblock


\bibitem[\protect\citeauthoryear{Minnen, Isbell, Essa, and Starner}{Minnen
  et~al\mbox{.}}{2007}]%
        {minnen2007discovering}
\bibfield{author}{\bibinfo{person}{D. Minnen} {}et al.}
  \bibinfo{year}{2007}\natexlab{}.
\newblock \showarticletitle{Discovering multivariate motifs using subsequence
  density estimation and greedy mixture learning}. In \bibinfo{booktitle}{{\em
  Natnl. Conf. on AI}}, Vol.~\bibinfo{volume}{22}. \bibinfo{pages}{615}.
\newblock


\bibitem[\protect\citeauthoryear{Miralles, van~den Berg, Gash, Parinussa,
  de~Jeu, Beck, Holmes, Jim{\'e}nez, Verhoest, Dorigo, et~al\mbox{.}}{Miralles
  et~al\mbox{.}}{2014}]%
        {miralles2014nino}
\bibfield{author}{\bibinfo{person}{D.~G. Miralles} {}et al.}
  \bibinfo{year}{2014}\natexlab{}.
\newblock \showarticletitle{El Ni{\~n}o--La Ni{\~n}a cycle and recent trends in
  continental evaporation}.
\newblock \bibinfo{journal}{{\em Nature Climate Change\/}} \bibinfo{volume}{4},
  \bibinfo{number}{2} (\bibinfo{year}{2014}), \bibinfo{pages}{122--126}.
\newblock


\bibitem[\protect\citeauthoryear{Mithal, Garg, Boriah, Steinbach, Kumar,
  Potter, Klooster, and Castilla-Rubio}{Mithal et~al\mbox{.}}{2011a}]%
        {mithal2011monitoring}
\bibfield{author}{\bibinfo{person}{V. Mithal} {}et al.}
  \bibinfo{year}{2011}\natexlab{a}.
\newblock \showarticletitle{Monitoring global forest cover using data mining}.
\newblock \bibinfo{journal}{{\em ACM Transactions on Intelligent Systems and
  Technology (TIST)\/}} \bibinfo{volume}{2}, \bibinfo{number}{4}
  (\bibinfo{year}{2011}), \bibinfo{pages}{36}.
\newblock


\bibitem[\protect\citeauthoryear{Mithal, Garg, Brugere, Boriah, Kumar,
  Steinbach, Potter, , and Klooster}{Mithal et~al\mbox{.}}{2011b}]%
        {mithalCIDU2011}
\bibfield{author}{\bibinfo{person}{V. Mithal} {}et al.}
  \bibinfo{year}{2011}\natexlab{b}.
\newblock \showarticletitle{Incorporating Natural Variation into Time
  Series-Based Land Cover Change Identification}. In \bibinfo{booktitle}{{\em
  CIDU'11: Proceedings of the 2011 NASA Conference on Intelligent Data
  Understanding}}.
\newblock


\bibitem[\protect\citeauthoryear{Mithal, O'Connor, Steinhaeuser, Boriah, Kumar,
  Potter, and Klooster}{Mithal et~al\mbox{.}}{2012}]%
        {mithal2012time}
\bibfield{author}{\bibinfo{person}{V. Mithal} {}et al.}
  \bibinfo{year}{2012}\natexlab{}.
\newblock \showarticletitle{Time series change detection using segmentation: A
  case study for land cover monitoring}. In \bibinfo{booktitle}{{\em
  Intelligent Data Understanding (CIDU), 2012 Conference on}}. IEEE,
  \bibinfo{pages}{63--70}.
\newblock


\bibitem[\protect\citeauthoryear{Mohan, Shekhar, Shine, and Rogers}{Mohan
  et~al\mbox{.}}{2010}]%
        {mohan2010cascading}
\bibfield{author}{\bibinfo{person}{P. Mohan} {}et al.}
  \bibinfo{year}{2010}\natexlab{}.
\newblock \showarticletitle{Cascading Spatio-temporal Pattern Discovery: A
  Summary of Results}. In \bibinfo{booktitle}{{\em SDM}}.
  \bibinfo{pages}{327--338}.
\newblock


\bibitem[\protect\citeauthoryear{Mohan, Shekhar, Shine, and Rogers}{Mohan
  et~al\mbox{.}}{2012}]%
        {mohan2012cascading}
\bibfield{author}{\bibinfo{person}{P. Mohan} {}et al.}
  \bibinfo{year}{2012}\natexlab{}.
\newblock \showarticletitle{Cascading spatio-temporal pattern discovery}.
\newblock \bibinfo{journal}{{\em TKDE\/}} \bibinfo{volume}{24},
  \bibinfo{number}{11} (\bibinfo{year}{2012}), \bibinfo{pages}{1977--1992}.
\newblock


\bibitem[\protect\citeauthoryear{Montgomery, Jennings, and Kulahci}{Montgomery
  et~al\mbox{.}}{2015}]%
        {montgomery2015introduction}
\bibfield{author}{\bibinfo{person}{D.~C. Montgomery} {}et al.}
  \bibinfo{year}{2015}\natexlab{}.
\newblock \bibinfo{booktitle}{{\em Introduction to time series analysis and
  forecasting}}.
\newblock \bibinfo{publisher}{John Wiley \& Sons}.
\newblock


\bibitem[\protect\citeauthoryear{Morris and Trivedi}{Morris and
  Trivedi}{2009}]%
        {morris2009learning}
\bibfield{author}{\bibinfo{person}{B. Morris} {}et al.}
  \bibinfo{year}{2009}\natexlab{}.
\newblock \showarticletitle{Learning trajectory patterns by clustering:
  Experimental studies and comparative evaluation}. In \bibinfo{booktitle}{{\em
  CVPR}}. IEEE, \bibinfo{pages}{312--319}.
\newblock


\bibitem[\protect\citeauthoryear{Moscheni, Bhattacharjee, and Kunt}{Moscheni
  et~al\mbox{.}}{1998}]%
        {moscheni1998spatio}
\bibfield{author}{\bibinfo{person}{F. Moscheni} {}et al.}
  \bibinfo{year}{1998}\natexlab{}.
\newblock \showarticletitle{Spatio-temporal segmentation based on region
  merging}.
\newblock \bibinfo{journal}{{\em TPAMI\/}} \bibinfo{volume}{20},
  \bibinfo{number}{9}, \bibinfo{pages}{897--915}.
\newblock


\bibitem[\protect\citeauthoryear{Mueen}{Mueen}{2014}]%
        {mueen2014time}
\bibfield{author}{\bibinfo{person}{A. Mueen}.} \bibinfo{year}{2014}\natexlab{}.
\newblock \showarticletitle{Time series motif discovery: dimensions and
  applications}.
\newblock \bibinfo{journal}{{\em Wiley IR: DMKD\/}} \bibinfo{volume}{4},
  \bibinfo{number}{2}, \bibinfo{pages}{152--159}.
\newblock


\bibitem[\protect\citeauthoryear{Newman}{Newman}{2006}]%
        {newman2006modularity}
\bibfield{author}{\bibinfo{person}{M.~E. Newman}.}
  \bibinfo{year}{2006}\natexlab{}.
\newblock \showarticletitle{Modularity and community structure in networks}.
\newblock \bibinfo{journal}{{\em PNAS\/}} \bibinfo{volume}{103},
  \bibinfo{number}{23} (\bibinfo{year}{2006}), \bibinfo{pages}{8577--8582}.
\newblock


\bibitem[\protect\citeauthoryear{Ng and Han}{Ng and Han}{2002}]%
        {ng2002clarans}
\bibfield{author}{\bibinfo{person}{R.~T. Ng} {}et al.}
  \bibinfo{year}{2002}\natexlab{}.
\newblock \showarticletitle{Clarans: A method for clustering objects for
  spatial data mining}.
\newblock \bibinfo{journal}{{\em TKDE\/}} \bibinfo{volume}{14},
  \bibinfo{number}{5}, \bibinfo{pages}{1003--1016}.
\newblock


\bibitem[\protect\citeauthoryear{Norman, Polyn, Detre, and Haxby}{Norman
  et~al\mbox{.}}{2006}]%
        {norman2006beyond}
\bibfield{author}{\bibinfo{person}{K.~A. Norman} {}et al.}
  \bibinfo{year}{2006}\natexlab{}.
\newblock \showarticletitle{Beyond mind-reading: multi-voxel pattern analysis
  of fMRI data}.
\newblock \bibinfo{journal}{{\em Trends in cognitive sciences\/}}
  \bibinfo{volume}{10}, \bibinfo{number}{9} (\bibinfo{year}{2006}),
  \bibinfo{pages}{424--430}.
\newblock


\bibitem[\protect\citeauthoryear{Oliver and Webster}{Oliver and
  Webster}{1990}]%
        {oliver1990kriging}
\bibfield{author}{\bibinfo{person}{M.~A. Oliver} {}et al.}
  \bibinfo{year}{1990}\natexlab{}.
\newblock \showarticletitle{Kriging: a method of interpolation for geographical
  information systems}.
\newblock \bibinfo{journal}{{\em International Journal of Geographical
  Information System\/}} \bibinfo{volume}{4}, \bibinfo{number}{3}
  (\bibinfo{year}{1990}), \bibinfo{pages}{313--332}.
\newblock


\bibitem[\protect\citeauthoryear{Pettersson-Yeo, Allen, Benetti, McGuire, and
  Mechelli}{Pettersson-Yeo et~al\mbox{.}}{2011}]%
        {pettersson2011dysconnectivity}
\bibfield{author}{\bibinfo{person}{W. Pettersson-Yeo} {}et al.}
  \bibinfo{year}{2011}\natexlab{}.
\newblock \showarticletitle{Dysconnectivity in schizophrenia: where are we
  now?}
\newblock \bibinfo{journal}{{\em Neuroscience \& Biobehavioral Reviews\/}}
  \bibinfo{volume}{35}, \bibinfo{number}{5} (\bibinfo{year}{2011}),
  \bibinfo{pages}{1110--1124}.
\newblock


\bibitem[\protect\citeauthoryear{Pillai, Angryk, and Aydin}{Pillai
  et~al\mbox{.}}{2013}]%
        {pillai2013filter}
\bibfield{author}{\bibinfo{person}{K.~G. Pillai} {}et al.}
  \bibinfo{year}{2013}\natexlab{}.
\newblock \showarticletitle{A filter-and-refine approach to mine spatiotemporal
  co-occurrences}. In \bibinfo{booktitle}{{\em SIGSPATIAL Intnl. Conf. on Adv.
  in GIS}}. ACM, \bibinfo{pages}{104--113}.
\newblock


\bibitem[\protect\citeauthoryear{Pillai, Angryk, Banda, Schuh, and
  Wylie}{Pillai et~al\mbox{.}}{2012}]%
        {pillai2012spatio}
\bibfield{author}{\bibinfo{person}{K.~G. Pillai} {}et al.}
  \bibinfo{year}{2012}\natexlab{}.
\newblock \showarticletitle{Spatio-temporal co-occurrence pattern mining in
  data sets with evolving regions}. In \bibinfo{booktitle}{{\em ICDM
  Workshops}}. IEEE, \bibinfo{pages}{805--812}.
\newblock


\bibitem[\protect\citeauthoryear{Pillai, Angryk, Banda, Wylie, and
  Schuh}{Pillai et~al\mbox{.}}{2014}]%
        {pillai2014spatiotemporal}
\bibfield{author}{\bibinfo{person}{K.~G. Pillai} {}et al.}
  \bibinfo{year}{2014}\natexlab{}.
\newblock \showarticletitle{Spatiotemporal co-occurrence rules}.
\newblock In \bibinfo{booktitle}{{\em New Trends in DB \& IS}}.
  \bibinfo{publisher}{Springer}, \bibinfo{pages}{27--35}.
\newblock


\bibitem[\protect\citeauthoryear{Prati, Mikic, Trivedi, and Cucchiara}{Prati
  et~al\mbox{.}}{2003}]%
        {prati2003detecting}
\bibfield{author}{\bibinfo{person}{A. Prati} {}et al.}
  \bibinfo{year}{2003}\natexlab{}.
\newblock \showarticletitle{Detecting moving shadows: algorithms and
  evaluation}.
\newblock \bibinfo{journal}{{\em TPAMI\/}} \bibinfo{volume}{25},
  \bibinfo{number}{7} (\bibinfo{year}{2003}), \bibinfo{pages}{918--923}.
\newblock


\bibitem[\protect\citeauthoryear{Rabiner and Juang}{Rabiner and Juang}{1986}]%
        {rabiner1986introduction}
\bibfield{author}{\bibinfo{person}{L. Rabiner} {}et al.}
  \bibinfo{year}{1986}\natexlab{}.
\newblock \showarticletitle{An introduction to hidden Markov models}.
\newblock \bibinfo{journal}{{\em ASSP\/}} \bibinfo{volume}{3},
  \bibinfo{number}{1} (\bibinfo{year}{1986}), \bibinfo{pages}{4--16}.
\newblock


\bibitem[\protect\citeauthoryear{Raichle and Snyder}{Raichle and
  Snyder}{2007}]%
        {raichle2007default}
\bibfield{author}{\bibinfo{person}{M.~E. Raichle} {}et al.}
  \bibinfo{year}{2007}\natexlab{}.
\newblock \showarticletitle{A default mode of brain function: a brief history
  of an evolving idea}.
\newblock \bibinfo{journal}{{\em Neuroimage\/}} \bibinfo{volume}{37},
  \bibinfo{number}{4} (\bibinfo{year}{2007}), \bibinfo{pages}{1083--1090}.
\newblock


\bibitem[\protect\citeauthoryear{Rubner, Tomasi, and Guibas}{Rubner
  et~al\mbox{.}}{2000}]%
        {rubner2000earth}
\bibfield{author}{\bibinfo{person}{Y. Rubner} {}et al.}
  \bibinfo{year}{2000}\natexlab{}.
\newblock \showarticletitle{The earth mover's distance as a metric for image
  retrieval}.
\newblock \bibinfo{journal}{{\em IJCV\/}} \bibinfo{volume}{40},
  \bibinfo{number}{2} (\bibinfo{year}{2000}), \bibinfo{pages}{99--121}.
\newblock


\bibitem[\protect\citeauthoryear{Ryan, LeMasters, Biswas, Levin, Hu, Lindsey,
  Bernstein, Lockey, Villareal, Hershey, et~al\mbox{.}}{Ryan
  et~al\mbox{.}}{2007}]%
        {ryan2007comparison}
\bibfield{author}{\bibinfo{person}{P.~H. Ryan} {}et al.}
  \bibinfo{year}{2007}\natexlab{}.
\newblock \showarticletitle{A comparison of proximity and land use regression
  traffic exposure models and wheezing in infants}.
\newblock \bibinfo{journal}{{\em Environmental health perspectives\/}}
  (\bibinfo{year}{2007}), \bibinfo{pages}{278--284}.
\newblock


\bibitem[\protect\citeauthoryear{Saha, Moorthi, Pan, Wu, Wang, Nadiga, Tripp,
  Kistler, Woollen, Behringer, et~al\mbox{.}}{Saha et~al\mbox{.}}{2010}]%
        {saha2010ncep}
\bibfield{author}{\bibinfo{person}{S. Saha} {}et al.}
  \bibinfo{year}{2010}\natexlab{}.
\newblock \showarticletitle{The NCEP climate forecast system reanalysis}.
\newblock \bibinfo{journal}{{\em Bulletin of the American Meteorological
  Society\/}} \bibinfo{volume}{91}, \bibinfo{number}{8} (\bibinfo{year}{2010}),
  \bibinfo{pages}{1015--1057}.
\newblock


\bibitem[\protect\citeauthoryear{Salmon, Olivier, Wessels, Kleynhans, van~den
  Bergh, and Steenkamp}{Salmon et~al\mbox{.}}{2011}]%
        {Salmon2011}
\bibfield{author}{\bibinfo{person}{B.~P. Salmon} {}et al.}
  \bibinfo{year}{2011}\natexlab{}.
\newblock \showarticletitle{{Unsupervised Land Cover Change Detection:
  Meaningful Sequential Time Series Analysis}}.
\newblock \bibinfo{journal}{{\em J of Selected Topics in Applied Earth Obs. and
  Remote Sensing\/}} \bibinfo{volume}{4}, \bibinfo{number}{2}
  (\bibinfo{date}{June} \bibinfo{year}{2011}), \bibinfo{pages}{327--335}.
\newblock
\showISSN{1939-1404}


\bibitem[\protect\citeauthoryear{Schroder, Rehrauer, Seidel, and
  Datcu}{Schroder et~al\mbox{.}}{1998}]%
        {schroder1998spatial}
\bibfield{author}{\bibinfo{person}{M. Schroder} {}et al.}
  \bibinfo{year}{1998}\natexlab{}.
\newblock \showarticletitle{Spatial information retrieval from remote-sensing
  images. II. Gibbs-Markov random fields}.
\newblock \bibinfo{journal}{{\em Geoscience and Remote Sensing, IEEE
  Transactions on\/}} \bibinfo{volume}{36}, \bibinfo{number}{5}
  (\bibinfo{year}{1998}), \bibinfo{pages}{1446--1455}.
\newblock


\bibitem[\protect\citeauthoryear{Shekhar and Chawla}{Shekhar and
  Chawla}{2003}]%
        {shashi2003spatial}
\bibfield{author}{\bibinfo{person}{S. Shekhar} {}et al.}
  \bibinfo{year}{2003}\natexlab{}.
\newblock \bibinfo{title}{Spatial databases: a tour}.
\newblock   (\bibinfo{year}{2003}).
\newblock


\bibitem[\protect\citeauthoryear{Shekhar, Evans, Kang, and Mohan}{Shekhar
  et~al\mbox{.}}{2011}]%
        {shekhar2011identifying}
\bibfield{author}{\bibinfo{person}{S. Shekhar} {}et al.}
  \bibinfo{year}{2011}\natexlab{}.
\newblock \showarticletitle{Identifying patterns in spatial information: A
  survey of methods}.
\newblock \bibinfo{journal}{{\em Wiley Interdisciplinary Reviews: Data Mining
  and Knowledge Discovery\/}} \bibinfo{volume}{1}, \bibinfo{number}{3}
  (\bibinfo{year}{2011}), \bibinfo{pages}{193--214}.
\newblock


\bibitem[\protect\citeauthoryear{Shekhar, Jiang, Ali, Eftelioglu, Tang,
  Gunturi, and Zhou}{Shekhar et~al\mbox{.}}{2015}]%
        {shekhar2015spatiotemporal}
\bibfield{author}{\bibinfo{person}{S. Shekhar} {}et al.}
  \bibinfo{year}{2015}\natexlab{}.
\newblock \showarticletitle{Spatiotemporal data mining: A computational
  perspective}.
\newblock \bibinfo{journal}{{\em ISPRS International Journal of
  Geo-Information\/}} \bibinfo{volume}{4}, \bibinfo{number}{4}
  (\bibinfo{year}{2015}), \bibinfo{pages}{2306--2338}.
\newblock


\bibitem[\protect\citeauthoryear{Shekhar, Lu, and Zhang}{Shekhar
  et~al\mbox{.}}{2001}]%
        {shekhar2001detecting}
\bibfield{author}{\bibinfo{person}{S. Shekhar} {}et al.}
  \bibinfo{year}{2001}\natexlab{}.
\newblock \showarticletitle{Detecting graph-based spatial outliers: algorithms
  and applications (a summary of results)}. In \bibinfo{booktitle}{{\em
  SIGKDD}}. ACM, \bibinfo{pages}{371--376}.
\newblock


\bibitem[\protect\citeauthoryear{Shekhar, Vatsavai, and Celik}{Shekhar
  et~al\mbox{.}}{2008}]%
        {shekhar2008spatial}
\bibfield{author}{\bibinfo{person}{S. Shekhar} {}et al.}
  \bibinfo{year}{2008}\natexlab{}.
\newblock \showarticletitle{Spatial and spatiotemporal data mining: Recent
  advances}.
\newblock \bibinfo{journal}{{\em Data Mining: Next Generation Challenges and
  Future Directions\/}} (\bibinfo{year}{2008}), \bibinfo{pages}{1--34}.
\newblock


\bibitem[\protect\citeauthoryear{Shen-Orr, Milo, Mangan, and Alon}{Shen-Orr
  et~al\mbox{.}}{2002}]%
        {shen2002network}
\bibfield{author}{\bibinfo{person}{S.~S. Shen-Orr} {}et al.}
  \bibinfo{year}{2002}\natexlab{}.
\newblock \showarticletitle{Network motifs in the transcriptional regulation
  network of Escherichia coli}.
\newblock \bibinfo{journal}{{\em Nature genetics\/}} \bibinfo{volume}{31},
  \bibinfo{number}{1} (\bibinfo{year}{2002}), \bibinfo{pages}{64--68}.
\newblock


\bibitem[\protect\citeauthoryear{Smith, Fox, Miller, Glahn, Fox, Mackay,
  Filippini, Watkins, Toro, Laird, et~al\mbox{.}}{Smith et~al\mbox{.}}{2009}]%
        {smith2009correspondence}
\bibfield{author}{\bibinfo{person}{S.~M. Smith} {}et al.}
  \bibinfo{year}{2009}\natexlab{}.
\newblock \showarticletitle{Correspondence of the brain's functional
  architecture during activation and rest}.
\newblock \bibinfo{journal}{{\em Proceedings of the National Academy of
  Sciences\/}} \bibinfo{volume}{106}, \bibinfo{number}{31}
  (\bibinfo{year}{2009}), \bibinfo{pages}{13040--13045}.
\newblock


\bibitem[\protect\citeauthoryear{Soundarajan, Eliassi-Rad, and
  Gallagher}{Soundarajan et~al\mbox{.}}{2013}]%
        {soundarajan2013network}
\bibfield{author}{\bibinfo{person}{S. Soundarajan} {}et al.}
  \bibinfo{year}{2013}\natexlab{}.
\newblock \showarticletitle{Which network similarity measure should you choose:
  an empirical study}. In \bibinfo{booktitle}{{\em Workshop on Information in
  Networks, New York, USA}}.
\newblock


\bibitem[\protect\citeauthoryear{Sporns and K{\"o}tter}{Sporns and
  K{\"o}tter}{2004}]%
        {sporns2004motifs}
\bibfield{author}{\bibinfo{person}{O. Sporns} {}et al.}
  \bibinfo{year}{2004}\natexlab{}.
\newblock \showarticletitle{Motifs in brain networks}.
\newblock \bibinfo{journal}{{\em PLoS Biol\/}} \bibinfo{volume}{2},
  \bibinfo{number}{11} (\bibinfo{year}{2004}), \bibinfo{pages}{e369}.
\newblock


\bibitem[\protect\citeauthoryear{Steinbach, Tan, Kumar, Klooster, and
  Potter}{Steinbach et~al\mbox{.}}{2003}]%
        {steinbach2003discovery}
\bibfield{author}{\bibinfo{person}{M. Steinbach} {}et al.}
  \bibinfo{year}{2003}\natexlab{}.
\newblock \showarticletitle{Discovery of climate indices using clustering}. In
  \bibinfo{booktitle}{{\em SIGKDD}}. ACM, \bibinfo{pages}{446--455}.
\newblock


\bibitem[\protect\citeauthoryear{Steinbach, Tan, Kumar, Potter, Klooster, and
  Torregrosa}{Steinbach et~al\mbox{.}}{2002}]%
        {steinbach2002data}
\bibfield{author}{\bibinfo{person}{M. Steinbach} {}et al.}
  \bibinfo{year}{2002}\natexlab{}.
\newblock \showarticletitle{Data mining for the discovery of ocean climate
  indices}. In \bibinfo{booktitle}{{\em Scientific Data Mining}}.
\newblock


\bibitem[\protect\citeauthoryear{Steinhaeuser and Tsonis}{Steinhaeuser and
  Tsonis}{2014}]%
        {steinhaeuser2014climate}
\bibfield{author}{\bibinfo{person}{K. Steinhaeuser} {}et al.}
  \bibinfo{year}{2014}\natexlab{}.
\newblock \showarticletitle{A climate model intercomparison at the dynamics
  level}.
\newblock \bibinfo{journal}{{\em Climate dynamics\/}} \bibinfo{volume}{42},
  \bibinfo{number}{5-6} (\bibinfo{year}{2014}), \bibinfo{pages}{1665--1670}.
\newblock


\bibitem[\protect\citeauthoryear{Stiles and Ghosh}{Stiles and Ghosh}{1997}]%
        {stiles1997habituation}
\bibfield{author}{\bibinfo{person}{B.~W. Stiles} {}et al.}
  \bibinfo{year}{1997}\natexlab{}.
\newblock \showarticletitle{Habituation based neural networks for
  spatio-temporal classification}.
\newblock \bibinfo{journal}{{\em Neurocomputing\/}} \bibinfo{volume}{15},
  \bibinfo{number}{3} (\bibinfo{year}{1997}), \bibinfo{pages}{273--307}.
\newblock


\bibitem[\protect\citeauthoryear{Sui, Adali, Pearlson, and Calhoun}{Sui
  et~al\mbox{.}}{2009}]%
        {sui2009ica}
\bibfield{author}{\bibinfo{person}{J. Sui} {}et al.}
  \bibinfo{year}{2009}\natexlab{}.
\newblock \showarticletitle{An ICA-based method for the identification of
  optimal FMRI features and components using combined group-discriminative
  techniques}.
\newblock \bibinfo{journal}{{\em Neuroimage\/}} \bibinfo{volume}{46},
  \bibinfo{number}{1} (\bibinfo{year}{2009}), \bibinfo{pages}{73--86}.
\newblock


\bibitem[\protect\citeauthoryear{Sun and Chawla}{Sun and Chawla}{2004}]%
        {sun2004local}
\bibfield{author}{\bibinfo{person}{P. Sun} {}et al.}
  \bibinfo{year}{2004}\natexlab{}.
\newblock \showarticletitle{On local spatial outliers}. In
  \bibinfo{booktitle}{{\em ICDM}}. IEEE, \bibinfo{pages}{209--216}.
\newblock


\bibitem[\protect\citeauthoryear{Takahashi, Kulldorff, Tango, and
  Yih}{Takahashi et~al\mbox{.}}{2008}]%
        {takahashi2008flexibly}
\bibfield{author}{\bibinfo{person}{K. Takahashi} {}et al.}
  \bibinfo{year}{2008}\natexlab{}.
\newblock \showarticletitle{A flexibly shaped space-time scan statistic for
  disease outbreak detection and monitoring}.
\newblock \bibinfo{journal}{{\em International Journal of Health
  Geographics\/}} \bibinfo{volume}{7}, \bibinfo{number}{1}
  (\bibinfo{year}{2008}), \bibinfo{pages}{14}.
\newblock


\bibitem[\protect\citeauthoryear{Takahashi, Hooi, and Faloutsos}{Takahashi
  et~al\mbox{.}}{2017}]%
        {takahashi2017autocyclone}
\bibfield{author}{\bibinfo{person}{T. Takahashi} {}et al.}
  \bibinfo{year}{2017}\natexlab{}.
\newblock \showarticletitle{AutoCyclone: Automatic Mining of Cyclic Online
  Activities with Robust Tensor Factorization}. In \bibinfo{booktitle}{{\em
  Proceedings of the 26th International Conference on World Wide Web}}.
  International World Wide Web Conferences Steering Committee,
  \bibinfo{pages}{213--221}.
\newblock


\bibitem[\protect\citeauthoryear{Takeishi and Yairi}{Takeishi and
  Yairi}{2014}]%
        {takeishi2014anomaly}
\bibfield{author}{\bibinfo{person}{N. Takeishi} {}et al.}
  \bibinfo{year}{2014}\natexlab{}.
\newblock \showarticletitle{Anomaly detection from multivariate time-series
  with sparse representation}. In \bibinfo{booktitle}{{\em Systems, Man and
  Cybernetics (SMC), 2014 IEEE International Conference on}}. IEEE,
  \bibinfo{pages}{2651--2656}.
\newblock


\bibitem[\protect\citeauthoryear{Tan, Steinbach, Karpatne, and Kumar}{Tan
  et~al\mbox{.}}{2017}]%
        {tan2018introduction}
\bibfield{author}{\bibinfo{person}{P.-N. Tan} {}et al.}
  \bibinfo{year}{2017}\natexlab{}.
\newblock \bibinfo{booktitle}{{\em Introduction to Data Mining}}.
\newblock \bibinfo{publisher}{Second Edition. Pearson Addison-Wesley (in
  print)}.
\newblock


\bibitem[\protect\citeauthoryear{Tang, Chang, and Liu}{Tang
  et~al\mbox{.}}{2014}]%
        {tang2014mining}
\bibfield{author}{\bibinfo{person}{J. Tang} {}et al.}
  \bibinfo{year}{2014}\natexlab{}.
\newblock \showarticletitle{Mining social media with social theories: a
  survey}.
\newblock \bibinfo{journal}{{\em KDD Explorations\/}} \bibinfo{volume}{15},
  \bibinfo{number}{2} (\bibinfo{year}{2014}), \bibinfo{pages}{20--29}.
\newblock


\bibitem[\protect\citeauthoryear{Tang, Rothbart, and Posner}{Tang
  et~al\mbox{.}}{2012}]%
        {tang2012neural}
\bibfield{author}{\bibinfo{person}{Y.-Y. Tang} {}et al.}
  \bibinfo{year}{2012}\natexlab{}.
\newblock \showarticletitle{Neural correlates of establishing, maintaining, and
  switching brain states}.
\newblock \bibinfo{journal}{{\em Trends in cognitive sciences\/}}
  \bibinfo{volume}{16}, \bibinfo{number}{6} (\bibinfo{year}{2012}),
  \bibinfo{pages}{330--337}.
\newblock


\bibitem[\protect\citeauthoryear{Tango, Takahashi, and Kohriyama}{Tango
  et~al\mbox{.}}{2011}]%
        {tango2011space}
\bibfield{author}{\bibinfo{person}{T. Tango} {}et al.}
  \bibinfo{year}{2011}\natexlab{}.
\newblock \showarticletitle{A space--time scan statistic for detecting emerging
  outbreaks}.
\newblock \bibinfo{journal}{{\em Biometrics\/}} \bibinfo{volume}{67},
  \bibinfo{number}{1} (\bibinfo{year}{2011}), \bibinfo{pages}{106--115}.
\newblock


\bibitem[\protect\citeauthoryear{Taylor, Fergus, LeCun, and Bregler}{Taylor
  et~al\mbox{.}}{2010}]%
        {taylor2010convolutional}
\bibfield{author}{\bibinfo{person}{G.~W. Taylor} {}et al.}
  \bibinfo{year}{2010}\natexlab{}.
\newblock \showarticletitle{Convolutional learning of spatio-temporal
  features}. In \bibinfo{booktitle}{{\em European conference on computer
  vision}}. Springer, \bibinfo{pages}{140--153}.
\newblock


\bibitem[\protect\citeauthoryear{Thompson, Rausch, Saari, and Selin}{Thompson
  et~al\mbox{.}}{2014}]%
        {thompson2014systems}
\bibfield{author}{\bibinfo{person}{T.~M. Thompson} {}et al.}
  \bibinfo{year}{2014}\natexlab{}.
\newblock \showarticletitle{A systems approach to evaluating the air quality
  co-benefits of US carbon policies}.
\newblock \bibinfo{journal}{{\em Nature Climate Change\/}} \bibinfo{volume}{4},
  \bibinfo{number}{10} (\bibinfo{year}{2014}), \bibinfo{pages}{917--923}.
\newblock


\bibitem[\protect\citeauthoryear{Tompson, Johnson, Ashby, Perkins, and
  Edwards}{Tompson et~al\mbox{.}}{2015}]%
        {tompson2015uk}
\bibfield{author}{\bibinfo{person}{L. Tompson} {}et al.}
  \bibinfo{year}{2015}\natexlab{}.
\newblock \showarticletitle{UK open source crime data: accuracy and
  possibilities for research}.
\newblock \bibinfo{journal}{{\em Cartography and Geographic Information
  Science\/}} \bibinfo{volume}{42}, \bibinfo{number}{2} (\bibinfo{year}{2015}),
  \bibinfo{pages}{97--111}.
\newblock


\bibitem[\protect\citeauthoryear{Toohey and Duckham}{Toohey and
  Duckham}{2015}]%
        {toohey2015trajectory}
\bibfield{author}{\bibinfo{person}{K. Toohey} {}et al.}
  \bibinfo{year}{2015}\natexlab{}.
\newblock \showarticletitle{Trajectory similarity measures}.
\newblock \bibinfo{journal}{{\em SIGSPATIAL Special\/}} \bibinfo{volume}{7},
  \bibinfo{number}{1} (\bibinfo{year}{2015}), \bibinfo{pages}{43--50}.
\newblock


\bibitem[\protect\citeauthoryear{Torkamani and Lohweg}{Torkamani and
  Lohweg}{2017}]%
        {torkamani2017survey}
\bibfield{author}{\bibinfo{person}{S. Torkamani} {}et al.}
  \bibinfo{year}{2017}\natexlab{}.
\newblock \showarticletitle{Survey on time series motif discovery}.
\newblock \bibinfo{journal}{{\em Wiley IR: DMKD\/}} \bibinfo{volume}{7},
  \bibinfo{number}{2} (\bibinfo{year}{2017}).
\newblock


\bibitem[\protect\citeauthoryear{Trasarti, Pinelli, Nanni, and
  Giannotti}{Trasarti et~al\mbox{.}}{2011}]%
        {trasarti2011mining}
\bibfield{author}{\bibinfo{person}{R. Trasarti} {}et al.}
  \bibinfo{year}{2011}\natexlab{}.
\newblock \showarticletitle{Mining mobility user profiles for car pooling}. In
  \bibinfo{booktitle}{{\em SIGKDD}}. ACM, \bibinfo{pages}{1190--1198}.
\newblock


\bibitem[\protect\citeauthoryear{Tsoukatos and Gunopulos}{Tsoukatos and
  Gunopulos}{2001}]%
        {tsoukatos2001efficient}
\bibfield{author}{\bibinfo{person}{I. Tsoukatos} {}et al.}
  \bibinfo{year}{2001}\natexlab{}.
\newblock \showarticletitle{Efficient mining of spatiotemporal patterns}. In
  \bibinfo{booktitle}{{\em ISSTD}}. Springer, \bibinfo{pages}{425--442}.
\newblock


\bibitem[\protect\citeauthoryear{Tye, Blenkinsop, Fowler, Stephenson, and
  Kilsby}{Tye et~al\mbox{.}}{2016}]%
        {tye2016simulating}
\bibfield{author}{\bibinfo{person}{M.~R. Tye} {}et al.}
  \bibinfo{year}{2016}\natexlab{}.
\newblock \showarticletitle{Simulating multimodal seasonality in extreme daily
  precipitation occurrence}.
\newblock \bibinfo{journal}{{\em Journal of Hydrology\/}}
  \bibinfo{volume}{537} (\bibinfo{year}{2016}), \bibinfo{pages}{117--129}.
\newblock


\bibitem[\protect\citeauthoryear{Van Den~Heuvel, Mandl, and Hulshoff~Pol}{Van
  Den~Heuvel et~al\mbox{.}}{2008}]%
        {van2008normalized}
\bibfield{author}{\bibinfo{person}{M. Van Den~Heuvel} {}et al.}
  \bibinfo{year}{2008}\natexlab{}.
\newblock \showarticletitle{Normalized cut group clustering of resting-state
  FMRI data}.
\newblock \bibinfo{journal}{{\em PloS one\/}} \bibinfo{volume}{3},
  \bibinfo{number}{4} (\bibinfo{year}{2008}), \bibinfo{pages}{e2001}.
\newblock


\bibitem[\protect\citeauthoryear{Vatsavai}{Vatsavai}{2008}]%
        {vatsavai2008machine}
\bibfield{author}{\bibinfo{person}{R.~R. Vatsavai}.}
  \bibinfo{year}{2008}\natexlab{}.
\newblock \bibinfo{booktitle}{{\em Machine learning algorithms for
  spatio-temporal data mining}}.
\newblock \bibinfo{publisher}{ProQuest}.
\newblock


\bibitem[\protect\citeauthoryear{Vatsavai, Ganguly, Chandola, Stefanidis,
  Klasky, and Shekhar}{Vatsavai et~al\mbox{.}}{2012}]%
        {vatsavai2012spatiotemporal}
\bibfield{author}{\bibinfo{person}{R.~R. Vatsavai} {}et al.}
  \bibinfo{year}{2012}\natexlab{}.
\newblock \showarticletitle{Spatiotemporal data mining in the era of big
  spatial data: algorithms and applications}. In \bibinfo{booktitle}{{\em
  SIGSPATIAL international workshop on analytics for big geospatial data}}.
  ACM, \bibinfo{pages}{1--10}.
\newblock


\bibitem[\protect\citeauthoryear{Verhein and Chawla}{Verhein and
  Chawla}{2006}]%
        {verhein2006mining}
\bibfield{author}{\bibinfo{person}{F. Verhein} {}et al.}
  \bibinfo{year}{2006}\natexlab{}.
\newblock \showarticletitle{Mining spatio-temporal association rules, sources,
  sinks, stationary regions and thoroughfares in object mobility databases}. In
  \bibinfo{booktitle}{{\em DASFAA}}, Vol.~\bibinfo{volume}{3882}. Springer,
  \bibinfo{pages}{187--201}.
\newblock


\bibitem[\protect\citeauthoryear{Verhein and Chawla}{Verhein and
  Chawla}{2008}]%
        {verhein2008mining}
\bibfield{author}{\bibinfo{person}{F. Verhein} {}et al.}
  \bibinfo{year}{2008}\natexlab{}.
\newblock \showarticletitle{Mining spatio-temporal patterns in object mobility
  databases}.
\newblock \bibinfo{journal}{{\em Data mining and knowledge discovery\/}}
  \bibinfo{volume}{16}, \bibinfo{number}{1} (\bibinfo{year}{2008}),
  \bibinfo{pages}{5--38}.
\newblock


\bibitem[\protect\citeauthoryear{Vieira, Bakalov, and Tsotras}{Vieira
  et~al\mbox{.}}{2009}]%
        {vieira2009line}
\bibfield{author}{\bibinfo{person}{M.~R. Vieira} {}et al.}
  \bibinfo{year}{2009}\natexlab{}.
\newblock \showarticletitle{On-line discovery of flock patterns in
  spatio-temporal data}. In \bibinfo{booktitle}{{\em SIGSPATIAL}}. ACM,
  \bibinfo{pages}{286--295}.
\newblock


\bibitem[\protect\citeauthoryear{Voldoire, Sanchez-Gomez, y~M{\'e}lia,
  Decharme, Cassou, S{\'e}n{\'e}si, Valcke, Beau, Alias, Chevallier,
  et~al\mbox{.}}{Voldoire et~al\mbox{.}}{2013}]%
        {voldoire2013cnrm}
\bibfield{author}{\bibinfo{person}{A. Voldoire} {}et al.}
  \bibinfo{year}{2013}\natexlab{}.
\newblock \showarticletitle{The CNRM-CM5. 1 global climate model: description
  and basic evaluation}.
\newblock \bibinfo{journal}{{\em Climate Dynamics\/}} \bibinfo{volume}{40},
  \bibinfo{number}{9-10} (\bibinfo{year}{2013}), \bibinfo{pages}{2091--2121}.
\newblock


\bibitem[\protect\citeauthoryear{Walther and Kaisser}{Walther and
  Kaisser}{2013}]%
        {walther2013geo}
\bibfield{author}{\bibinfo{person}{M. Walther} {}et al.}
  \bibinfo{year}{2013}\natexlab{}.
\newblock \showarticletitle{Geo-spatial event detection in the twitter stream}.
  In \bibinfo{booktitle}{{\em ECIR}}. Springer, \bibinfo{pages}{356--367}.
\newblock


\bibitem[\protect\citeauthoryear{Wang, Wu, and Chen}{Wang
  et~al\mbox{.}}{2013}]%
        {wang2013finding}
\bibfield{author}{\bibinfo{person}{L. Wang} {}et al.}
  \bibinfo{year}{2013}\natexlab{}.
\newblock \showarticletitle{Finding probabilistic prevalent colocations in
  spatially uncertain data sets}.
\newblock \bibinfo{journal}{{\em TKDE\/}} \bibinfo{volume}{25},
  \bibinfo{number}{4} (\bibinfo{year}{2013}), \bibinfo{pages}{790--804}.
\newblock


\bibitem[\protect\citeauthoryear{Wang, Yin, Chen, Sun, Sadiq, and Zhou}{Wang
  et~al\mbox{.}}{2017}]%
        {wang2017st}
\bibfield{author}{\bibinfo{person}{W. Wang} {}et al.}
  \bibinfo{year}{2017}\natexlab{}.
\newblock \showarticletitle{ST-SAGE: A Spatial-Temporal Sparse Additive
  Generative Model for Spatial Item Recommendation}.
\newblock \bibinfo{journal}{{\em ACM Transactions on Intelligent Systems and
  Technology (TIST)\/}} \bibinfo{volume}{8}, \bibinfo{number}{3}
  (\bibinfo{year}{2017}), \bibinfo{pages}{48}.
\newblock


\bibitem[\protect\citeauthoryear{Wei, Kumar, Lolla, Keogh, Lonardi, and
  Ratanamahatana}{Wei et~al\mbox{.}}{2005}]%
        {wei2005assumption}
\bibfield{author}{\bibinfo{person}{L. Wei} {}et al.}
  \bibinfo{year}{2005}\natexlab{}.
\newblock \showarticletitle{Assumption-Free Anomaly Detection in Time Series.}.
  In \bibinfo{booktitle}{{\em SSDBM}}, Vol.~\bibinfo{volume}{5}.
  \bibinfo{pages}{237--242}.
\newblock


\bibitem[\protect\citeauthoryear{Weng and Lee}{Weng and Lee}{2011}]%
        {weng2011event}
\bibfield{author}{\bibinfo{person}{J. Weng} {}et al.}
  \bibinfo{year}{2011}\natexlab{}.
\newblock \showarticletitle{Event Detection in Twitter}. In
  \bibinfo{booktitle}{{\em AAAI Conf. on Weblogs and Social Media}}.
\newblock


\bibitem[\protect\citeauthoryear{Whitcher, Guttorp, and Percival}{Whitcher
  et~al\mbox{.}}{2000}]%
        {whitcher2000multiscale}
\bibfield{author}{\bibinfo{person}{B. Whitcher} {}et al.}
  \bibinfo{year}{2000}\natexlab{}.
\newblock \showarticletitle{Multiscale detection and location of multiple
  variance changes in the presence of long memory}.
\newblock \bibinfo{journal}{{\em Journal of Statistical Computation and
  Simulation\/}} \bibinfo{volume}{68}, \bibinfo{number}{1}
  (\bibinfo{year}{2000}), \bibinfo{pages}{65--87}.
\newblock


\bibitem[\protect\citeauthoryear{Wu, Liu, and Chawla}{Wu et~al\mbox{.}}{2010}]%
        {wu2010spatio}
\bibfield{author}{\bibinfo{person}{E. Wu} {}et al.}
  \bibinfo{year}{2010}\natexlab{}.
\newblock \showarticletitle{Spatio-temporal outlier detection in precipitation
  data}.
\newblock In \bibinfo{booktitle}{{\em Knowledge discovery from sensor data}}.
  \bibinfo{publisher}{Springer}, \bibinfo{pages}{115--133}.
\newblock


\bibitem[\protect\citeauthoryear{Xiao, Xie, Luo, and Ma}{Xiao
  et~al\mbox{.}}{2008}]%
        {xiao2008density}
\bibfield{author}{\bibinfo{person}{X. Xiao} {}et al.}
  \bibinfo{year}{2008}\natexlab{}.
\newblock \showarticletitle{Density based co-location pattern discovery}. In
  \bibinfo{booktitle}{{\em SIGSPATIAL}}. ACM, \bibinfo{pages}{29}.
\newblock


\bibitem[\protect\citeauthoryear{Yang and DelSole}{Yang and DelSole}{2012}]%
        {yang2012systematic}
\bibfield{author}{\bibinfo{person}{X. Yang} {}et al.}
  \bibinfo{year}{2012}\natexlab{}.
\newblock \showarticletitle{Systematic comparison of ENSO teleconnection
  patterns between models and observations}.
\newblock \bibinfo{journal}{{\em Journal of Climate\/}} \bibinfo{volume}{25},
  \bibinfo{number}{2} (\bibinfo{year}{2012}), \bibinfo{pages}{425--446}.
\newblock


\bibitem[\protect\citeauthoryear{Ye and Keogh}{Ye and Keogh}{2009}]%
        {ye2009time}
\bibfield{author}{\bibinfo{person}{L. Ye} {}et al.}
  \bibinfo{year}{2009}\natexlab{}.
\newblock \showarticletitle{Time series shapelets: a new primitive for data
  mining}. In \bibinfo{booktitle}{{\em SIGKDD}}. ACM,
  \bibinfo{pages}{947--956}.
\newblock


\bibitem[\protect\citeauthoryear{Yeh, Zhu, Ulanova, Begum, Ding, Dau, Silva,
  Mueen, and Keogh}{Yeh et~al\mbox{.}}{2016}]%
        {yeh2016matrix}
\bibfield{author}{\bibinfo{person}{C.-C.~M. Yeh} {}et al.}
  \bibinfo{year}{2016}\natexlab{}.
\newblock \showarticletitle{Matrix Profile I: All Pairs Similarity Joins for
  Time Series: A Unifying View that Includes Motifs, Discords and Shapelets}.
  In \bibinfo{booktitle}{{\em IEEE ICDM}}.
\newblock


\bibitem[\protect\citeauthoryear{Yu, Erhardt, Sui, Du, He, Hjelm, Cetin,
  Rachakonda, Miller, Pearlson, et~al\mbox{.}}{Yu et~al\mbox{.}}{2015b}]%
        {yu2015assessing}
\bibfield{author}{\bibinfo{person}{Q. Yu} {}et al.}
  \bibinfo{year}{2015}\natexlab{b}.
\newblock \showarticletitle{Assessing dynamic brain graphs of time-varying
  connectivity in fMRI data: Application to healthy controls and patients with
  schizophrenia}.
\newblock \bibinfo{journal}{{\em NeuroImage\/}}  \bibinfo{volume}{107}
  (\bibinfo{year}{2015}), \bibinfo{pages}{345--355}.
\newblock


\bibitem[\protect\citeauthoryear{Yu, Cheng, and Liu}{Yu et~al\mbox{.}}{2015a}]%
        {yu2015accelerated}
\bibfield{author}{\bibinfo{person}{R. Yu} {}et al.}
  \bibinfo{year}{2015}\natexlab{a}.
\newblock \showarticletitle{Accelerated online low rank tensor learning for
  multivariate spatiotemporal streams}. In \bibinfo{booktitle}{{\em
  International Conference on Machine Learning}}. \bibinfo{pages}{238--247}.
\newblock


\bibitem[\protect\citeauthoryear{Zeinalipour-Yazti, Lin, and
  Gunopulos}{Zeinalipour-Yazti et~al\mbox{.}}{2006}]%
        {zeinalipour2006distributed}
\bibfield{author}{\bibinfo{person}{D. Zeinalipour-Yazti} {}et al.}
  \bibinfo{year}{2006}\natexlab{}.
\newblock \showarticletitle{Distributed spatio-temporal similarity search}. In
  \bibinfo{booktitle}{{\em Proceedings of the 15th ACM international conference
  on Information and knowledge management}}. ACM, \bibinfo{pages}{14--23}.
\newblock


\bibitem[\protect\citeauthoryear{Zhang, Zhang, Yuan, Zhang, Hanratty, and
  Han}{Zhang et~al\mbox{.}}{2016}]%
        {zhang2016gmove}
\bibfield{author}{\bibinfo{person}{C. Zhang} {}et al.}
  \bibinfo{year}{2016}\natexlab{}.
\newblock \showarticletitle{GMove: Group-Level Mobility Modeling Using
  Geo-Tagged Social Media}. In \bibinfo{booktitle}{{\em KDD}}.
  \bibinfo{pages}{1305--1314}.
\newblock


\bibitem[\protect\citeauthoryear{Zhao, Sun, Ye, Chen, Lu, and
  Ramakrishnan}{Zhao et~al\mbox{.}}{2015}]%
        {zhao2015multi}
\bibfield{author}{\bibinfo{person}{L. Zhao} {}et al.}
  \bibinfo{year}{2015}\natexlab{}.
\newblock \showarticletitle{Multi-task learning for spatio-temporal event
  forecasting}. In \bibinfo{booktitle}{{\em Proceedings of the 21th ACM SIGKDD
  International Conference on Knowledge Discovery and Data Mining}}. ACM,
  \bibinfo{pages}{1503--1512}.
\newblock


\bibitem[\protect\citeauthoryear{Zhao, Zhang, Li, and Huang}{Zhao
  et~al\mbox{.}}{2007}]%
        {zhao2007classification}
\bibfield{author}{\bibinfo{person}{Y. Zhao} {}et al.}
  \bibinfo{year}{2007}\natexlab{}.
\newblock \showarticletitle{Classification of high spatial resolution imagery
  using improved Gaussian Markov random-field-based texture features}.
\newblock \bibinfo{journal}{{\em Trans. on GeoSc. and Remote Sensing\/}}
  \bibinfo{volume}{45}, \bibinfo{number}{5} (\bibinfo{year}{2007}),
  \bibinfo{pages}{1458--1468}.
\newblock


\bibitem[\protect\citeauthoryear{Zheng}{Zheng}{2015}]%
        {zheng2015trajectory}
\bibfield{author}{\bibinfo{person}{Y. Zheng}.} \bibinfo{year}{2015}\natexlab{}.
\newblock \showarticletitle{Trajectory data mining: an overview}.
\newblock \bibinfo{journal}{{\em ACM Transactions on Intelligent Systems and
  Technology (TIST)\/}} \bibinfo{volume}{6}, \bibinfo{number}{3}
  (\bibinfo{year}{2015}), \bibinfo{pages}{29}.
\newblock


\bibitem[\protect\citeauthoryear{Zheng, Zha, and Chua}{Zheng
  et~al\mbox{.}}{2012}]%
        {zheng2012mining}
\bibfield{author}{\bibinfo{person}{Y.-T. Zheng} {}et al.}
  \bibinfo{year}{2012}\natexlab{}.
\newblock \showarticletitle{Mining travel patterns from geotagged photos}.
\newblock \bibinfo{journal}{{\em TIST\/}} \bibinfo{volume}{3},
  \bibinfo{number}{3} (\bibinfo{year}{2012}), \bibinfo{pages}{56}.
\newblock


\bibitem[\protect\citeauthoryear{Zhou, Li, and Zhu}{Zhou
  et~al\mbox{.}}{2013a}]%
        {zhou2013tensor}
\bibfield{author}{\bibinfo{person}{H. Zhou} {}et al.}
  \bibinfo{year}{2013}\natexlab{a}.
\newblock \showarticletitle{Tensor regression with applications in neuroimaging
  data analysis}.
\newblock \bibinfo{journal}{{\em JASA\/}} \bibinfo{volume}{108},
  \bibinfo{number}{502} (\bibinfo{year}{2013}), \bibinfo{pages}{540--552}.
\newblock


\bibitem[\protect\citeauthoryear{Zhou, Shekhar, and Ali}{Zhou
  et~al\mbox{.}}{2014}]%
        {zhou2014spatiotemporal}
\bibfield{author}{\bibinfo{person}{X. Zhou} {}et al.}
  \bibinfo{year}{2014}\natexlab{}.
\newblock \showarticletitle{Spatiotemporal change footprint pattern discovery:
  an inter-disciplinary survey}.
\newblock \bibinfo{journal}{{\em Wiley IR: DMKD\/}} \bibinfo{volume}{4},
  \bibinfo{number}{1} (\bibinfo{year}{2014}), \bibinfo{pages}{1--23}.
\newblock


\bibitem[\protect\citeauthoryear{Zhou, Shekhar, Mohan, Liess, and Snyder}{Zhou
  et~al\mbox{.}}{2011}]%
        {zhou2011discovering}
\bibfield{author}{\bibinfo{person}{X. Zhou} {}et al.}
  \bibinfo{year}{2011}\natexlab{}.
\newblock \showarticletitle{Discovering interesting sub-paths in spatiotemporal
  datasets: A summary of results}. In \bibinfo{booktitle}{{\em SIGSPATIAL Intl.
  Conf. on advances in geographic information systems}}. ACM,
  \bibinfo{pages}{44--53}.
\newblock


\bibitem[\protect\citeauthoryear{Zhou, Shekhar, and Oliver}{Zhou
  et~al\mbox{.}}{2013b}]%
        {zhou2013discovering}
\bibfield{author}{\bibinfo{person}{X. Zhou} {}et al.}
  \bibinfo{year}{2013}\natexlab{b}.
\newblock \showarticletitle{Discovering persistent change windows in
  spatiotemporal datasets: a summary of results}. In \bibinfo{booktitle}{{\em
  SIGSPATIAL International Workshop on Analytics for Big Geospatial Data}}.
  ACM, \bibinfo{pages}{37--46}.
\newblock


\bibitem[\protect\citeauthoryear{Zhou and Matteson}{Zhou and Matteson}{2015}]%
        {zhou2015predicting}
\bibfield{author}{\bibinfo{person}{Z. Zhou} {}et al.}
  \bibinfo{year}{2015}\natexlab{}.
\newblock \showarticletitle{Predicting ambulance demand: A spatio-temporal
  kernel approach}. In \bibinfo{booktitle}{{\em ACM SIGKDD}}. ACM,
  \bibinfo{pages}{2297--2303}.
\newblock


\bibitem[\protect\citeauthoryear{Zhu, Zimmerman, Senobari, Yeh, Funning, Mueen,
  Brisk, and Keogh}{Zhu et~al\mbox{.}}{2016}]%
        {zhumatrix}
\bibfield{author}{\bibinfo{person}{Y. Zhu} {}et al.}
  \bibinfo{year}{2016}\natexlab{}.
\newblock \showarticletitle{Matrix Profile II: Exploiting a Novel Algorithm and
  GPUs to Break the One Hundred Million Barrier for Time Series Motifs and
  Joins}. In \bibinfo{booktitle}{{\em IEEE ICDM}}.
\newblock


\bibitem[\protect\citeauthoryear{Zhu, Woodcock, and Olofsson}{Zhu
  et~al\mbox{.}}{2012}]%
        {Zhu2012}
\bibfield{author}{\bibinfo{person}{Z. Zhu} {}et al.}
  \bibinfo{year}{2012}\natexlab{}.
\newblock \showarticletitle{Continuous monitoring of forest disturbance using
  all available Landsat imagery}.
\newblock \bibinfo{journal}{{\em Remote Sensing of Environment\/}}
  \bibinfo{volume}{122} (\bibinfo{year}{2012}), \bibinfo{pages}{75--91}.
\newblock


\end{thebibliography}
%\bibliography{acmsmall-sample-bibfile}
                             % Sample .bib file with references that match those in
                             % the 'Specifications Document (V1.5)' as well containing
                             % 'legacy' bibs and bibs with 'alternate codings'.
                             % Gerry Murray - March 2012

\end{document}